\documentclass{article} 
\usepackage{arxiv}

\usepackage[utf8]{inputenc} 
\usepackage[T1]{fontenc}    
\usepackage{hyperref}       
\usepackage{url}            
\usepackage{booktabs}       
\usepackage{amsfonts}       
\usepackage{nicefrac}       
\usepackage{microtype}      
\usepackage{lipsum}		

\usepackage{graphicx}
\usepackage{subfig}
\usepackage{amsmath}
\usepackage{graphicx}
\usepackage{algpseudocode}
\usepackage{algorithm}  
\usepackage{algorithmicx} 
\usepackage{algpseudocode} 
\usepackage{amssymb}
\usepackage{amsmath}
\usepackage{bbding}
\usepackage{multirow}
\usepackage{setspace}
\usepackage{appendix}

\title{A novel tree-structured point cloud dataset for skeletonization algorithm evaluation}

\author{
  Yan Lin \\
  College of Computer Science\\
  Chongqing University\\
  Chongqing 400044, China \\
  \texttt{lyfairy@cqu.edu.cn} \\
   \And
  Ji Liu \\
  College of Computer Science\\
  Chongqing University\\
  Chongqing 400044, China \\
  \texttt{liujiboy@cqu.edu.cn} \\
   \And
  Jianlin Zhou \\
  College of Computer Science\\
  Chongqing University\\
  Chongqing 400044, China \\
  \texttt{jlzhou@cqu.edu.cn} \\
}

\begin{document}
\maketitle
\begin{abstract}
Curve skeleton extraction from unorganized point cloud is a fundamental task of computer vision and three-dimensional data preprocessing and visualization. A great amount of work has been done to extract skeleton from point cloud. but the lack of standard datasets of point cloud with ground truth skeleton makes it difficult to evaluate these algorithms. In this paper, we construct a brand new tree-structured point cloud dataset, including ground truth skeletons, and point cloud models. In addition, four types of point cloud are built on clean point cloud: point clouds with noise, point clouds with missing data, point clouds with different density, and point clouds with uneven density distribution. We first use  tree editor to build the tree skeleton and corresponding mesh model. Since the implicit surface is sufficiently expressive to retain the edges and details of the complex branches model, we use the implicit surface to model the triangular mesh. With the implicit surface, virtual scanner is applied to the sampling of point cloud. Finally, considering the challenges in skeleton extraction, we introduce different methods to build four different types of point cloud models. This dataset can be used as standard dataset for skeleton extraction algorithms. And the evaluation between skeleton extraction algorithms can be performed by comparing the ground truth skeleton with the extracted skeleton.
\end{abstract}


\section{Introduction}
As one-dimensional topological description method of 3D geometric model, curve skeleton extracted from point cloud provides an intuitive and effective representation of underlying geometry feature, has been used in many areas such as computer animation and geometric deformation, model segmentation and 3D reconstruction. Extracting skeleton from point cloud has become a popular research topic in the fields of computer version and data visualization .

There are many ways to capture the point cloud, such as 3D terrestrial laser scanners (TLS), image-based 3D reconstruction methods and multiple view stereo (MVS) technique. Due to the limitation of acquisition method and technology, the obtained point cloud will be disturbed by noise and outliers, heavy data occlusion and uneven density distribution. Therefore, it is very challenging for extracting skeleton from the point cloud.

Util now, many methods have been proposed to extract skeletons from point cloud. For instance, Tagliasacchi et al.\cite{Tagliasacchi2009Curve}propose an algorithm for extracting curve skeleton from imperfect point cloud. Huang et al.\cite{huang2013l}, Mei et al.\cite{Jie20163D} extract skeletons from point cloud with missing and noisy data. Distance field guided $L_1$-median method \cite{Song2018Distance} solves the problem of skeleton extraction from point cloud with uneven density. In these algorithms, the point cloud datasets for extracting skeleton are not unified, coming from public point cloud dataset or raw scanned point cloud. And these datasets usually do not contain ground truth skeletons. Thus, evaluation and comparison between these algorithms has become a difficult problem to solve.

Among all types of point cloud, tree-structured point cloud model is complex case with complex branches. In this paper, we create a new tree-structured point cloud dataset programmatically containing 900 tree models. Each tree model contains point cloud model, and corresponding ground truth skeleton. In the process of modeling, we take into account tree models with different scales, different branch densities, and different branch complexity levels. Besides, these program-built point cloud models will simulate the point cloud scanned from natural environment and are challenging for typical skeleton extraction algorithms. We construct different types of point cloud models with noise, missing data and non-uniform point distribution and different point density. Building this dataset includes the following steps:
\begin{enumerate}
	\item
	Using the tree editor to outline the skeleton of the tree model and generate the corresponding mesh model for later processing.
	\item
	Modeling triangle mesh with implicit surface and simulating the distance scan to obtain point cloud.
	\item
	Constructing four different types of point clouds based on the scanned point cloud : different density of points, noise, missing data, and varying point density distribution.
\end{enumerate}

To demonstrate the effectiveness of our dateset, we test $L_1$ method\cite{huang2013l} on tree point cloud from visually and quantitative aspects. Our dataset is publicly available at \url{https://github.com/liujiboy/TreePointCloud.git}. The point cloud models can be used as standard dataset for skeletonization algorithm. Four different types of point cloud models can be used to assess the robustness of algorithms. In addition,the ground truth skeleton can be used to quantitatively evaluate the skeleton extraction algorithms.

\section{Related Work}
There are many algorithms aimming at extracting curve skeleton. The latest researches are presented at SkelNetOn@CVPR19 workshop, such as Liu et al.\cite{Liu_2019_CVPR_Workshops}, Yang et al.\cite{Yang_2019_CVPR_Workshops}, Atienza et al.\cite{Atienza_2019_CVPR_Workshops} and Demir et al.\cite{Demir_2019_CVPR_Workshops}.  However, these methods are mainly based on deep learning to extract skeletons from 2D images. We only summary the latest works on skeleton extraction from unorganized 3D point cloud.
\subsection{Graph-based method}
 Ogniewicz and Ilg .\cite{Ogniewicz1992Voronoi} use the Voronoi graph of the boundary point and the residual function to reduce the spurious branch from the topological branch. Bucksch et al.\cite{Bucksch2008CAMPINO} propose  a graph-based approach that divides points into octree cells and removes the cycles in this graph to get the skeleton. The result of the skeletonization depends on the depth of the octree, and if the octree cell is large enough, the results of the skeletonization are satisfactory. However, the topology correctness of the skeleton still has problems.

Works\cite{Natali2011Graph} introduce a skeleton representation of a point cloud graph, extending the Reeb diagram to high-latitude  point cloud space representation. They later make some improvements in this method. They\cite{Bucksch2010SkelTre} build octrees and use the extracted octree diagrams to retract the point clouds back to the skeletons and embed them in the Reeb diagrams in order to preserve the topology information. Aim to reconstruct a large number of overlapping tree point clouds, Li et al .\cite{Livny2010Automatic}propose an optical flow based feature-matching algorithm to acquire high-quality point clouds. This method  apply a global optimization strategy to build a branch structure graph(BSG) for each tree and then synthesize a finer structure from the BSG.

Staff's method\cite{Sharf2010On} base on the idea of deformable model. This method tracks the center of the front of the point cloud, then merges and filters the generated arc according to the branch structure to get the curve skeleton of the object. Guo at el.\cite{Guo2010Analysis} propose a deformable model called an arterial snake in order to extract the structure of complex and fine man-made objects. The surface of the objects are defined as tubular shapes. The scan data is fitted by growing a short snake-shaped curved segment from the tubular region.
\subsection{Optimization-based method}
Tagliasacchi et al.\cite{Tagliasacchi2009Curve}propose an algorithm for extracting a curve skeleton from an imperfect point cloud. This method constructs a generalized rotational symmetry axis (ROSA) based on a set of directional points, depending on the a priori of the generally cylindrical shape of the curve skeleton. The skeleton is obtained by minimizing the sum of the projection distances extended from the point cloud normal. It also requires pre-processing of data containing noise and topological defects.

Cao et al.\cite{Cao2010Point} extend the work of Au et al.\cite{au2008skeleton}by making Laplacian-based contraction in slightly incomplete point cloud point cloud data and use opological thinning method to shrink curve skeleton. The algorithm is not ideal for datasets with severe data missing, and the quality of the acquired skeleton is heavily dependent on the tuning parameters.

In order to deal with incomplete point cloud data, Wang et al.\cite{A} propose a global optimization method based on structure-aware global optimization (SAGO). Firstly, the approximate tree skeleton is obtained from the distance minimum spanning tree (DMst). Then stretching direction toward the tree skeleton branch, SAGO method recovers lost data from an incomplete TLS point cloud. Finally, DMst is applied to obtain a refined tree skeleton.

Wang at el.\cite{Zhen2016A} propose a data-driven modeling approach that effectively reconstructs trees from incomplete TLS point clouds. They first extract the rough skeleton from the original point cloud by the minimum distance spanning tree method (DMst). A skeleton-based Laplacian contraction process is proposed to shrink the tree point cloud into a skeleton point cloud. Redundant dictionary is used to learn and reconstruct the local structure to obtain a more accurate point density. 

Huang et al.\cite{huang2013l} restore the key structures from the 3D point cloud by using the $L_1$-median projection operator. The L1 median is used to describe the unique global center of a given set of points. This method obtains the skeleton by gradually expanding the size of the neighborhood to grow and iterate. As an advanced method, the algorithm can directly operate on raw scan data with noise, outliers and large areas of missing data without pre-processing.

A lot of work has been improved on the basis of $L_1$-median. Mei et al.\cite{Jie20163D} improve the method based on the $L_1$-median. The $L_1$-median was used to extract the rough tree skeleton. Considering the dominant direction of each point and the point density of the local region, an iterative optimization process is developed to recover missing data. Finally, they propose a $L_1$-minimum spanning tree (MST) algorithm for refining the tree skeleton from an optimized point cloud.

Song et al.\cite{Song2018Distance}combine the advantages of the distance field and the $L_1$-median method. First, the input point cloud is voxelized to obtain the voxel representation of the point cloud. Then the distance field of the point cloud is calculated according to the theory that the distance between each voxel and the model boundary is the closest. Finally, according to the distance field, a more appropriate domain size can be specified in the $L_1$-median optimization process, avoiding repeated iteration problems. This method is robust and effective to the raw scanned point cloud data.

In a recent work, qin et al.\cite{Qin2019Mass} introduce the concept of mass drive curve skeleton (MdCS) while retaining the geometric properties and mass distribution of the curve skeleton. Skeleton extraction by minimizing the Wasserstein distance between the point cloud and its curved skeleton. The curve skeleton extraction is formulated as the optimal mass transfer problem. The algorithm is directly applied to extract curve skeletons from incomplete point clouds. Not relying on the appropriate neighborhood size makes the algorithm more robust.

\subsection{Point cloud datasets}
In previous works, different algorithms use different point cloud datasets to extract skeleton. Works\cite{Bucksch2010SkelTre}\cite{Zhang20163D} \cite{Jie20163D}\cite{Zhen2016A}focus on extracting skeletons from tree point clouds. TLS is used to scan real trees to acquire tree point cloud with different density, missing and noise features. Tagliasacchi et al.\cite{Tagliasacchi2009Curve} use a virtual scanner to collect point cloud from a set of views around a complete surface model, and gradually remove points captured from different views to simulate incomplete data. Researcher\cite{huang2013l}\cite{Cao2010Point}\cite{Qin2019Mass}\cite{Song2018Distance}use public datasets and various raw scanned point cloud to validate their skeleton extraction algorithms, which usually contain 3D general models of different geometric and topological features: curved grids, highgenus shapes, thin cylindrical components, non-cylindrical geometric shapes and so on.

In recent work\cite{demir2019skelneton}, a 2D point cloud dataset with ground truth is proposed for skeleton extraction by deep learning. They use a 2D skeletonization algorithm\cite{durix2019the} to generate the skeleton and manually adjust it as the ground true skeleton. Our tree-structured dataset is 3d unorganized point cloud models constructed from the corresponding ground truth skeletons.

In a word, there is no comprehensive and consistent dataset to verify the performance of 3D skeletonization algorithms. And these scanned point cloud models also have no ground truth skeletons. Therefore, we devote to constructing a point cloud dataset containing ground truth skeleton.

\section{Synthetic point cloud dataset}
\begin{figure}
	\centering
	{
		\includegraphics[width=0.7\textwidth]{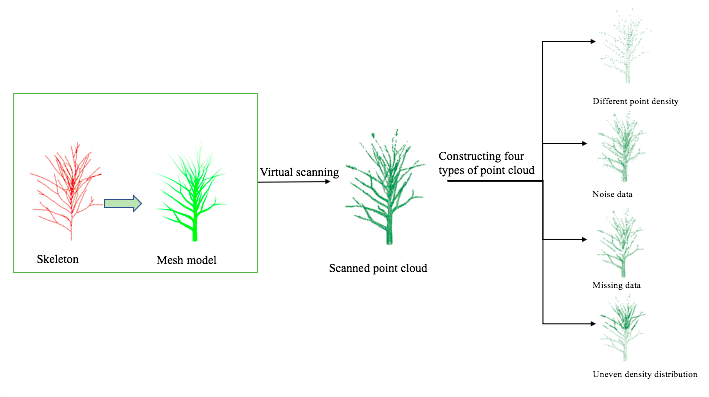}
	}
	\caption{An overview of our point cloud datasets. The left part  is mesh model generated from tree skeleton. The middle figure is the point cloud obtained by implicit surface modeling and virtual scanning. The figure on the right is four different types of point clouds from the scanned point cloud.}
	\label{overview}
\end{figure}

We construct dataset that includes point cloud models of trees and corresponding ground truth skeletons. The construction process of the dataset is shown in the Fig \ref{overview}, divided into four steps: 
\begin{itemize}
	\item 
	 Firstly,we use tree editor to generate tree model, and export mesh model and corresponding ground truth skeleton.
	\item 
	We use the implicit surface to model the smooth triangular mesh.
	\item 
	With implicit surface, a virtual scanner is used to simulate a distance scan to obtain  point cloud.
	\item 
	Generating four types of point clouds by different methods.
\end{itemize}

\subsection{Tree model construction }
Tree is described as a trunk, which is then divided into large branches and then into smaller branches. We use a data-driven editor that combines interactive control and parameter adjustment to easily draw the branch structure and curve skeleton, and save the completed tree as a mesh model for later processing. We model each branch as a cylinder whose skeleton is a 3D linear curve. The advantage of the whole process is that the trees can be modeled finely. the branches of the trees can be stacked multiple times, and the parameters of the trunk and branches are respectively controlled to generate trees with different attitudes.

\begin{figure}
	\centering
	\subfloat[] {
		\includegraphics[width=0.15\textwidth]{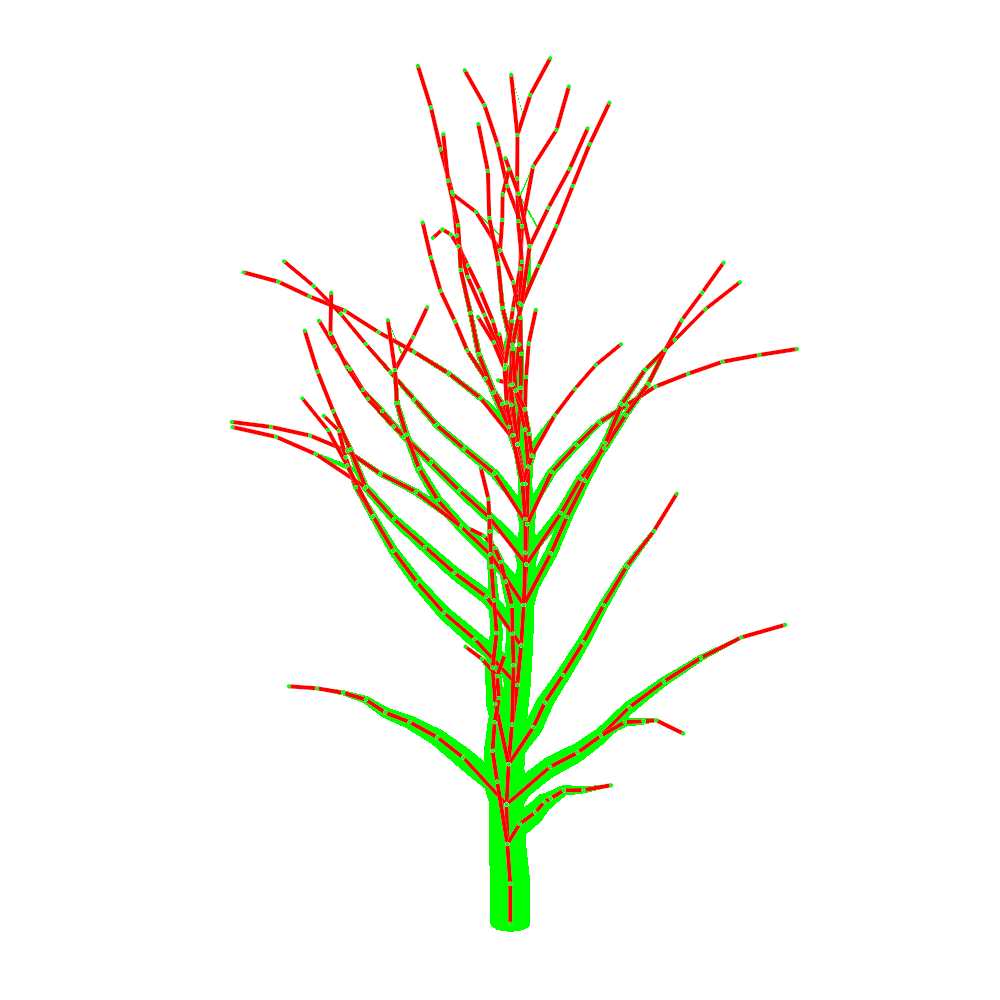}
	}
	\subfloat[] {
		\includegraphics[width=0.15\textwidth]{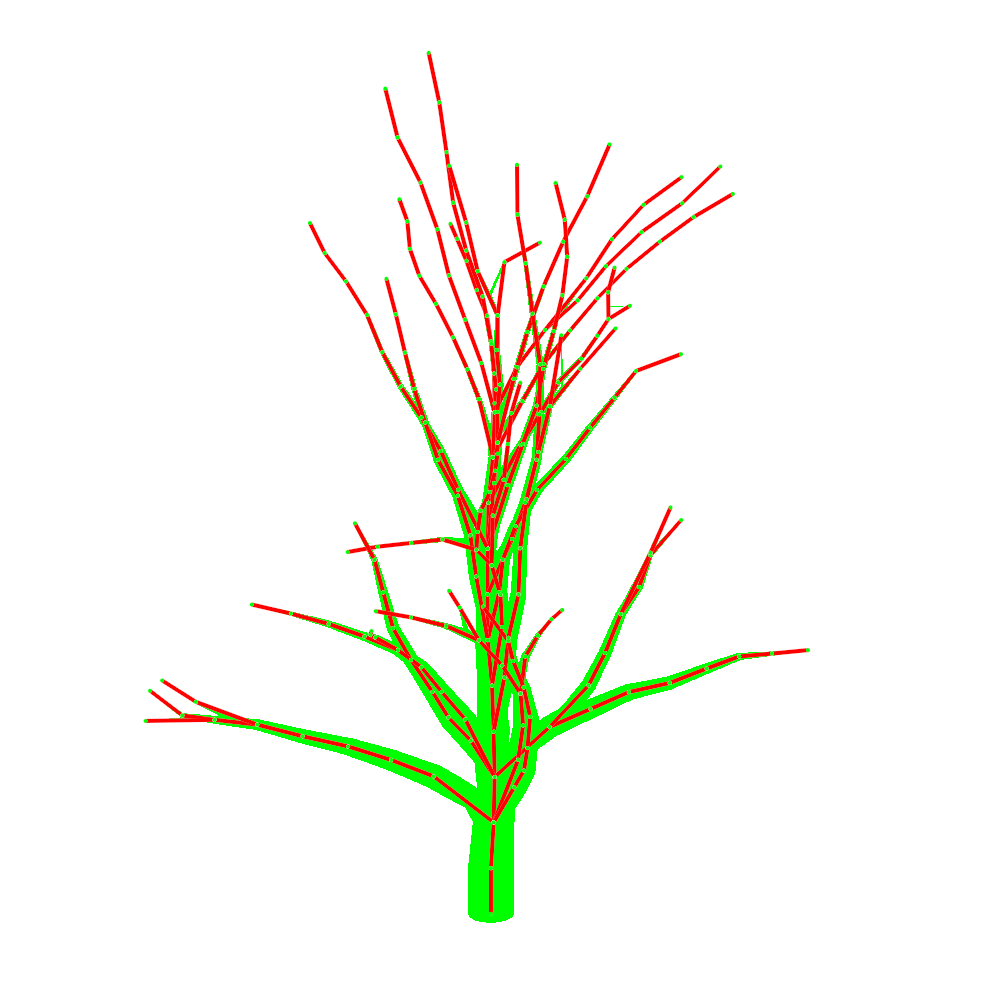}
	}
	\subfloat[] {
		\includegraphics[width=0.15\textwidth]{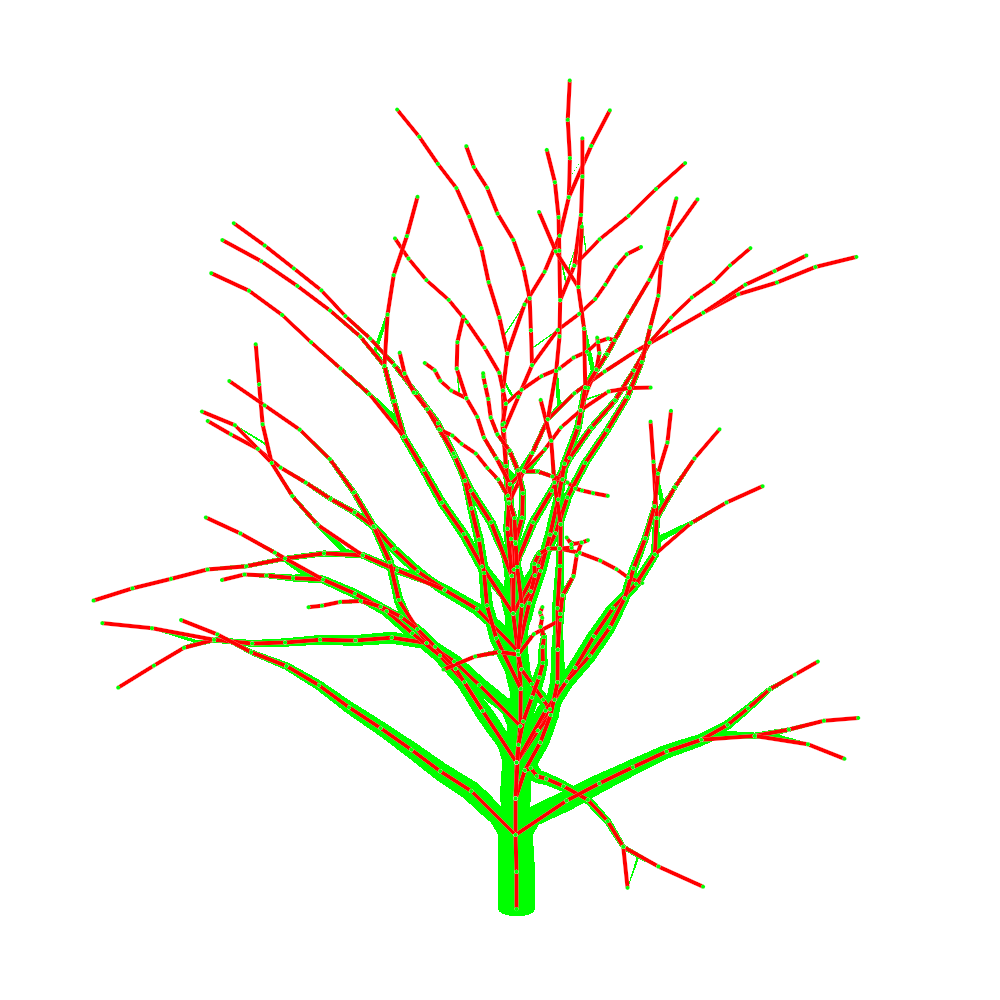}
	}
	\subfloat[] {
		\includegraphics[width=0.15\textwidth]{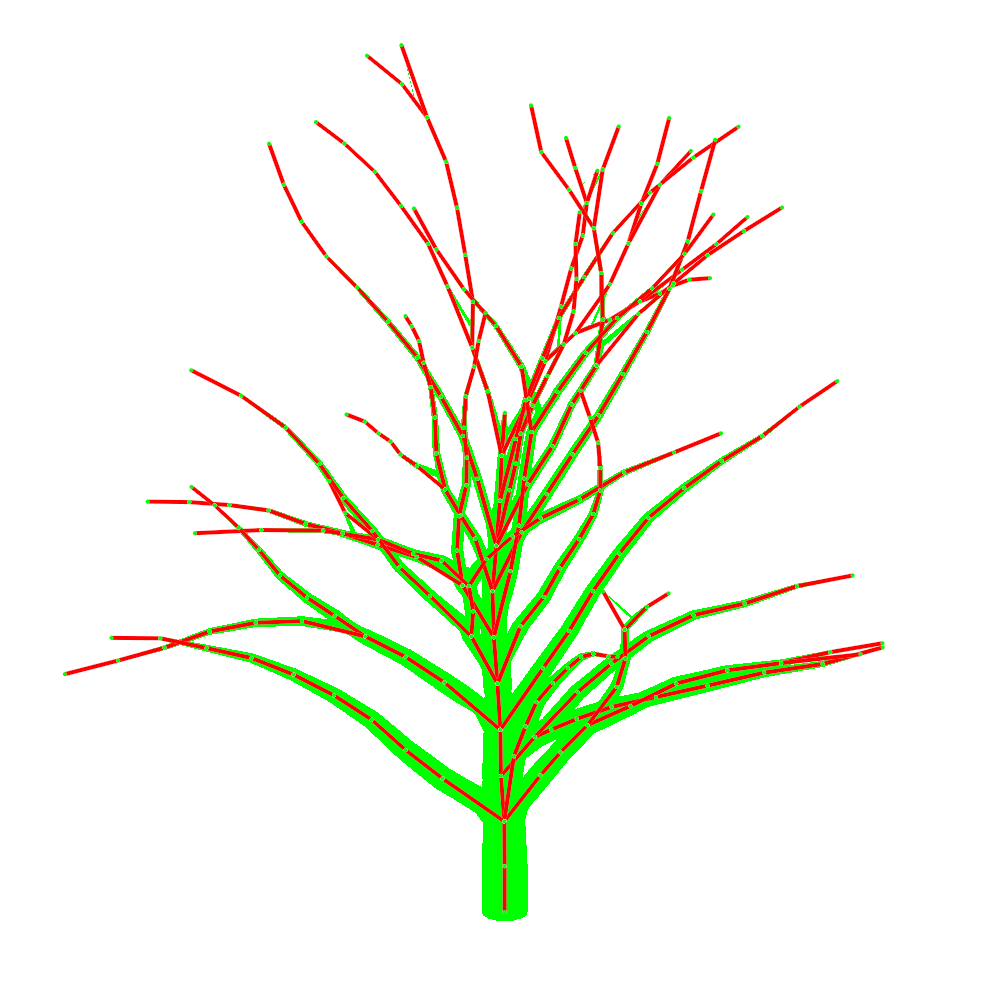}
	}
	\vfill
	\subfloat[] {
		\includegraphics[width=0.15\textwidth]{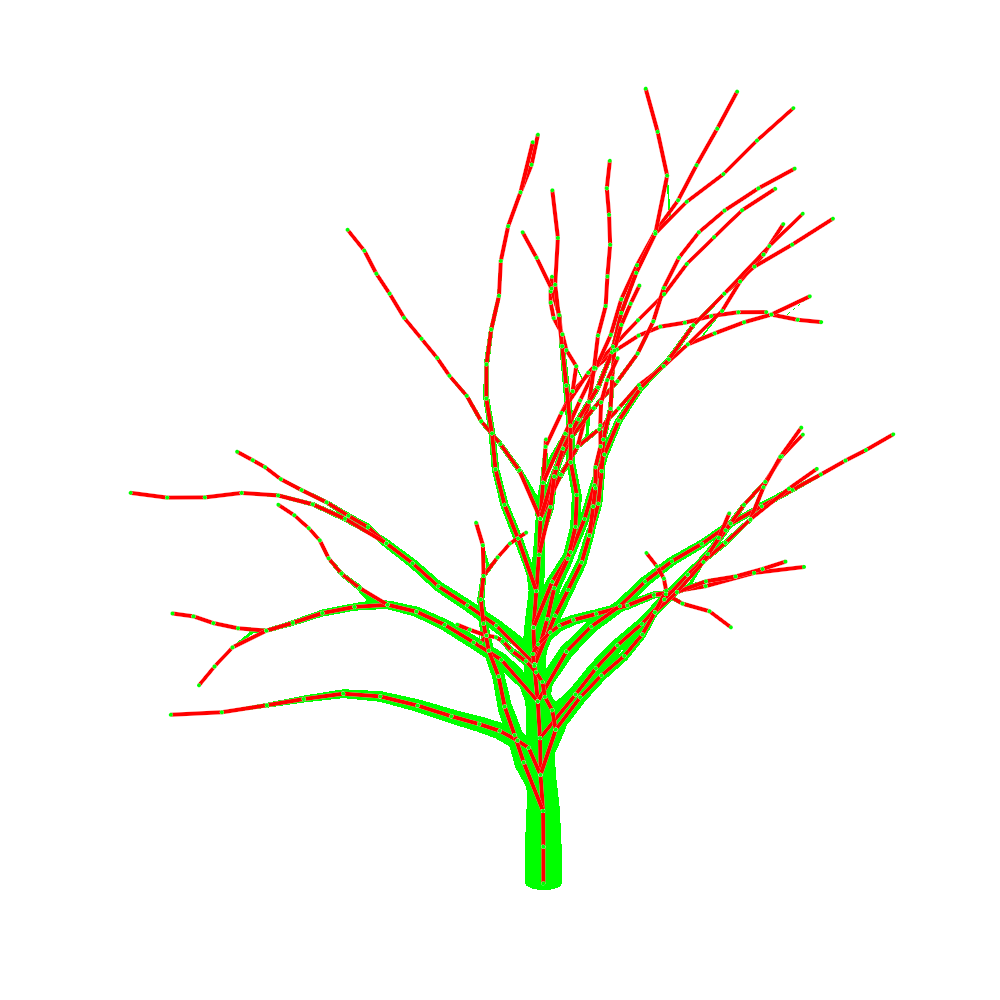}
	}
	\subfloat[] {
		\includegraphics[width=0.15\textwidth]{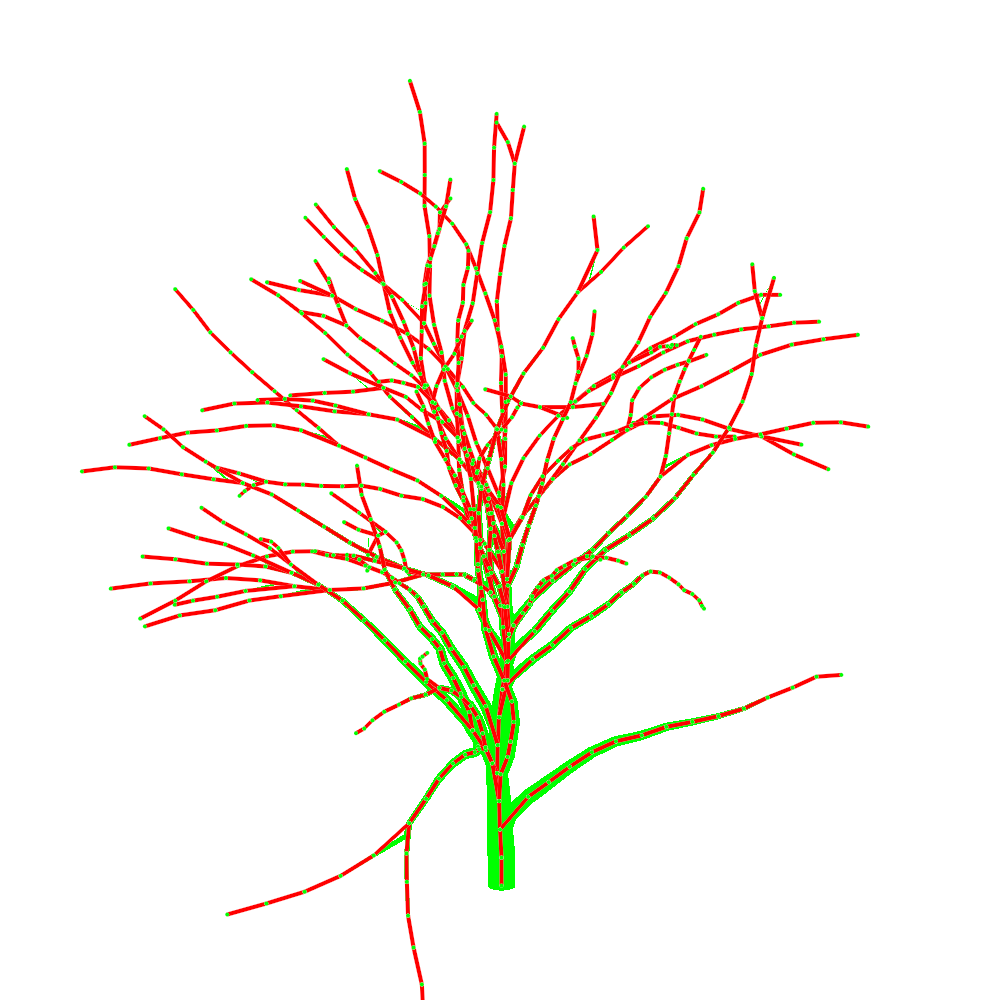}
	}
	\subfloat[] {
		\includegraphics[width=0.15\textwidth]{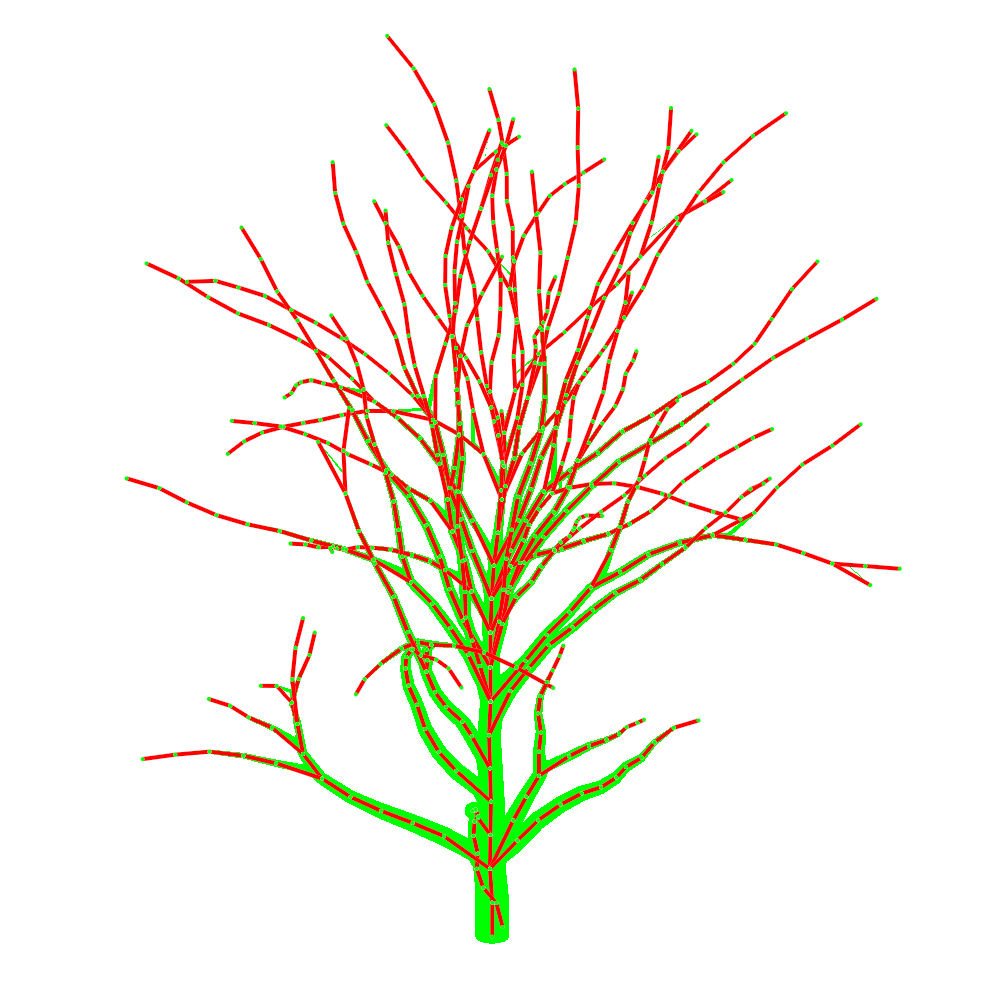}
	}
	\subfloat[] {
		\includegraphics[width=0.15\textwidth]{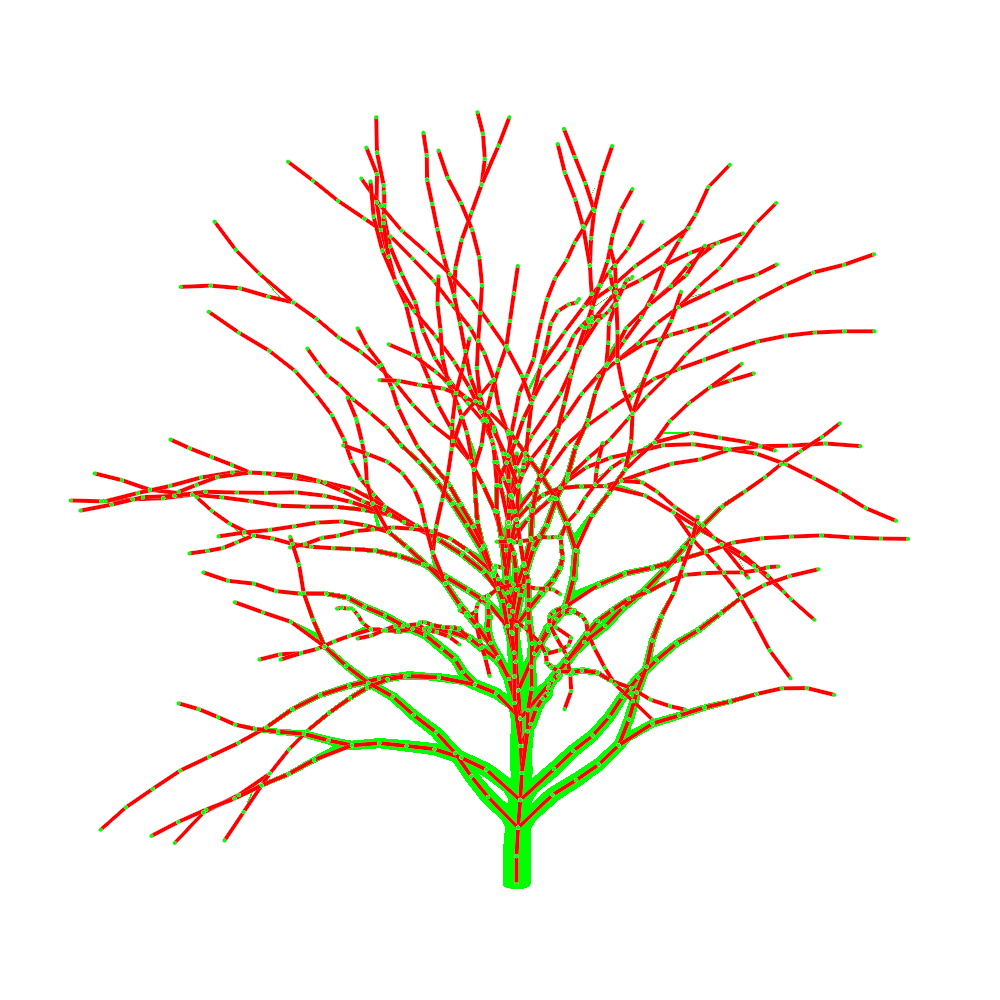}
	}
	\vfill
	\subfloat[] {
		\includegraphics[width=0.15\textwidth]{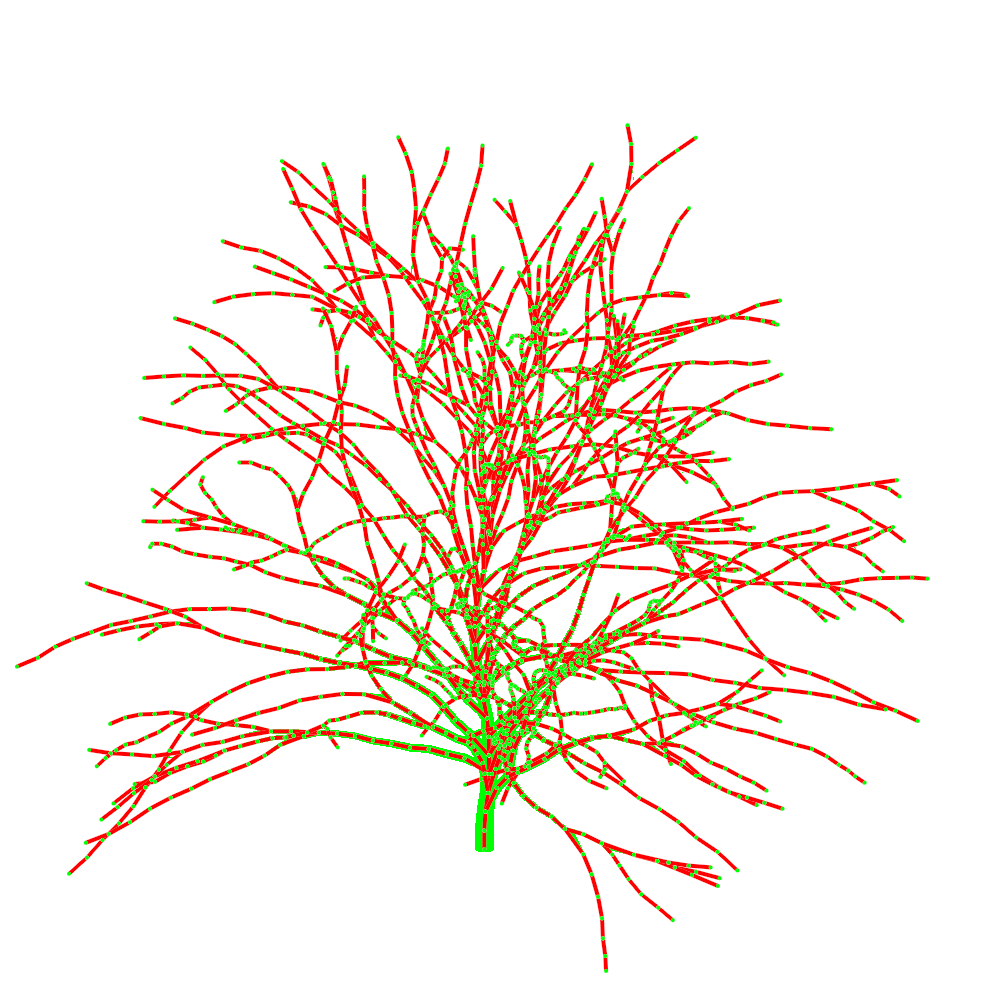}
	}
	\subfloat[] {
		\includegraphics[width=0.15\textwidth]{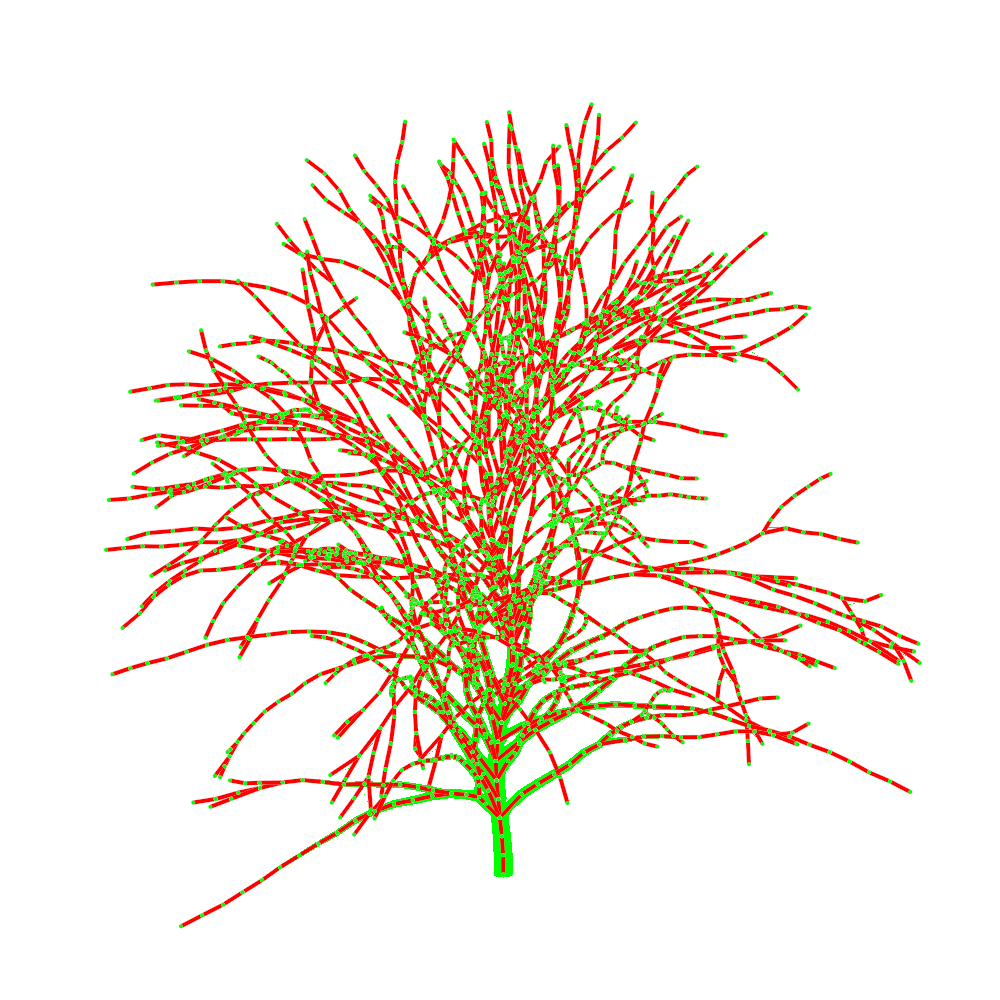}
	}
	\subfloat[] {
		\includegraphics[width=0.15\textwidth]{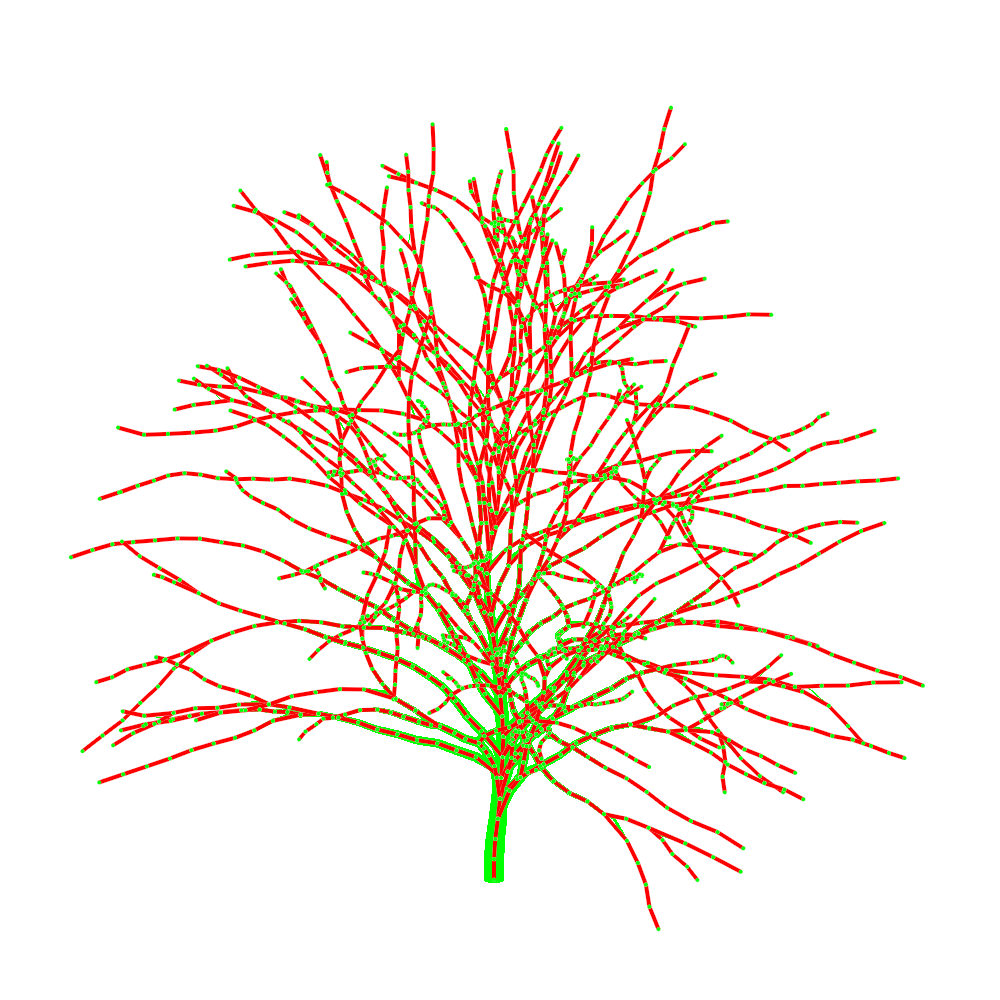}
	}
	\subfloat[] {
		\includegraphics[width=0.15\textwidth]{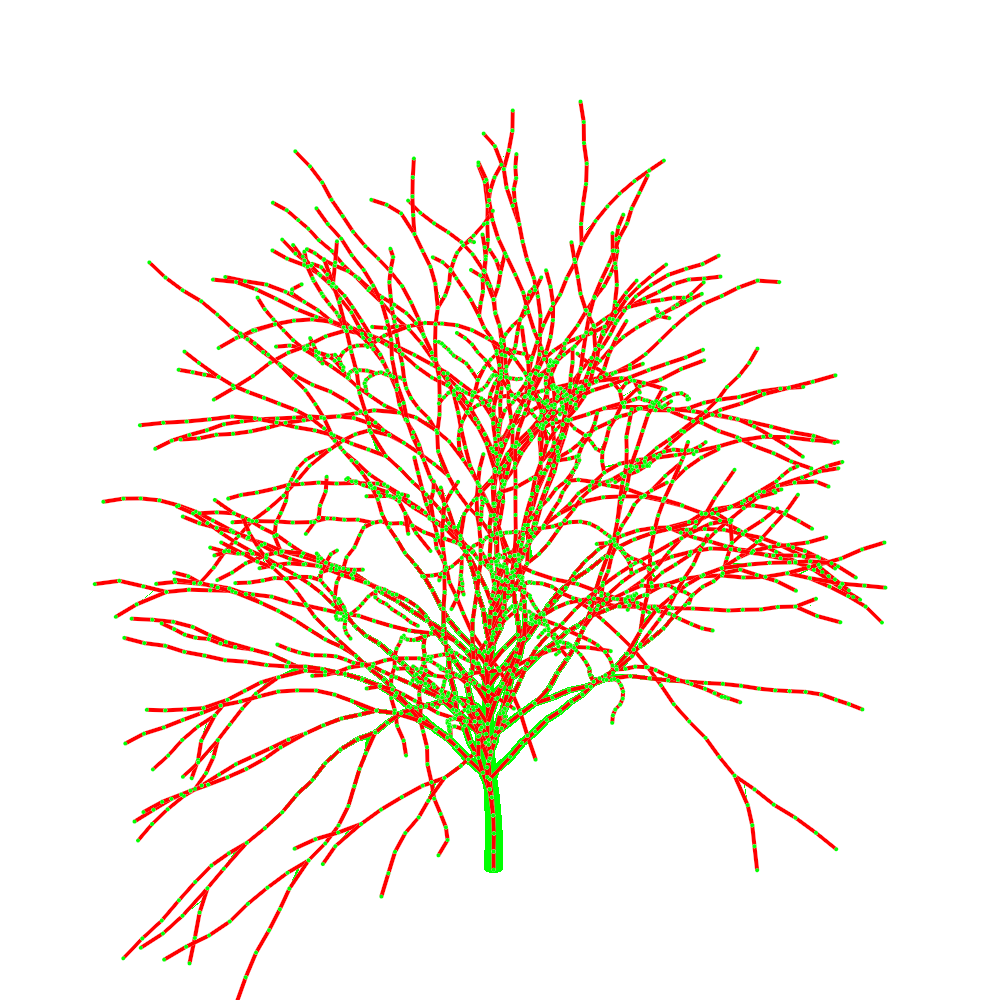}
	}
	\vfill
	\caption{ Several types of trees are generated including ground truth skeletons and mesh models. From the first row to the third row,  the branch complexity and the number of trees increase.}
	\label{skeleton_tree}
\end{figure}

First of all, we create a procedural trunk and modify the parameters in the trunk node editor to control the shape of the trunk: length, radius, bending degree, etc. Then branch nodes grow from the trunk. The trunk is split, and the branch grows. The amount , radius, and growth angle of branches vary in space. Finally, we use the node editor to manage the result trees of all previously created nodes and display the representation in a 3d view. We create tree models of three sizes: small, medium, and large, in which the branches and complexity increase. The skeleton and mesh of the created tree model are show in Fig \ref{skeleton_tree}.

\begin{figure}
	\subfloat[] {
		\includegraphics[width=3.5cm,height=4.5cm]{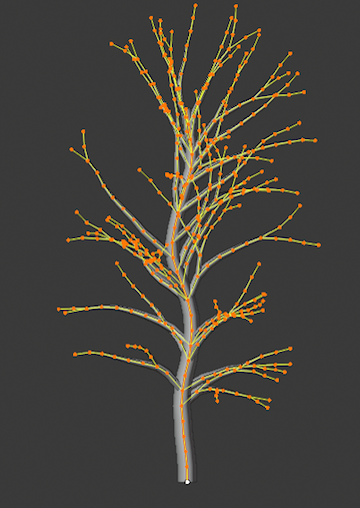}
	}
	\subfloat[] {
		\includegraphics[width=3.5cm,height=4.5cm]{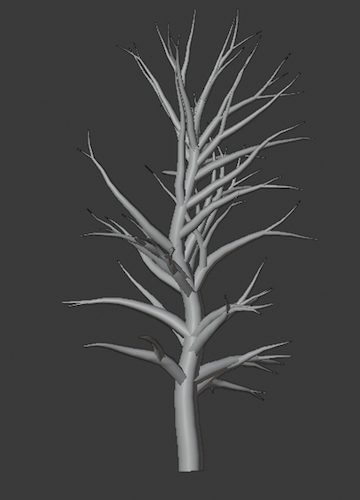}
	}
	\subfloat[] {
		\includegraphics[width=3.5cm,height=4.5cm]{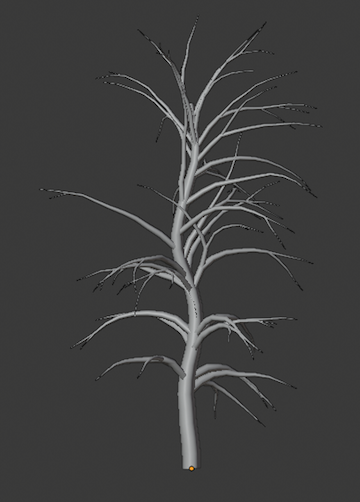}
	}
	\subfloat[] {
		\includegraphics[width=3.5cm,height=4.5cm]{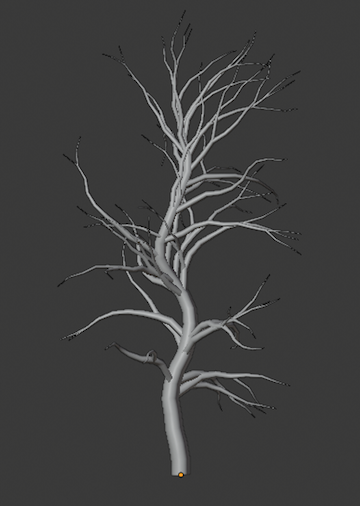}
	}
	
	\vfill
	\caption{Some operations to create a tree model in the tree editor. \textbf{a} is a mesh model generated from a tree skeleton. \textbf{b} is the effect of increasing the trunk and branch radius. \textbf{c} is gravity effect. \textbf{d} is the effect of branch bending}
	\label{tree_editor}
\end{figure}

In this editor, we perform the following basic operations to control the shape of tree are show in Fig \ref{tree_editor}: 

\textbf{Draw curves}: The skeleton of trunk is drawn. And the length and bending of the trunk are adjusted according to parameters. Then the curve can be drawn at any point on the trunk to split the trunk. 

\textbf{Edit radius}: Control the mesh properties of the drawn cylindrical branches by modifying the radius parameter.

\textbf{Gravity effect}: In order to increase the reality of the trees, the gravity bending degree of the branches is adjusted by adding a negative z-axis gravity parameter vector to the original direction of the branches.

\textbf{Bending degree}: Representing the grid geometry as a set of rigid prisms connected by elastic joints, the bending effect is achieved by modifying the positional constraints on a subset of these prisms. When the shaft deforms, the other branches maintain relative positions and orientations.
	
\subsection{Implicit surface modeling}
In order to sample uniformly distributed point cloud, the laser-based virtual scanner require scan surface with smooth normal field. However, the normals of triangular mesh are not continuous and smooth between faces. We approximate an implicit surface from a triangle mesh. Since the implicit surface\cite{ohtake2003multi-level} can provide high-precision shape approximations and accurately reconstruct sharp and small features, it is used to reconstruct surfaces from a large number of points and polygons\cite{berger2013a}. We create a smooth implicit surface by integrating the weight functions on the polygons\cite{shen2005interpolating}.

Given a point $\mathbf{x} \in \mathbb{R}^{3}$, $\mathbf{t}$ is any point in the triangular $T$, $\mathbf{t} \in {T}$ . To approximate the triangular surface, the weight function is defined as:

\begin{equation}
w(\mathbf{x}, \mathbf{t})=\frac{1}{\left(|\mathbf{x}-\mathbf{t}|^{2}+\epsilon^{2}\right)^{2}}
\end{equation}

$\epsilon$ is the smoothing factor. With the value of $\epsilon$ is from small to large, the implicit surface is from interpolation to extreme smoothness, for appropriate value, can completely fit the model and reconstruct small features and sharp edges.

The shape function is fitted to the triangular mesh for MPU processing. We adaptively construct an octree in a triangular mesh for sphere filling. The radius of the sphere is the length of the octree diagonal. Then, the shape function is fitted to the triangle contained within the sphere. In each cell $i$, for all triangles contained in the sphere,$T_{i} \in {T}$, $n_{t}$ is the triangle normal of $T_{i}$ and $c_{i}$ is the center of the cell $i$, the shape function is defined as :

\begin{equation}
{\mathbf{s}_{i}(\mathbf{x})=\mathbf{x}^{\mathrm{T}}\frac{\sum_{T_{i} \in T} \mathbf{n}_{t} \int_{\mathbf{t} \in T_{i}} w\left(\mathbf{c}_{i}, \mathbf{t}\right) \mathrm{d} \mathbf{t}}{\sum_{T_{i} \in T} \int_{\mathbf{t} \in T_{i}} w\left(\mathbf{c}_{i}, \mathbf{t}\right) \mathrm{d} \mathbf{t}}}-\left\langle\frac{\sum_{T_{i} \in T} \int_{\mathbf{t} \in T_{i}} \mathbf{t} w\left(\mathbf{c}_{i}, \mathbf{t}\right) \mathrm{d} \mathbf{t}}{\sum_{T_{i} \in T} \int_{\mathbf{t} \in T_{i}} w\left(\mathbf{c}_{i}, \mathbf{t}\right) \mathrm{d} \mathbf{t}}, \mathbf{n}_{i}\right\rangle
\end{equation}

We cover the space with spheres. If the sphere is empty, the radius of the sphere is expanded until it contains a sufficient number of triangles to provide fit. 

The implicit function is estimated by blending the shape functions of all spheres containing the point, $q_{i}$ is a quadratic b-spline function centered at $s_{i}$:

\begin{equation}
f(\mathbf{x})=\frac{\sum q_{i}(\mathbf{x}) s_{i}(\mathbf{x})}{\sum q_{i}(\mathbf{x})}
\end{equation}

\subsection{Sampling}
The point cloud module is the result of laser bars projecting onto implicit surfaces. We simulate the LiDAR (Light Detection and Ranging) scanning process to perform distance scanning to obtain point cloud data with normals and directions. 

\begin{figure}
	\centering
	\subfloat[] {
		\includegraphics[width=0.15\textwidth]{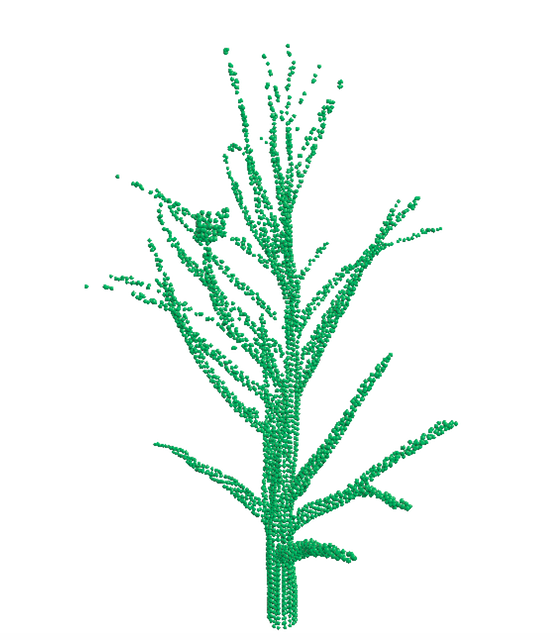}
	}
	\subfloat[] {
		\includegraphics[width=0.15\textwidth]{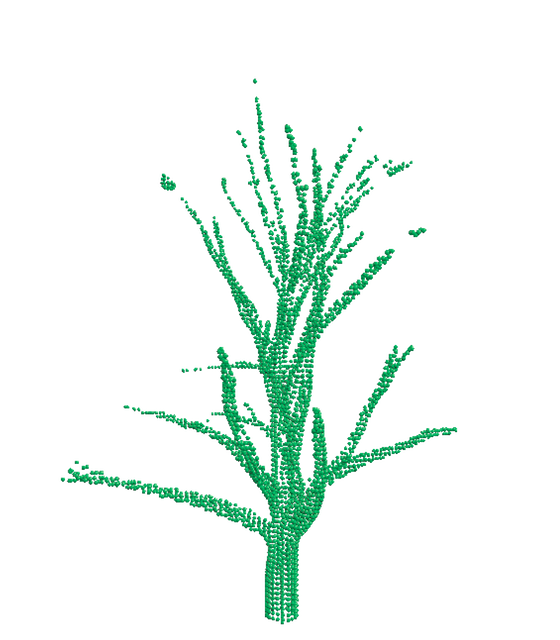}
	}
	\subfloat[] {
		\includegraphics[width=0.15\textwidth]{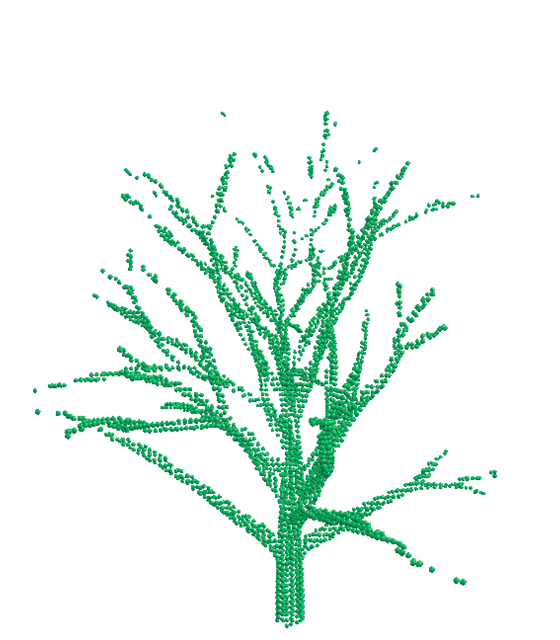}
	}
	\subfloat[] {
		\includegraphics[width=0.15\textwidth]{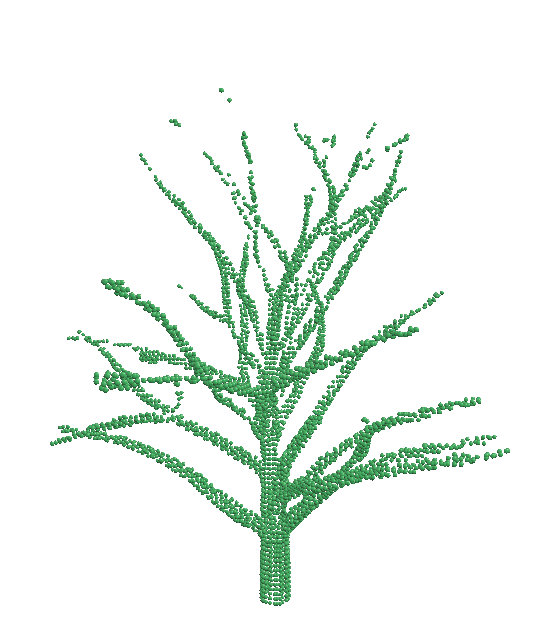}
	}
	\vfill
	\subfloat[] {
		\includegraphics[width=0.15\textwidth]{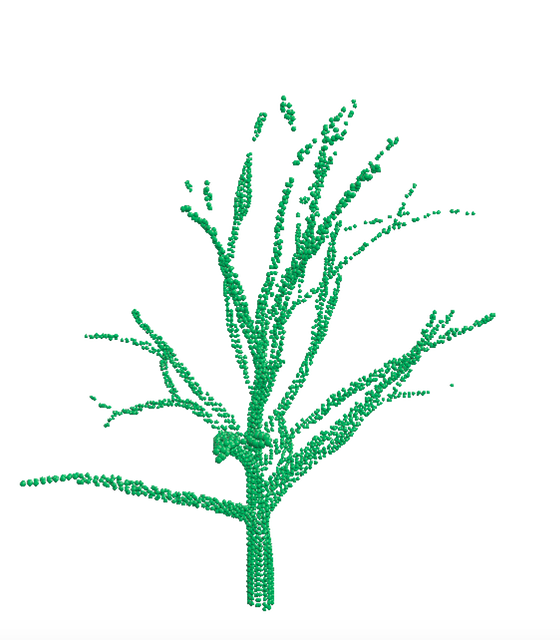}
	}
	\subfloat[] {
		\includegraphics[width=0.15\textwidth]{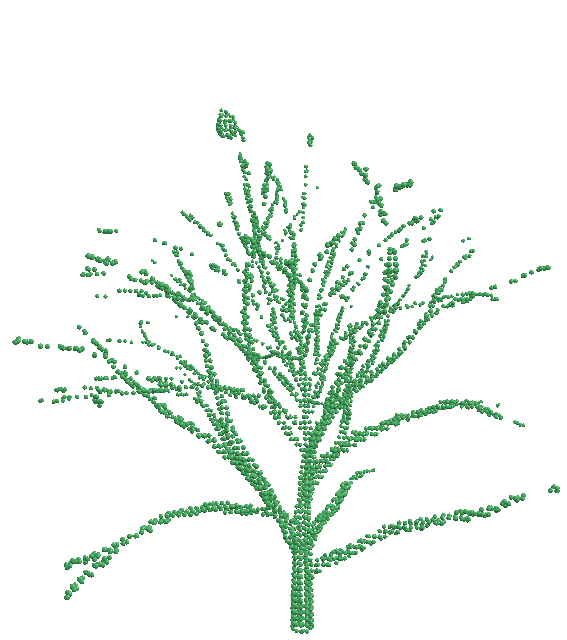}
	}
	\subfloat[] {
		\includegraphics[width=0.15\textwidth]{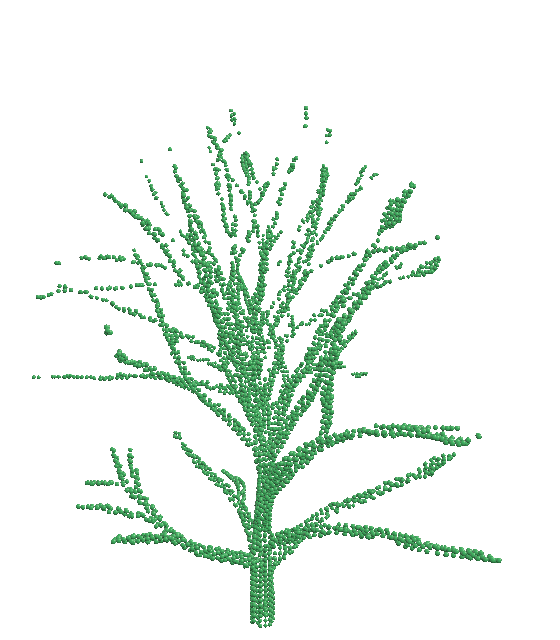}
	}
	\subfloat[] {
		\includegraphics[width=0.15\textwidth]{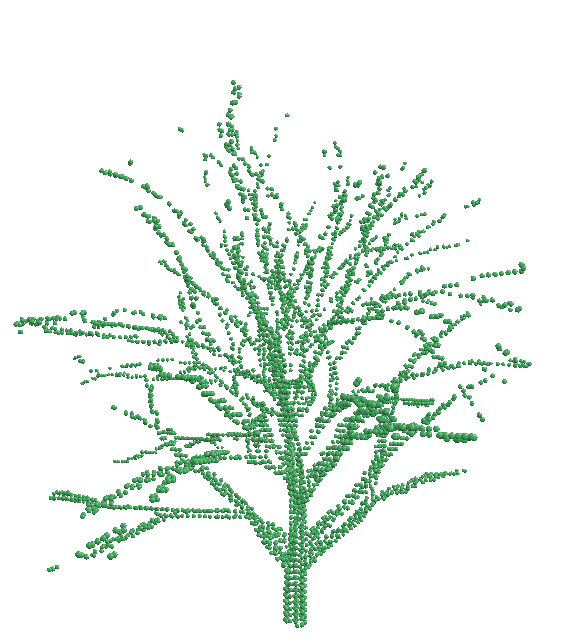}
	}
	\vfill
	\subfloat[] {
		\includegraphics[width=0.15\textwidth]{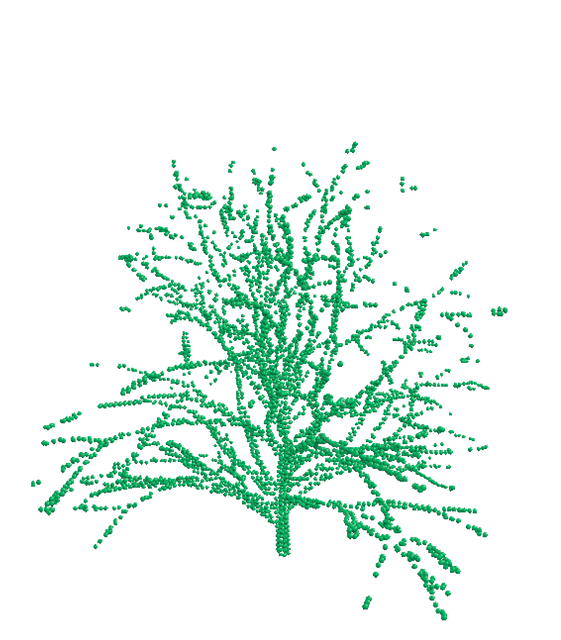}
	}
	\subfloat[] {
		\includegraphics[width=0.15\textwidth]{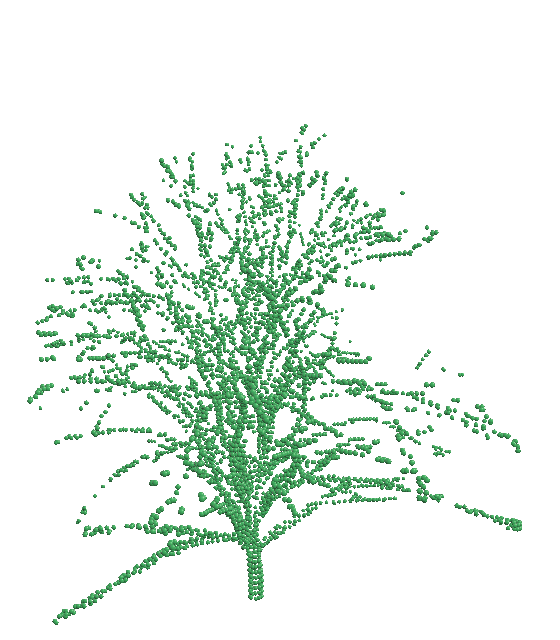}
	}
	\subfloat[] {
		\includegraphics[width=0.15\textwidth]{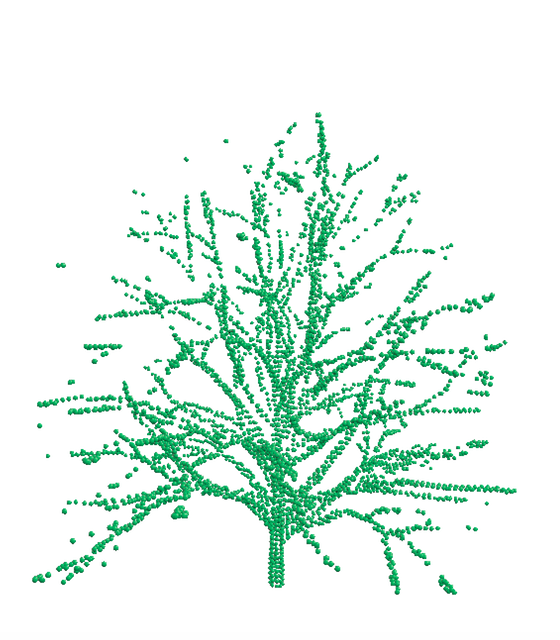}
	}
	\subfloat[] {
		\includegraphics[width=0.15\textwidth]{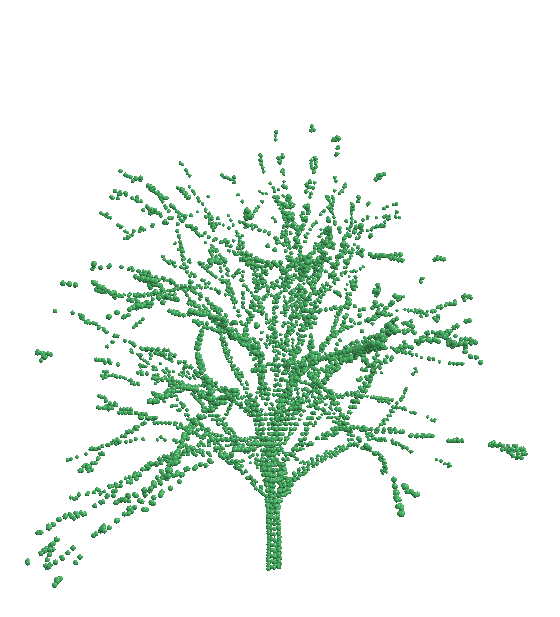}
	}
	\vfill
	\caption{ (\textbf{a}) to (\textbf{i}) are some point cloud models scanned by virtual scanner.}
	\label{scan_data}
\end{figure}

Firstly determining the scan location and register the scan. We place the scanner on a uniformly spaced position on the object's bounding sphere so that the camera's orientation is aligned with the object's centroid\cite{vlasic2009dynamic}. A synthetic range scan is then performed. The exact range data is first generated by ray tracing the implicit surfaces. A single laser stripe is then projected onto the range geometry to collect all of the range geometry contained in the stripe.

With these separate range scans, we next perform point cloud registration to register them in a single coordinate system. The range scans are slightly overlapped and the registration algorithm\cite{brown2007global} is used to align the scans. Finally, we assign a normal to each point cloud. The PCA method is used to estimate the local tangent plane to estimate the normals, and the minimum spanning tree method\cite{hoppe1992surface} is used to calculate the normal directions. The scaned point cloud models are shown in Fig \ref{scan_data}.

\subsection{Build different types of point cloud models}

\begin{figure}[htb]
	\centering
	\subfloat[Sparse density] {
		\includegraphics[width=0.3\textwidth]{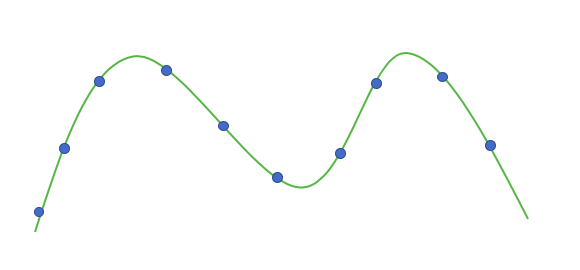}
	}
	\subfloat[Noisy data] {
		\includegraphics[width=0.3\textwidth]{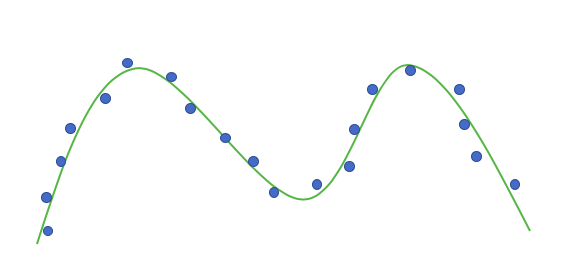}
	}
	\vfill
	\subfloat[Missing data] {
		\includegraphics[width=0.3\textwidth]{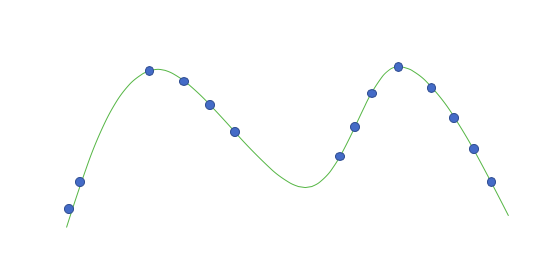}
	}
	\subfloat[Uneven sampling] {
		\includegraphics[width=0.3\textwidth]{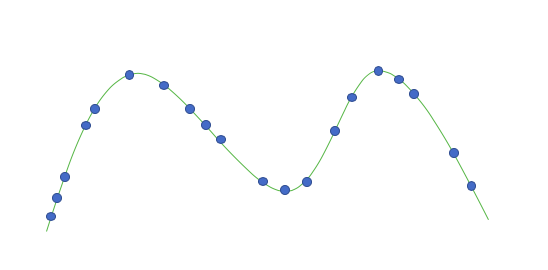}
	}
	
	\vfill
	\caption{Common properties of scanned data on a sampled curve. The green curve is real curve, and the blue points are the points sampled from the curve. }
	\label{attribute}
\end{figure}

We build four different point cloud models to simulate common attributes found in scanned point clouds, as shown in Fig \ref{attribute}. Different methods are used to construct four types of point cloud datasets: point cloud datasets with different density of points, noisy point cloud datasets, point cloud datasets with missing data, and point cloud datasets with uneven density distribution. See the appendix \ref{supply} for more details.

\begin{figure}[htb]
	\centering
	\subfloat[resolution=50] {
		\includegraphics[width=0.15\textwidth]{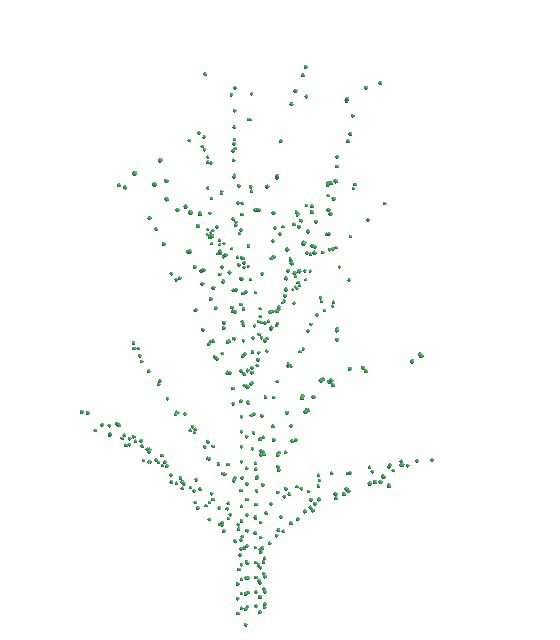}
	}
	\subfloat[resolution=100] {
		\includegraphics[width=0.15\textwidth]{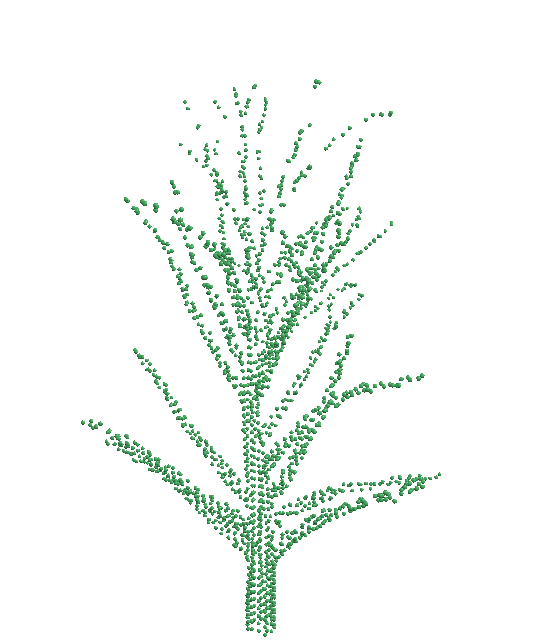}
	}
	\subfloat[resolution=150] {
		\includegraphics[width=0.15\textwidth]{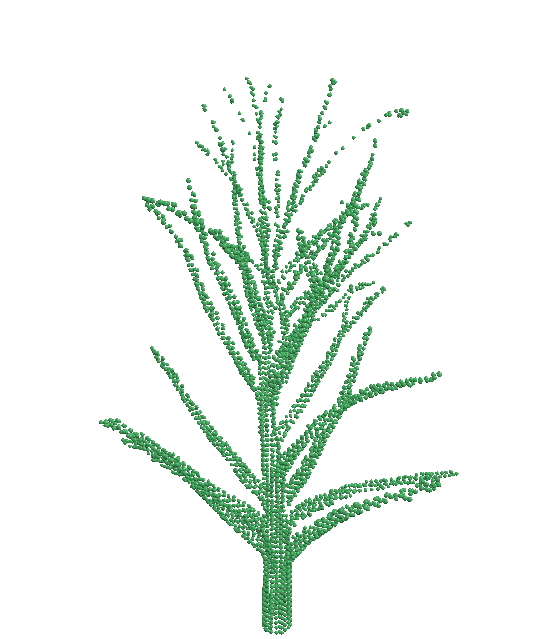}
	}
	\subfloat[resolution=50] {
		\includegraphics[width=0.15\textwidth]{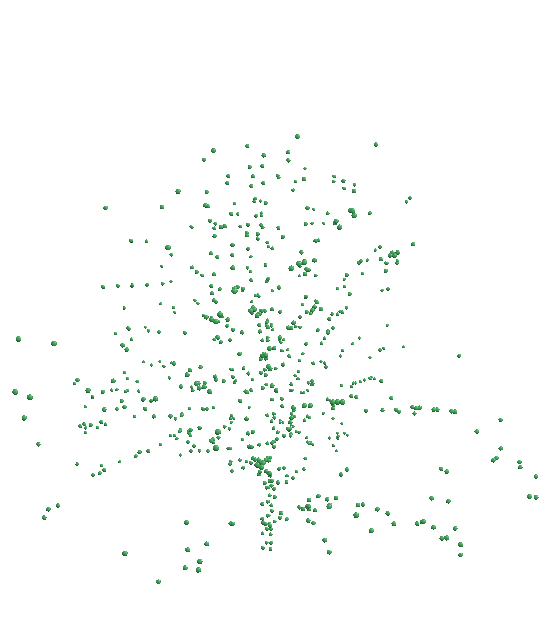}
	}
	\subfloat[resolution=100] {
		\includegraphics[width=0.15\textwidth]{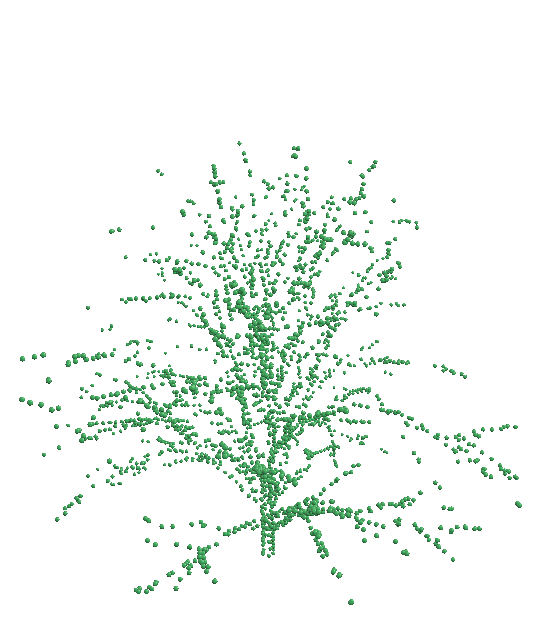}
	}
	\subfloat[resolution=150] {
		\includegraphics[width=0.15\textwidth]{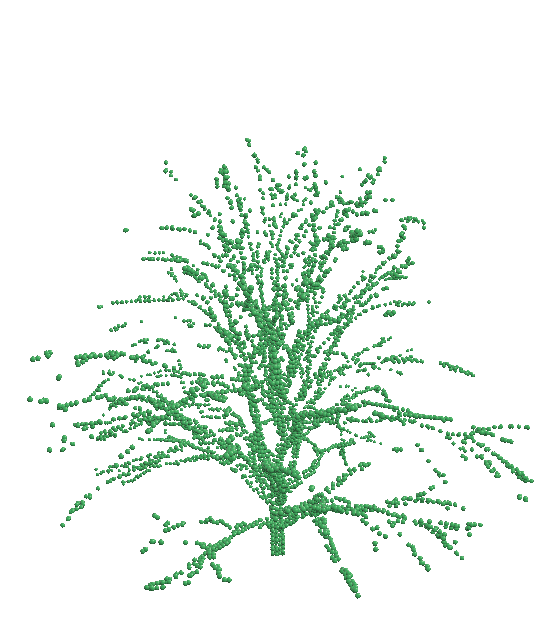}
	}
	\caption{ Some examples of different density point cloud models. In (\textbf{a}) to (\textbf{c}) and (\textbf{d}) to (\textbf{f}), the density of points increases with resolution.}
	\label{diferent_density}
\end{figure}

The density of the scanned point clouds is controlled by adjusting the resolution of the virtual scanner. When scanning, We set the resolution at three scales: 50, 100, 150. The higher the resolution, the denser the point clouds. Some cases are shown in Fig \ref{diferent_density}.

\begin{figure}[htb]
	\centering
	\subfloat[] {
		\includegraphics[width=0.15\textwidth]{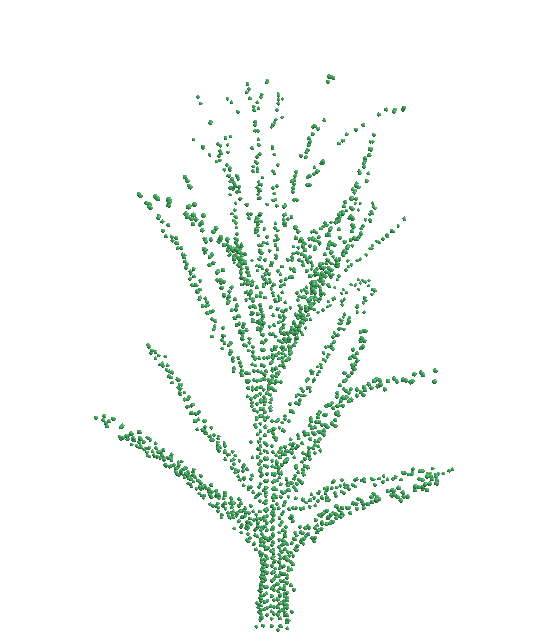}
	}
	\subfloat[] {
		\includegraphics[width=0.15\textwidth]{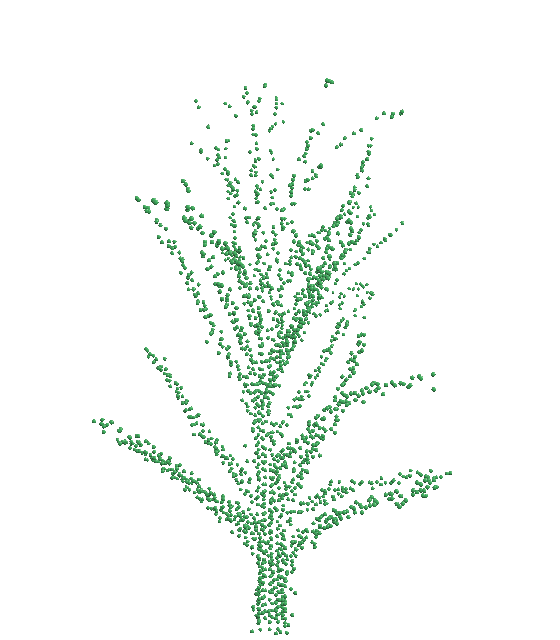}
	}
	\subfloat[] {
		\includegraphics[width=0.15\textwidth]{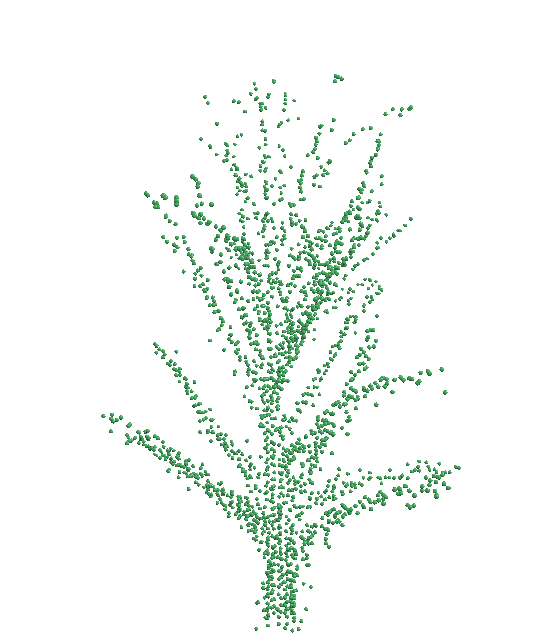}
	}
	\subfloat[] {
		\includegraphics[width=0.15\textwidth]{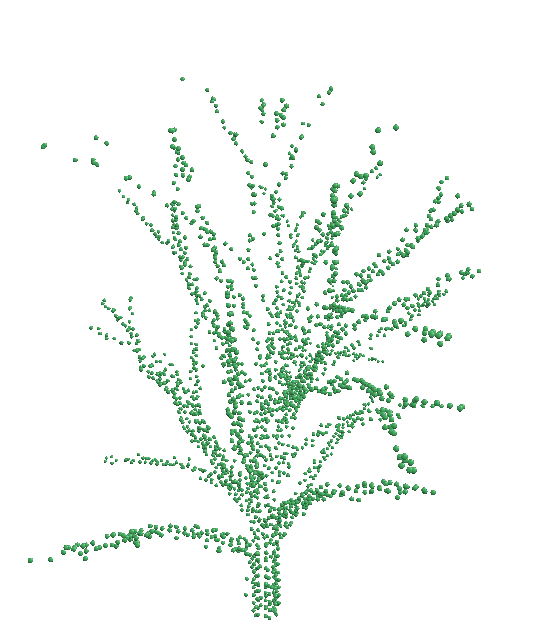}
	}
	\subfloat[] {
		\includegraphics[width=0.15\textwidth]{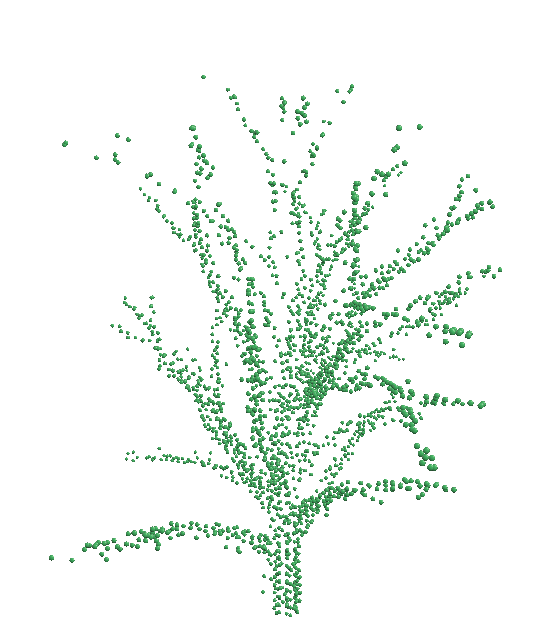}
	}
	\subfloat[] {
		\includegraphics[width=0.15\textwidth]{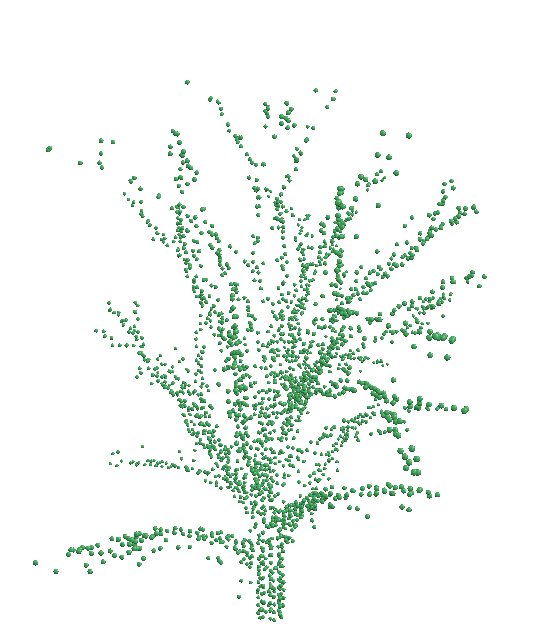}
	}
	\caption{ Some examples of point cloud models with noise.}
	\label{noise}
\end{figure}

A certain number of points are selected from the original point clouds, and point cloud models with noise are constructed by adding Gaussian noise points in the range of thier normal direction are shown in Fig \ref{noise}. The pseudocode of this method is presented in Algorithm.\ref{alg:one}. $s$ represents the range of noise points inserted in the normal direction. Noise point density $d$ means insert a noise point every $d$ points.

\begin{algorithm}[H]  
	\caption{Point cloud with noise}  
	\setstretch{1.5}
	\label{alg:one}  
	\begin{algorithmic}[1]  
		\Require  
		Tree point cloud $Q=\{q_i\}_{i \in I} \subset {\mathbf R}^3$;
		Normal vector reduction factor $s$; Noise point density $d$;

		\For{ $i \in I$}
		\State Calculate  point $q_i$ normal $n_i$;
		\State Generate a random number $G_i$ that obeys a Gaussian distribution;
		\State New point $p_{i}$ =$q_{i}$ + $G_{i}$$n_{i}$$s$;
		\State Insert $p_i$ to $Q$;
		\State $i$ = $i$ + $d$;
		\EndFor 

		\\
		\Return $Q$;  
	\end{algorithmic}  
\end{algorithm}  

\begin{figure}[htb]
	\centering
	\subfloat[n=2,r=0.05] {
		\includegraphics[width=0.15\textwidth]{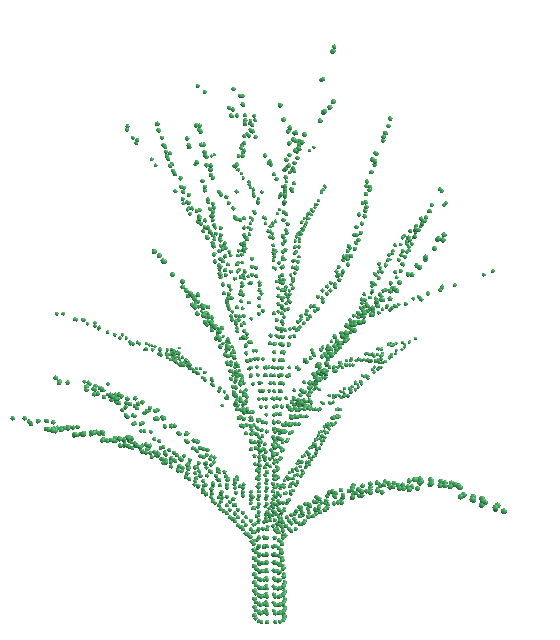}
	}
	\subfloat[n=3,r=0.09] {
		\includegraphics[width=0.15\textwidth]{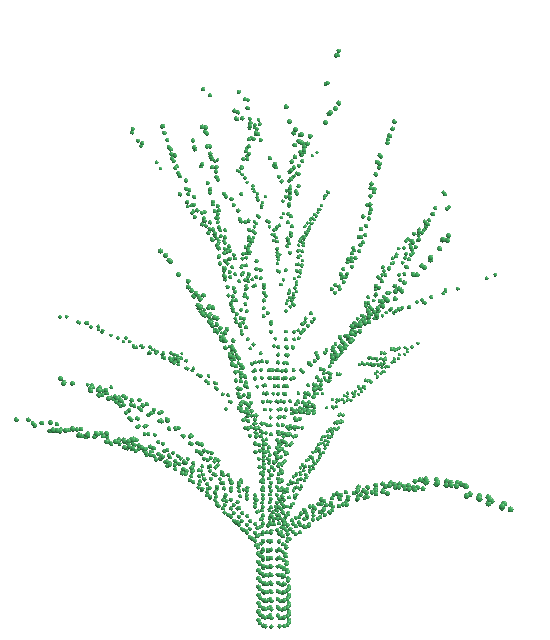}
	}
	\subfloat[n=2,r=0.05] {
		\includegraphics[width=0.15\textwidth]{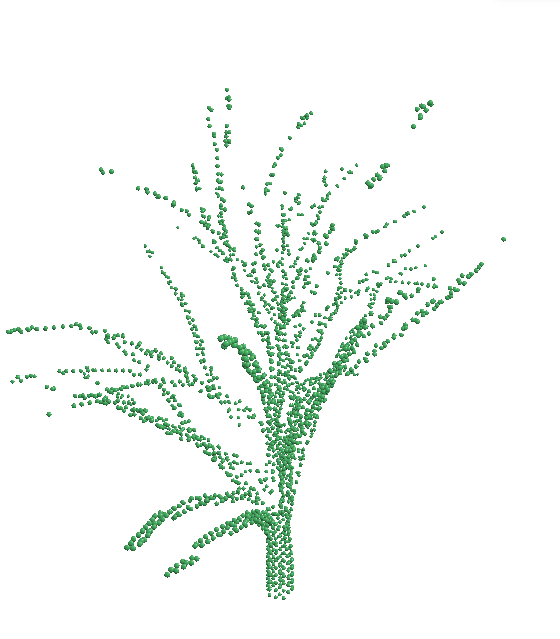}
	}
	\subfloat[n=3,r=0.09] {
		\includegraphics[width=0.15\textwidth]{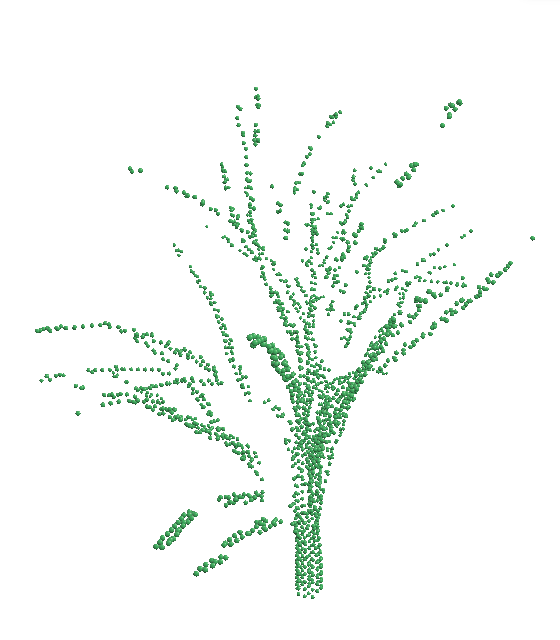}
	}
	\subfloat[n=2,r=0.05] {
		\includegraphics[width=0.15\textwidth]{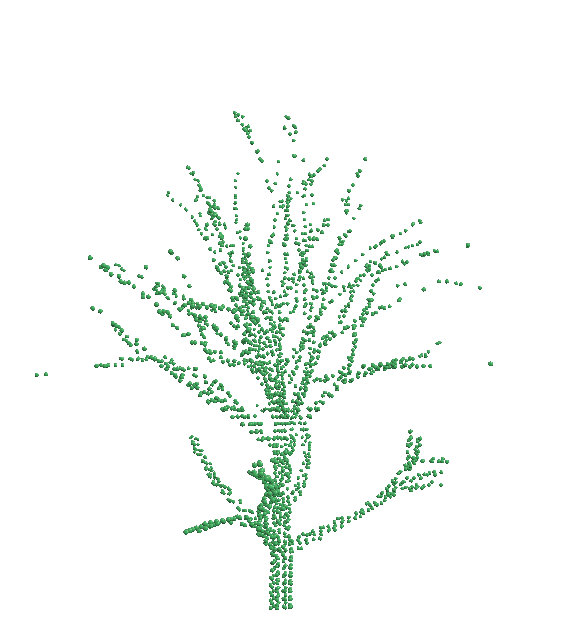}
	}
	\subfloat[n=3,r=0.09] {
		\includegraphics[width=0.15\textwidth]{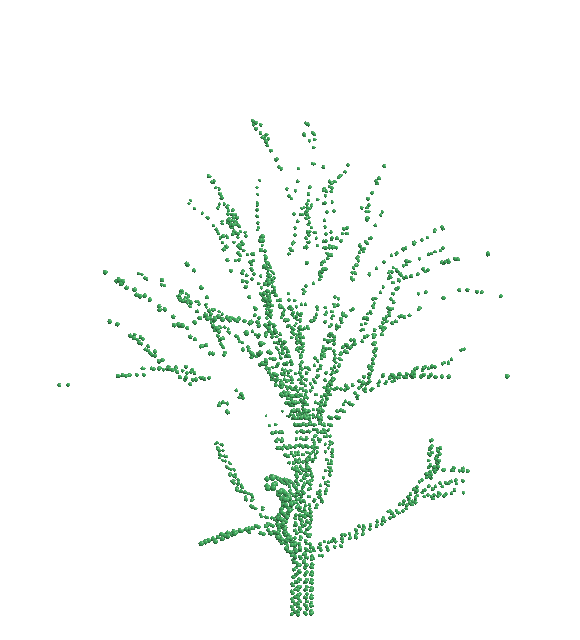}
	}
	
	\vfill
	\caption{ Some examples of point cloud models with missing data. $n$ is the number of occlusion ball, $r$ is the radius of the ball.}
	\label{missing}
\end{figure}

Point cloud models with missing data are constructed by building occlusion balls and removing points within the balls from the original point cloud. The center of the ball is a random point in the original point cloud, the radius of the ball is $\lambda$ times of  the norm of the boundingbox extent vector $l$. The number of balls is $N$. Some example are shown in Fig \ref{missing}. The pseudocode is presented in Algorithm.\ref{alg:two}.
\begin{algorithm}[H]  
	\caption{Point cloud with missing data}  
	\setstretch{1.5}
	\label{alg:two}  
	\begin{algorithmic}[1]  
		\Require  
		Tree point cloud $Q=\{q_i\}_{i \in I} \subset {\mathbf R}^3$;
		 Number of occlusion balls $N$; The norm of the boundingbox extent vector $l$;
		 Random ratio $\lambda$;
		 
		 \For{ $n \in N$}
		 \State $ball_n$.center=random point  $q_n$;
		 \State $ball_n$.radius=$l$$\lambda$;
		 \State Add $ball_n$ to  occlusion balls;
		 \EndFor 
		 
		\For{ $i \in I$}
		\If {$q_i$ in occlusion balls }
		\State Remove point $q_i$ form $Q$;
		\EndIf
		\EndFor 
		
		\Return $Q$;  
	\end{algorithmic}  
\end{algorithm}

\begin{figure}[htb]
	\centering
	\vfill
	\subfloat[] {
		\includegraphics[width=0.15\textwidth]{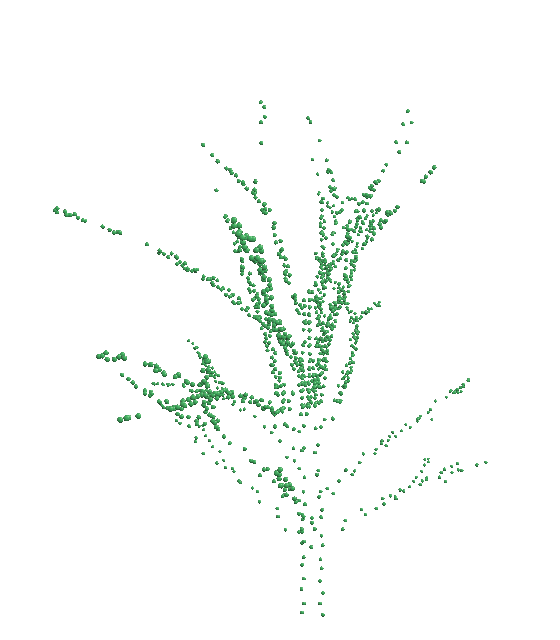}
	}
	\subfloat[] {
		\includegraphics[width=0.15\textwidth]{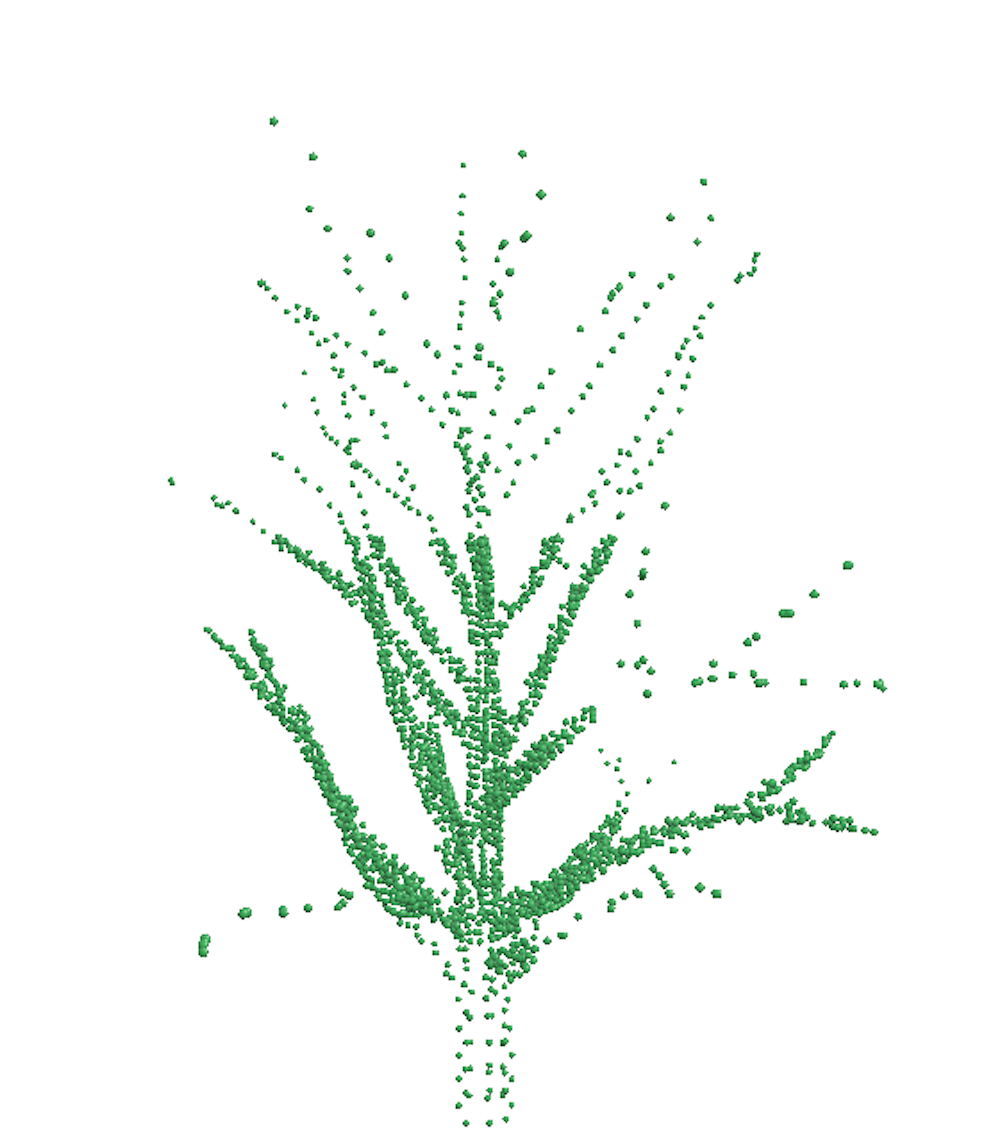}
	}
	\subfloat[]{
		\includegraphics[width=0.15\textwidth]{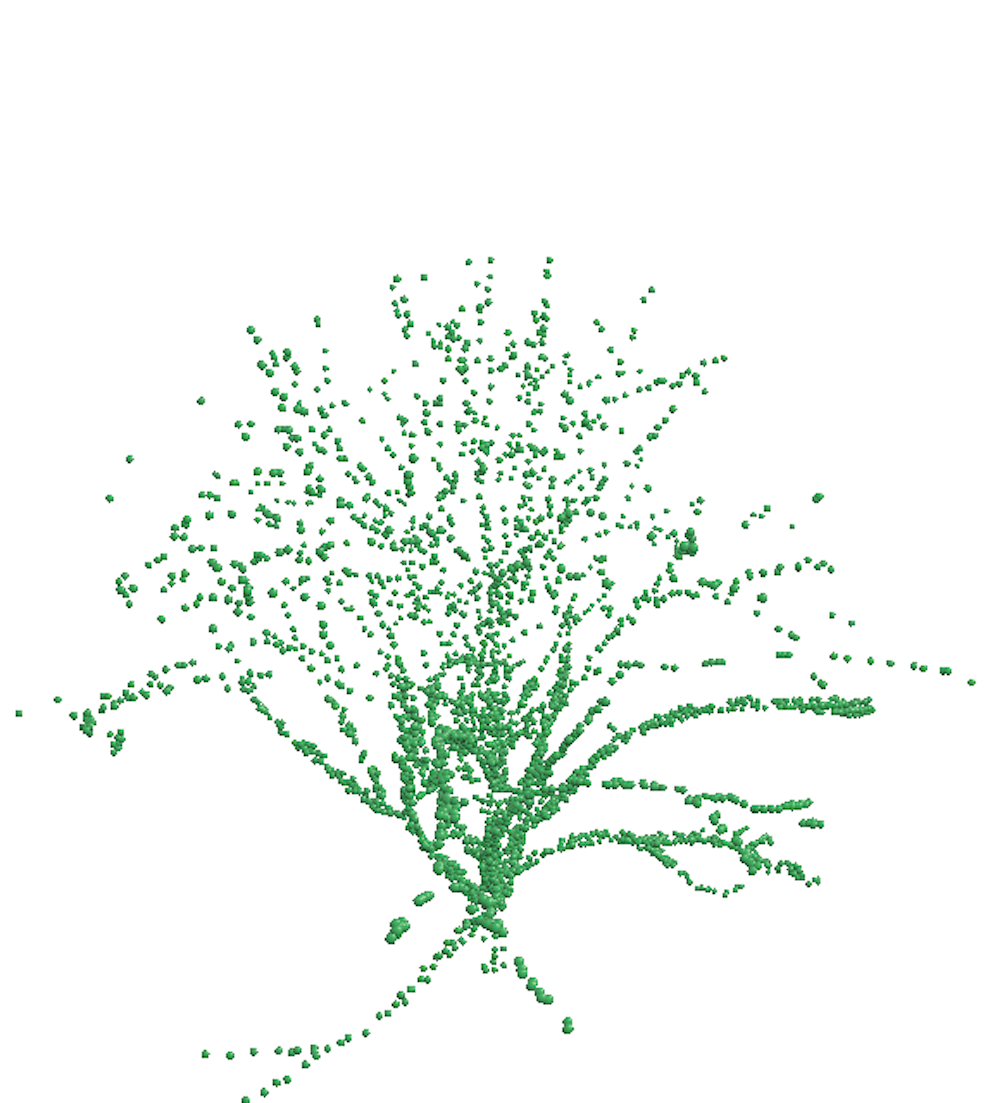}
	}
	\subfloat[] {
		\includegraphics[width=0.15\textwidth]{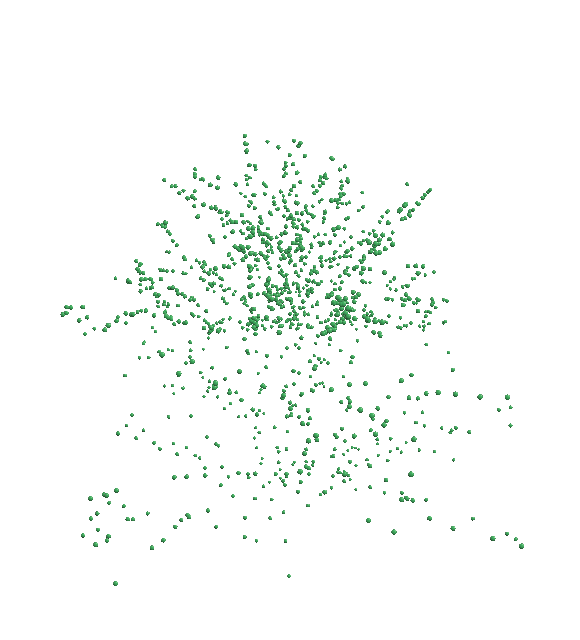}
	}
	\subfloat[] {
		\includegraphics[width=0.15\textwidth]{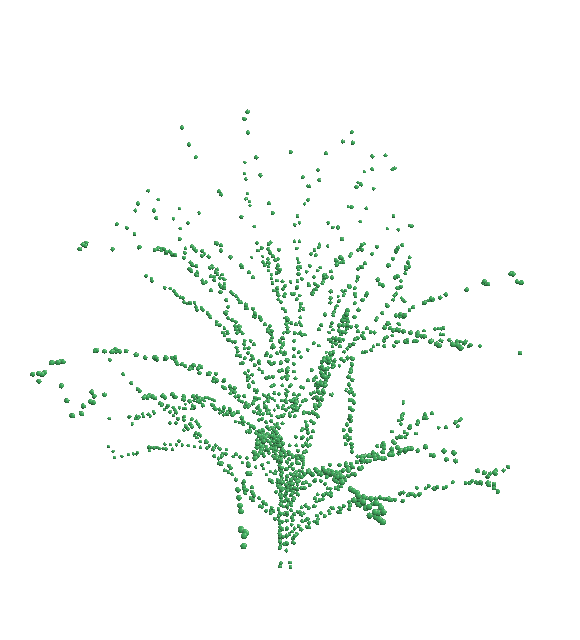}
	}
	\subfloat[] {
		\includegraphics[width=0.15\textwidth]{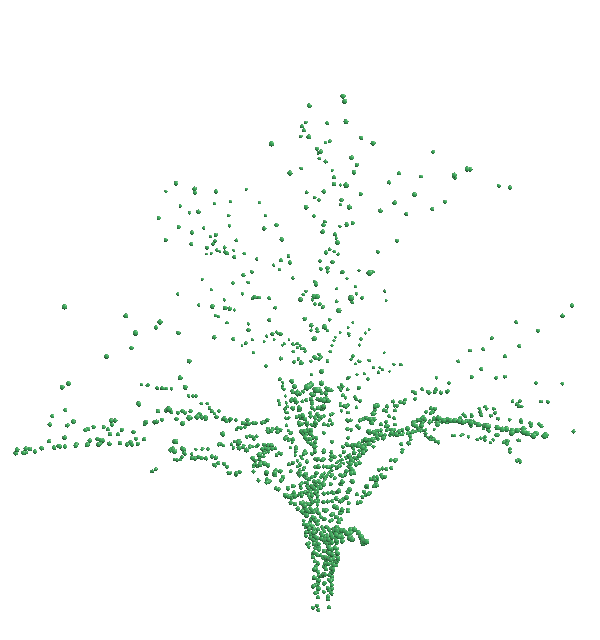}
	}
	\caption{ Some examples of point cloud models with varying point density. $l_1$  and $l_2$  are the ranges of density variation.}
	\label{vary density}
\end{figure}

Point cloud dataset with uneven density distribution are constructed by inserting points within in local area of the point cloud model. Within the scope of boundingbox, we randomly choose a range $R$ of point density variation. In the points within neighborhood radius $r$, the PCA is performed to calculate the first and second principal component. And the added point is obtained by random multiple of the principal direction. Finally, the calculated point is  inserted into the density variation range. Some example are shown in Fig \ref{vary density}. The pseudocode is presented in Algorithm.\ref{alg:three}.

\begin{algorithm}[H]  
	\caption{Point cloud with uneven density distribution}  
	\setstretch{1.5}
	\label{alg:three}  
	\begin{algorithmic}[1]  
		\Require  
		Tree point cloud $Q=\{q_i\}_{i \in I} \subset {\mathbf R}^3$;
		Points $N_{r}\left\{q_{i}\right\}$ within neighborhood radius $r$;
		Range of point density variation $R$;
		Random constraint factor $\lambda_{1}$,$\lambda_{2}$;
	
		\For{ each $i \in I$}
		\If {$q_i$ in $R$}
		\State In $N_{r}\left\{q_{i}\right\}$, use PCA to calculate the first principal direction $P_D$ and the second principal direction $S_D$ ;
		\State New point $p_{i}$ =$q_{i}$ + $\lambda_{1}$$P_D$ + $\lambda_{2}$$S_D$;
		\State Insert $p_i$ to $Q$;
		\Else
		\State continue;
		\EndIf
		\EndFor 
		
		\\
		\Return $Q$;  
	\end{algorithmic}  
\end{algorithm}  

\section{Result and analysis}
To verify the effectiveness of our dataset, we use state-of-the-art $L_1$ method\cite{huang2013l} to extract skeleton from the tree point cloud. $L_1$ method uses a weighting function with a support radius to define the size of the local neighborhood, then gradually increases the support radius and produces a clean and well-connected skeleton. We download the software provided in the paper. In order to be suitable for tree structure, we reduce initial neighborhood size and slow the growth of neighborhood size of the algorithm to ensure the topological correctness of the tree skeleton.

\begin{figure}[htb]
	\centering
	\vfill
{
		\includegraphics[width=0.15\textwidth]{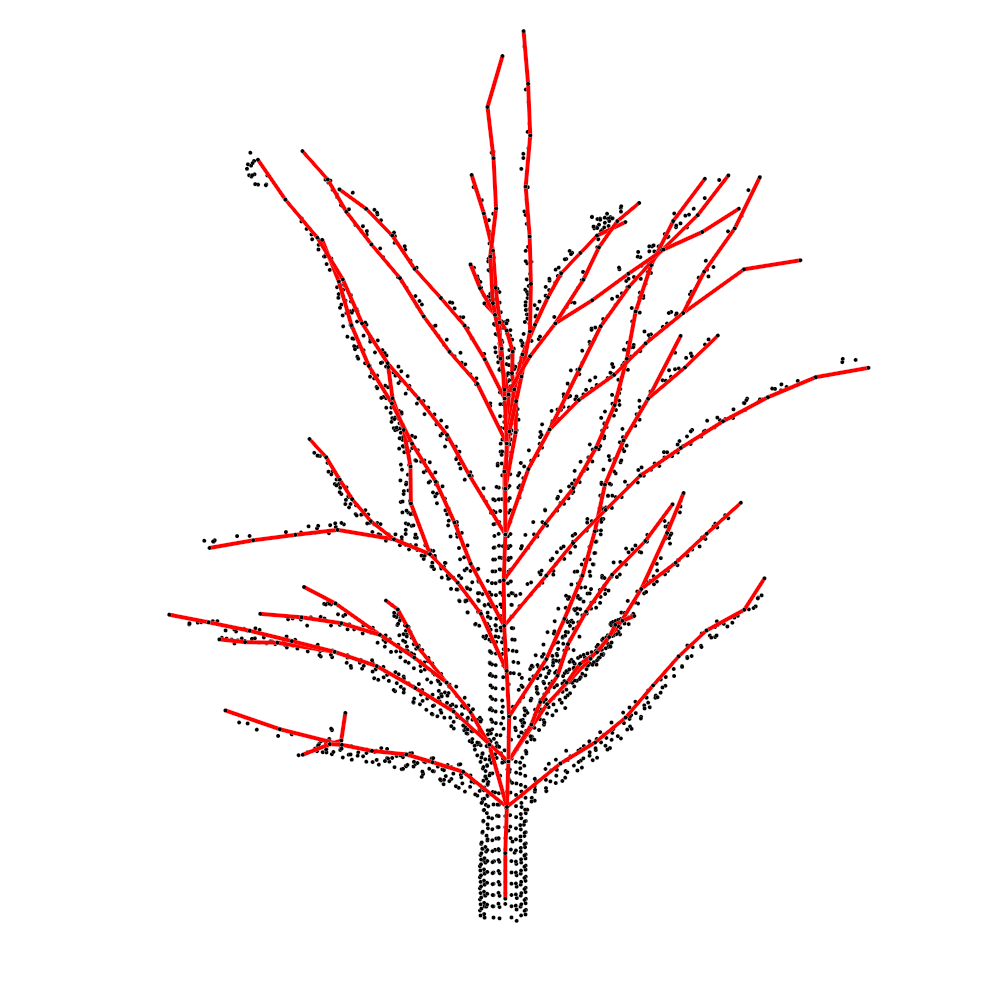}
	}
{
		\includegraphics[width=0.15\textwidth]{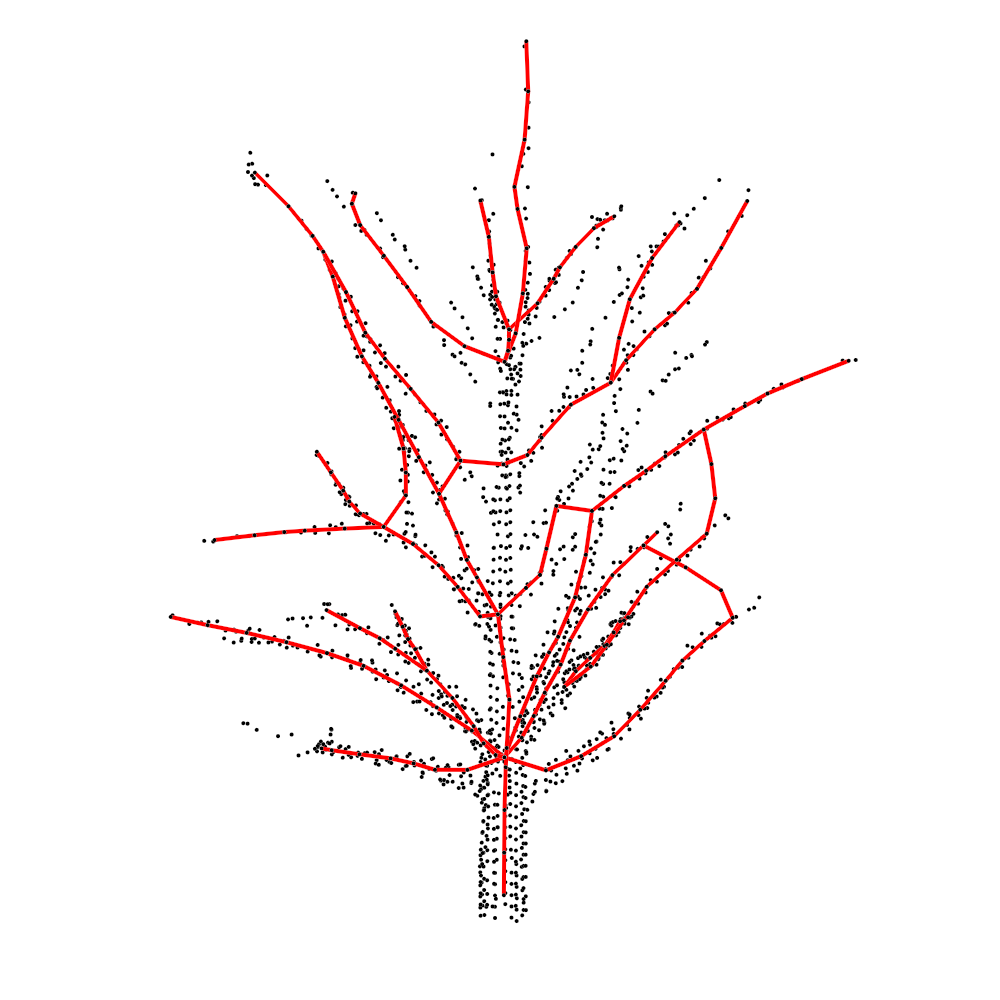}
	}
 {
			\includegraphics[width=0.15\textwidth]{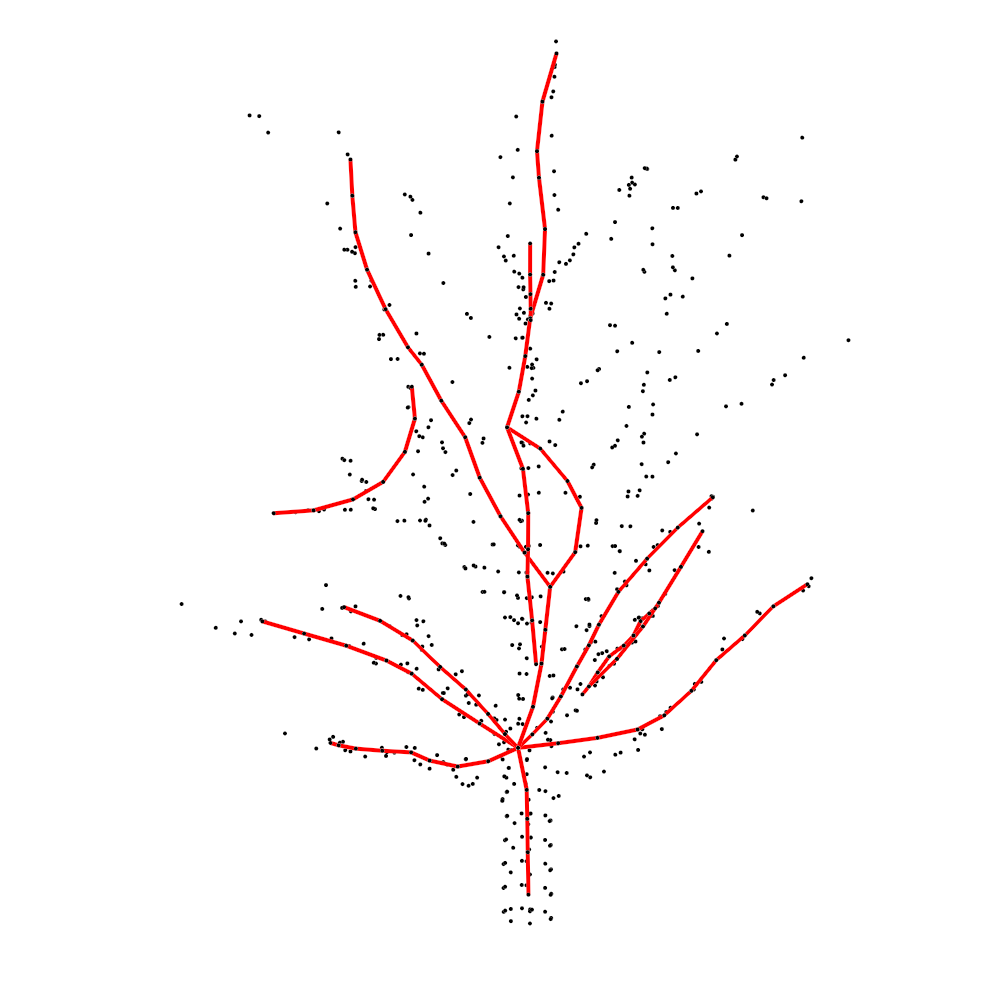}
		}
{
		\includegraphics[width=0.15\textwidth]{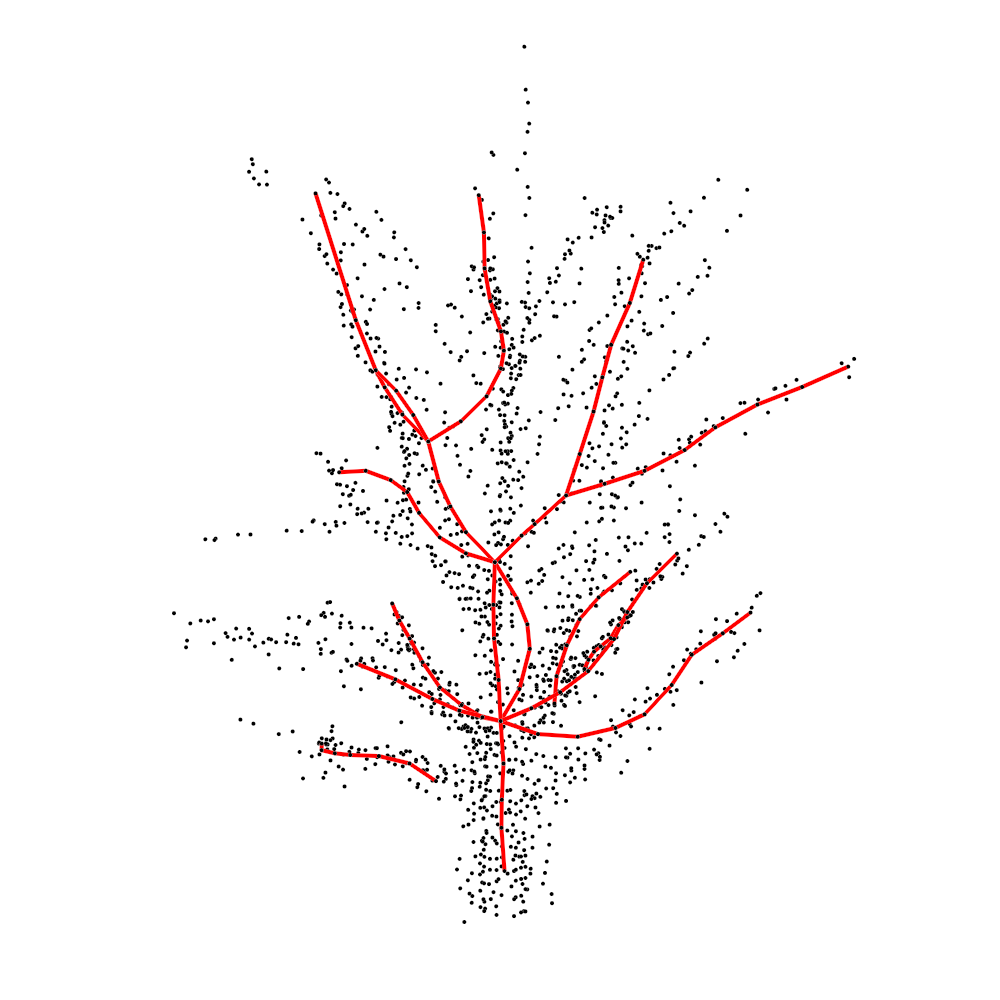}
	}
 {
		\includegraphics[width=0.15\textwidth]{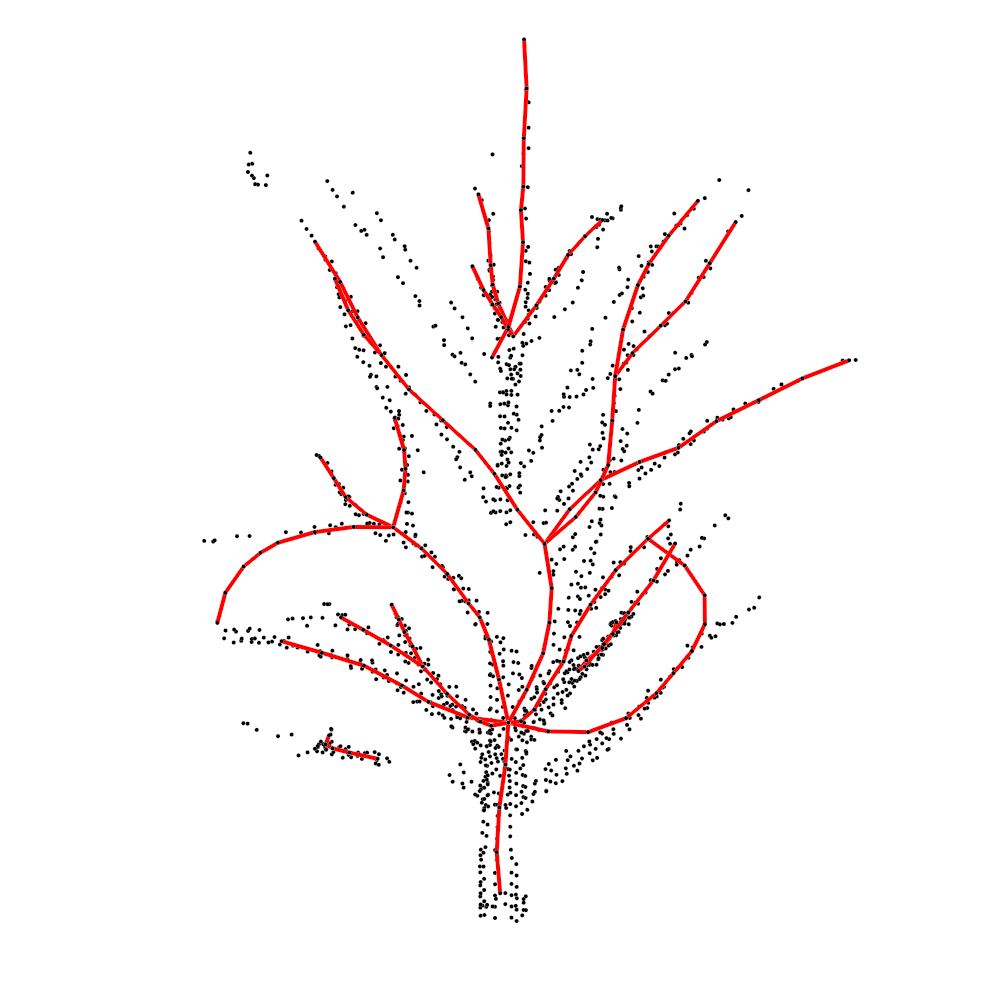}
	}
 {
		\includegraphics[width=0.15\textwidth]{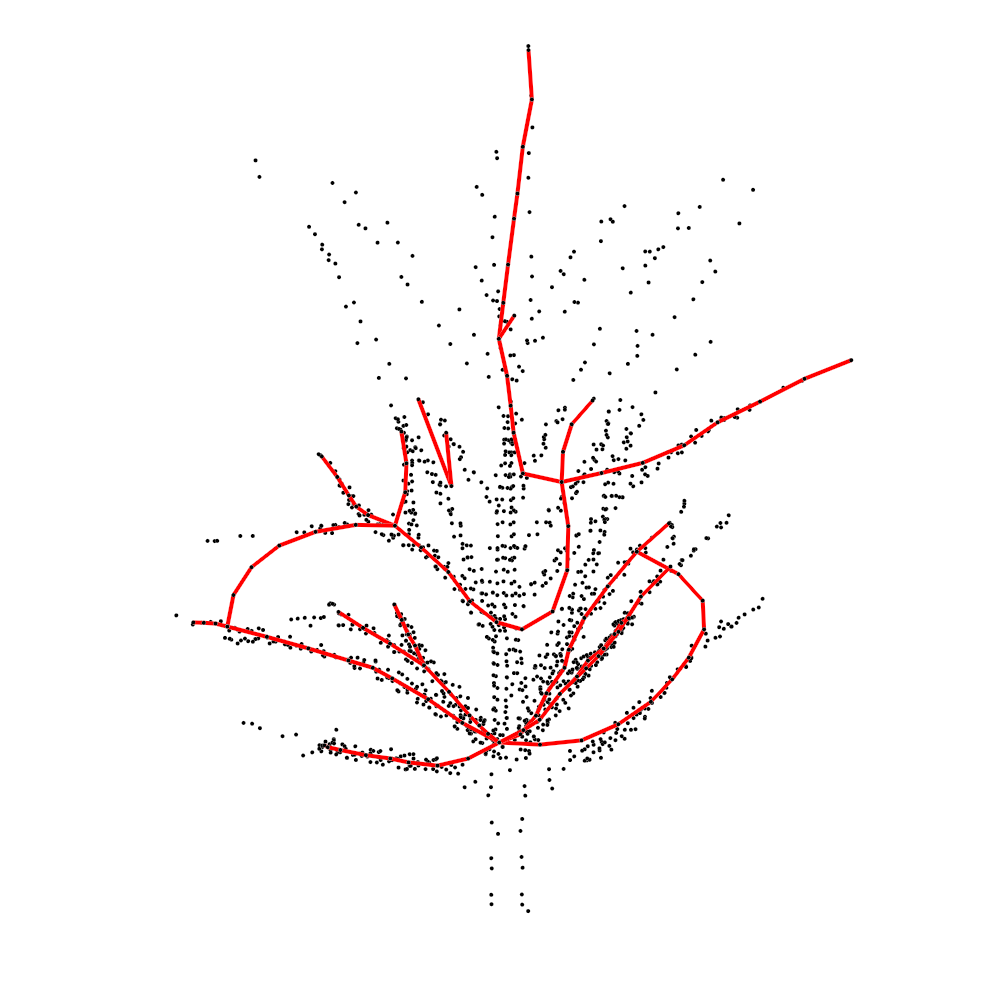}
	}
	
	{
		\includegraphics[width=0.15\textwidth]{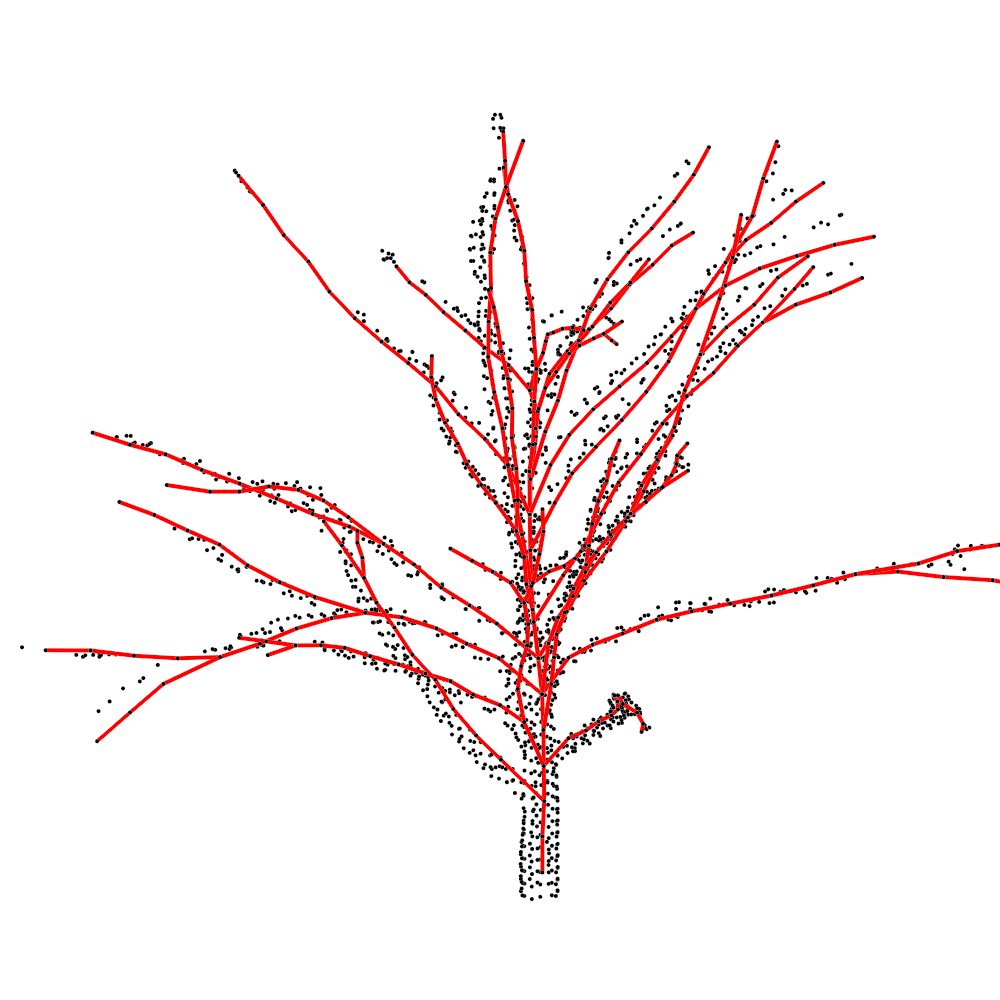}
	}
	{
		\includegraphics[width=0.15\textwidth]{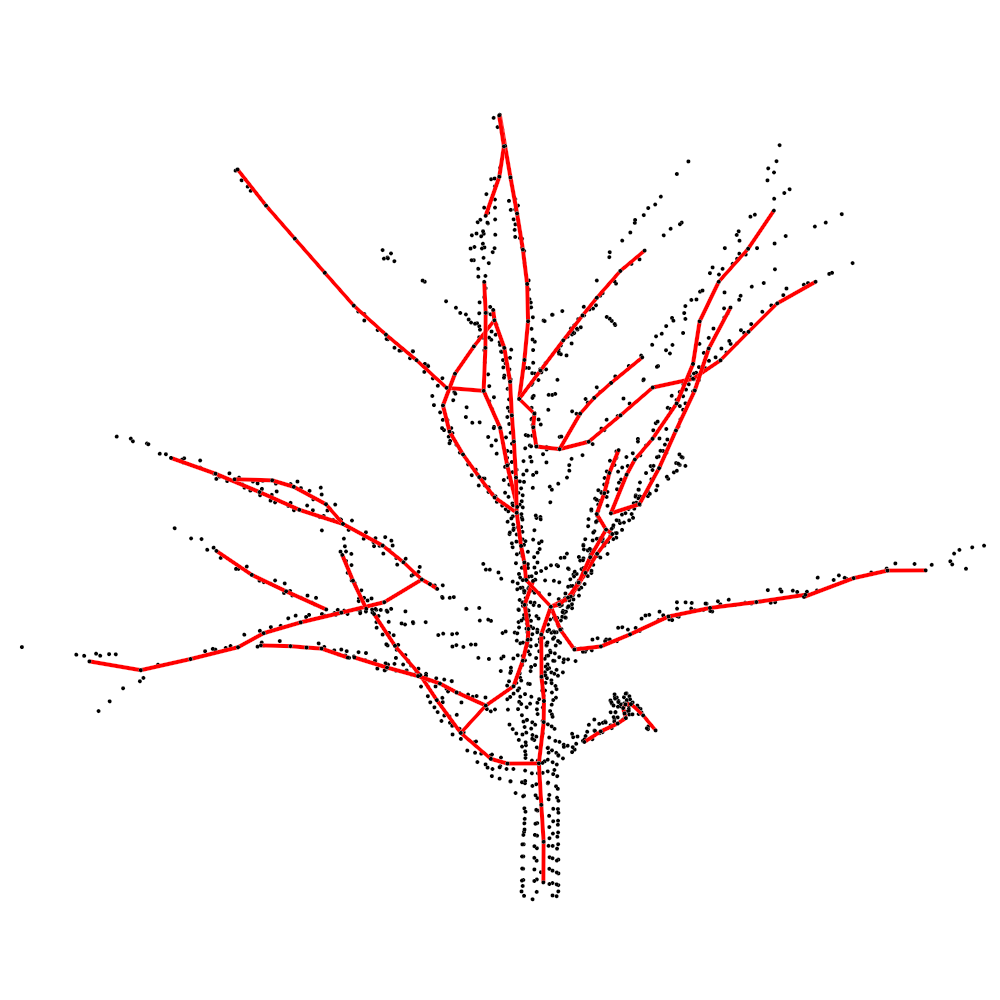}
	}
	{
		\includegraphics[width=0.15\textwidth]{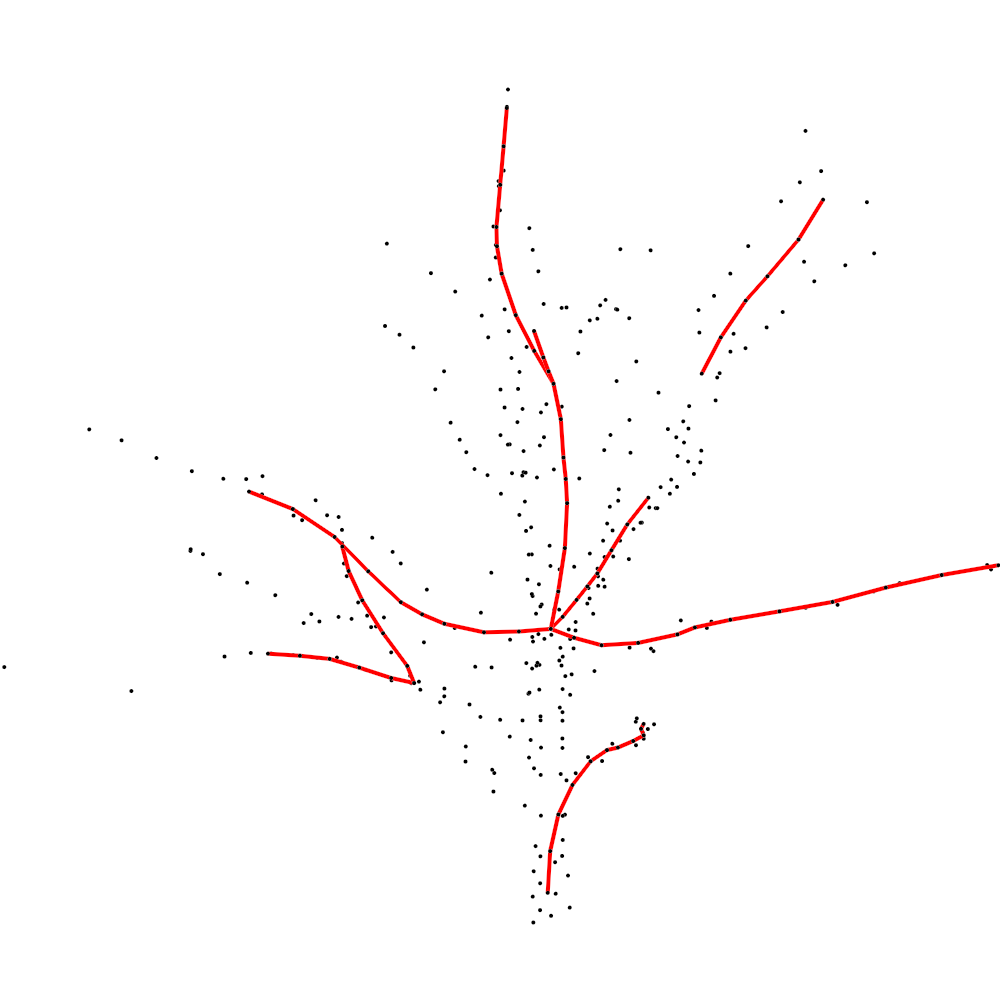}
	}
	{
		\includegraphics[width=0.15\textwidth]{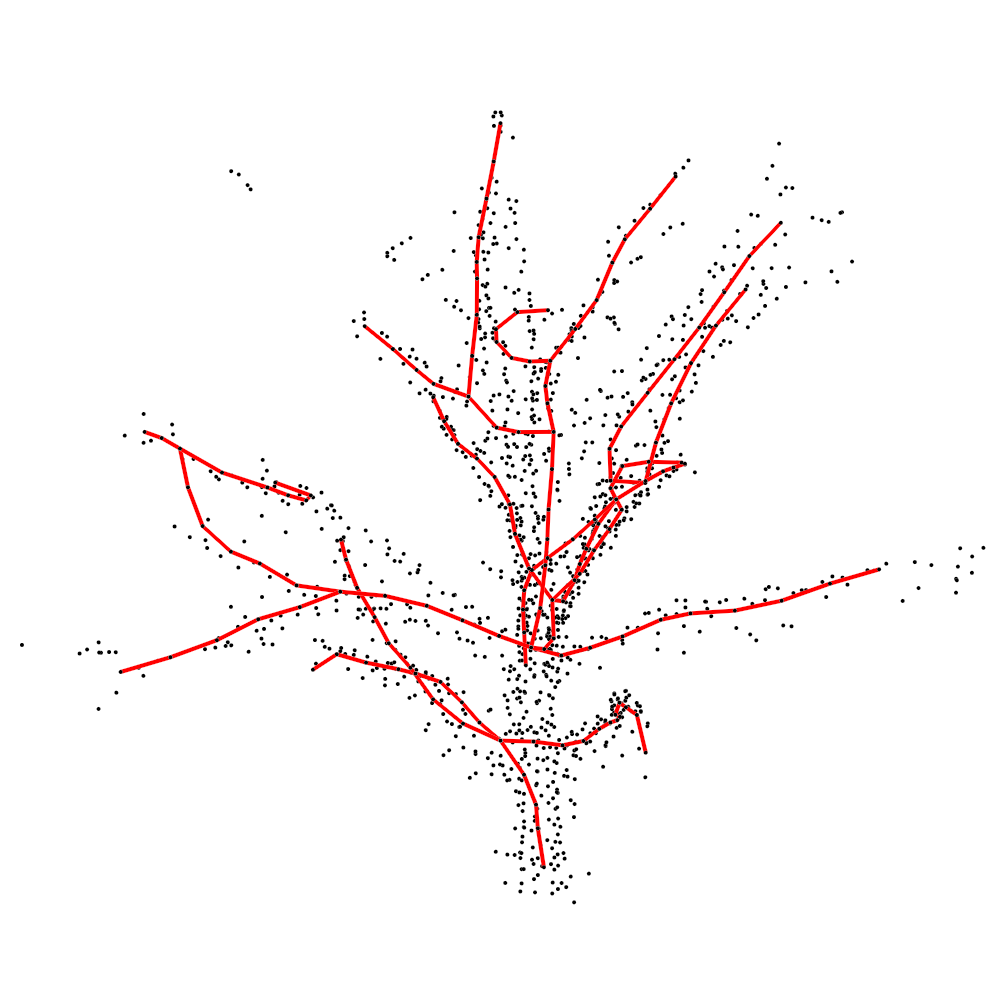}
	}
	{
		\includegraphics[width=0.15\textwidth]{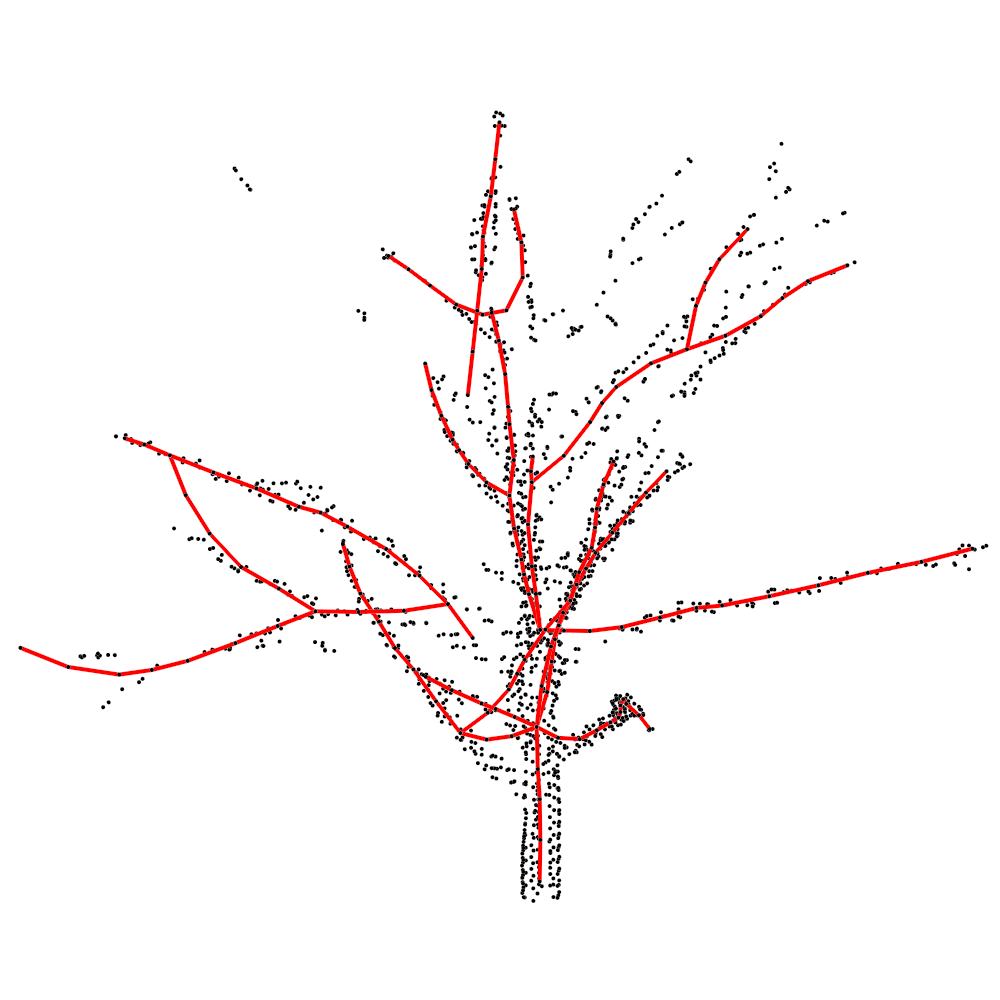}
	}
	{
		\includegraphics[width=0.15\textwidth]{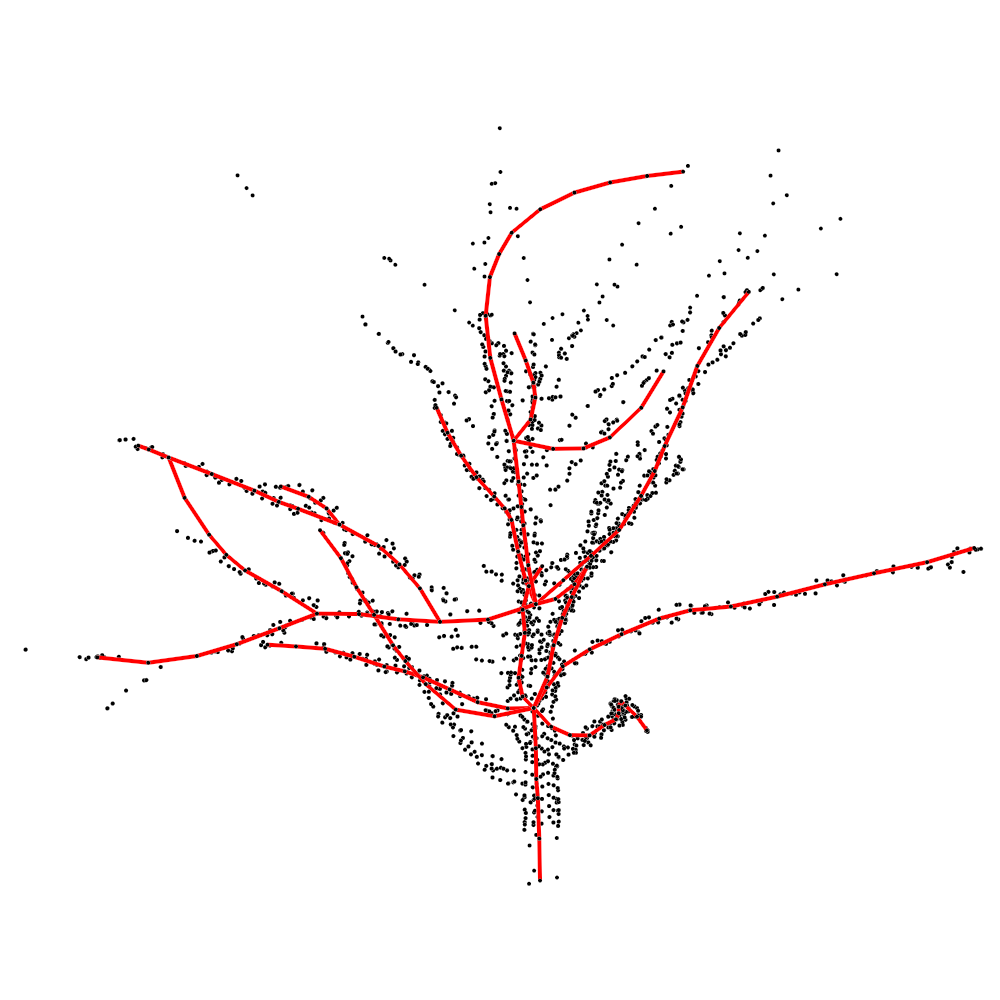}
	}

		{
		\includegraphics[width=0.15\textwidth]{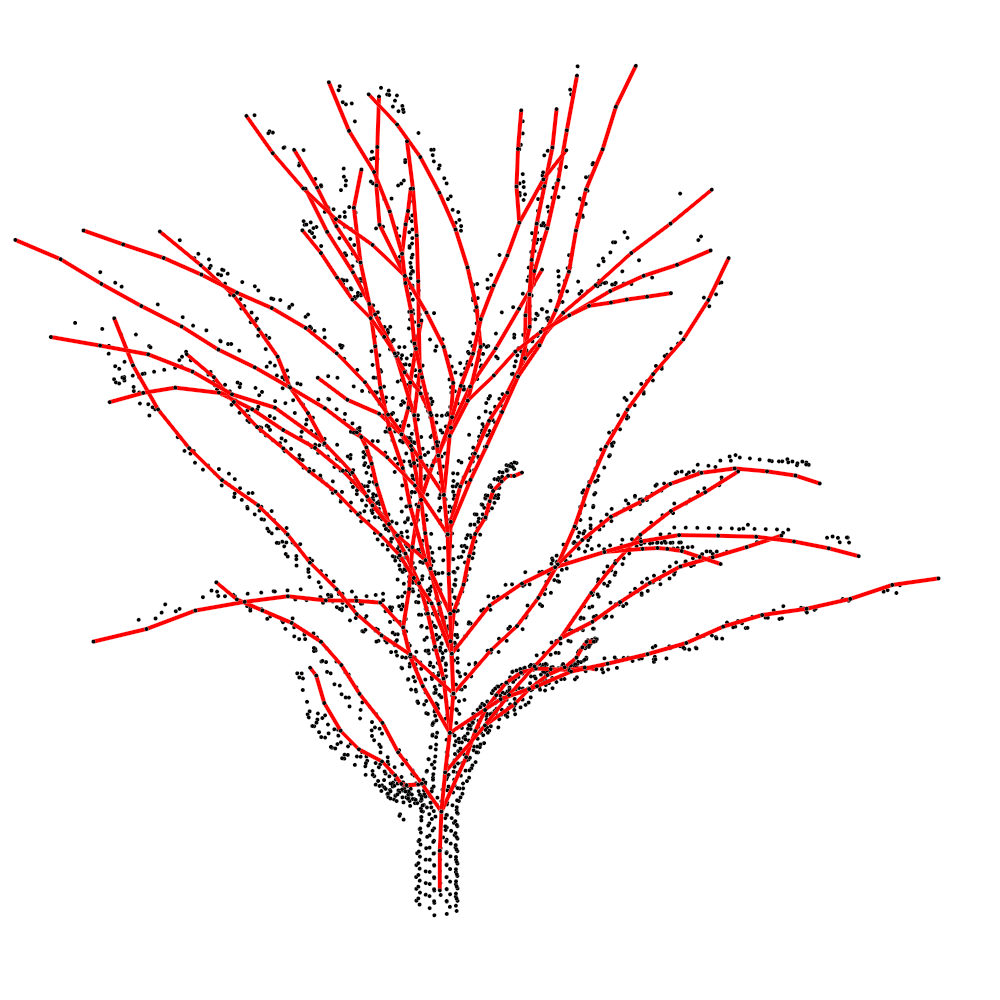}
	}
		{
		\includegraphics[width=0.15\textwidth]{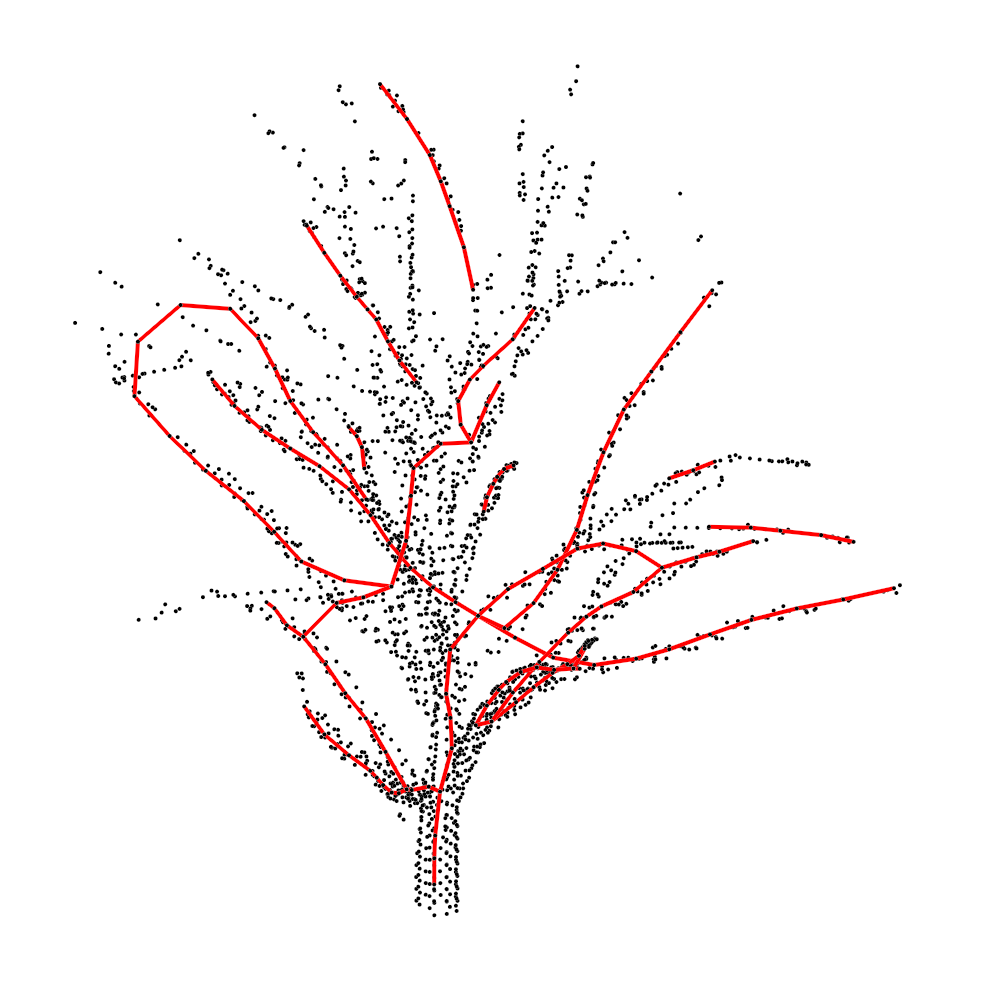}
	}
	{
		\includegraphics[width=0.15\textwidth]{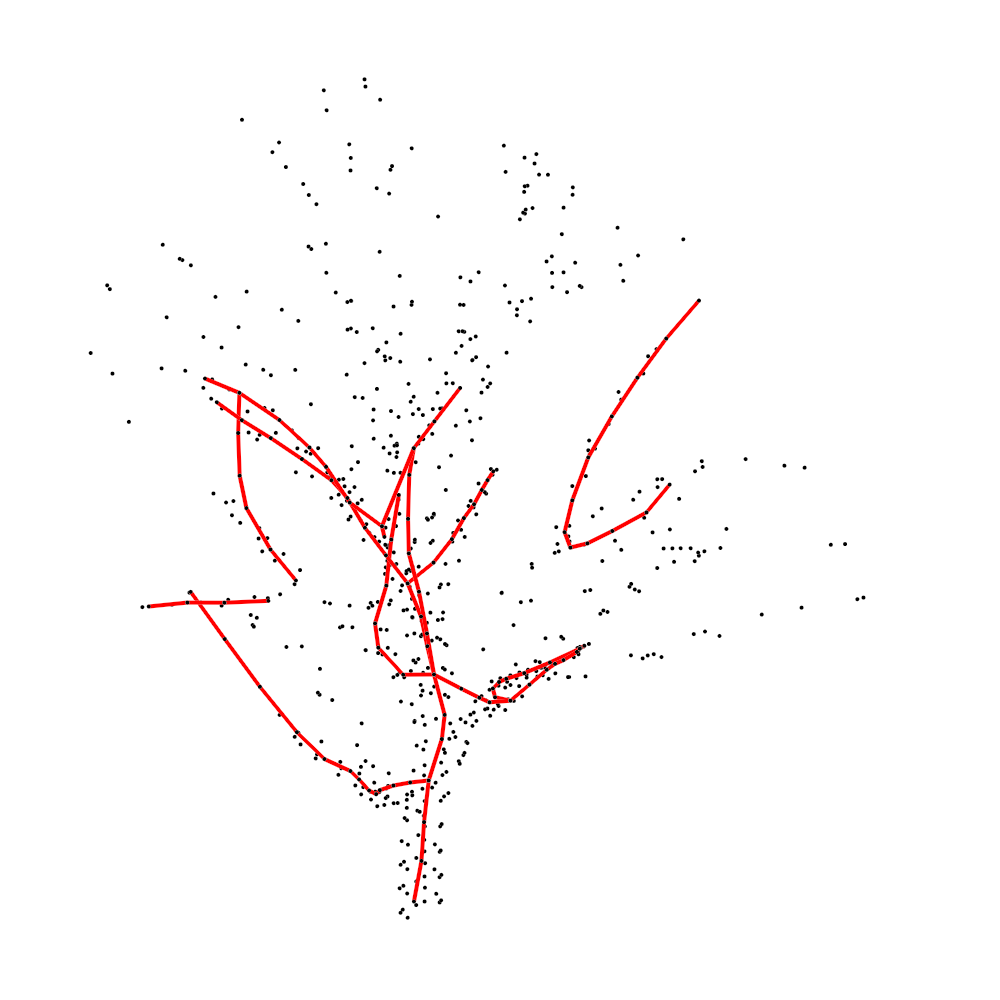}
	}
	{
		\includegraphics[width=0.15\textwidth]{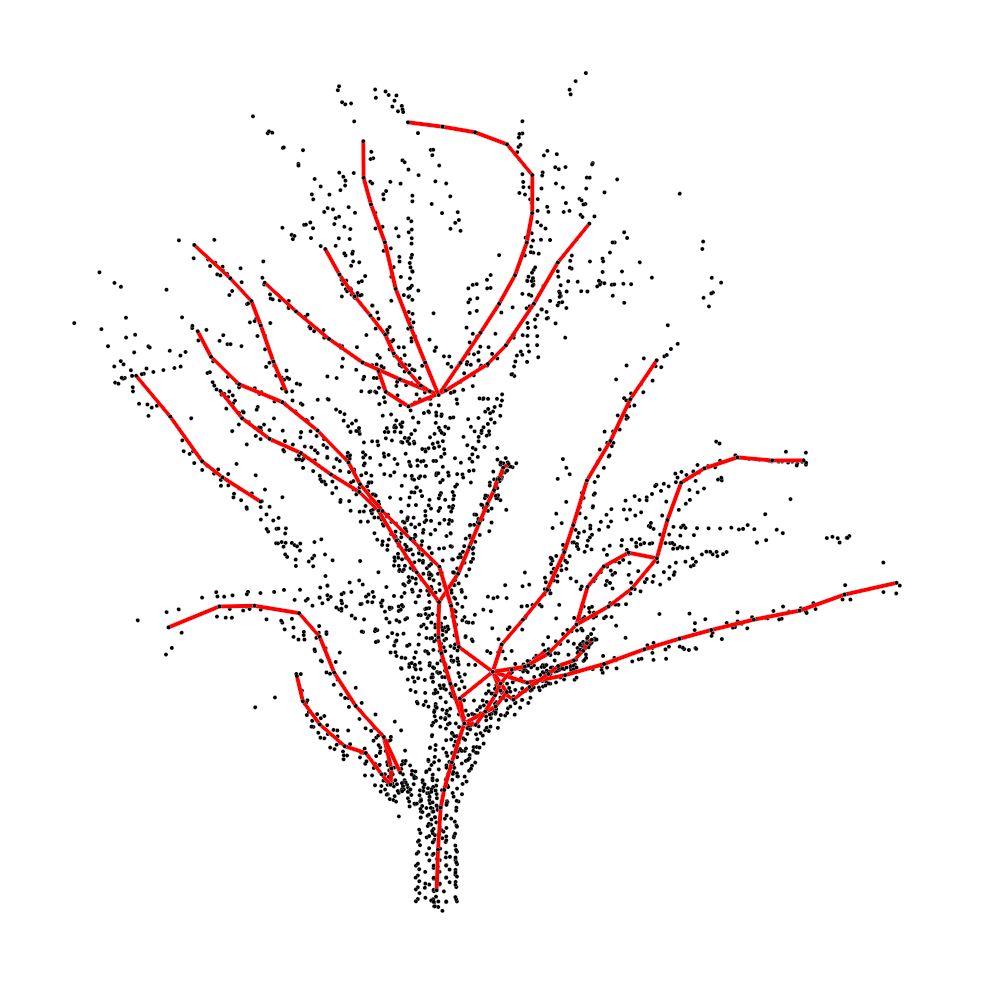}
	}
	{
		\includegraphics[width=0.15\textwidth]{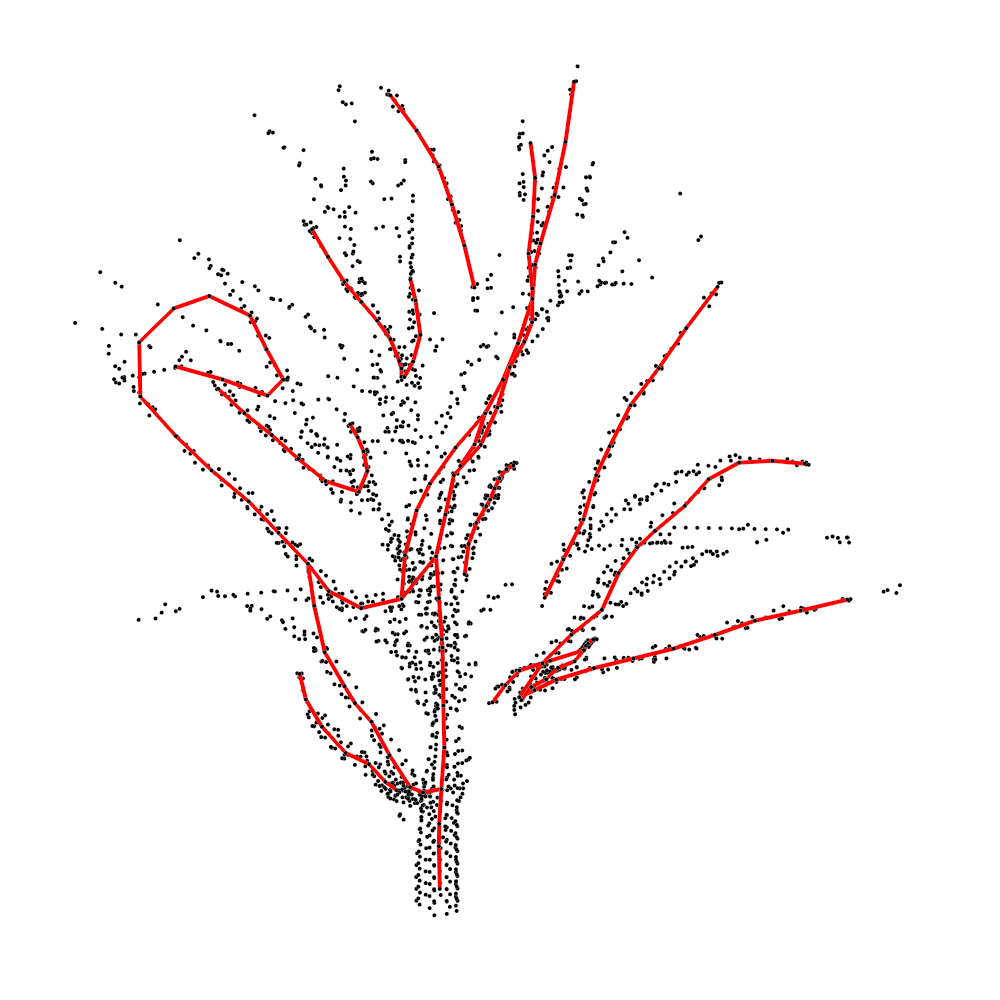}
	}
	{
		\includegraphics[width=0.15\textwidth]{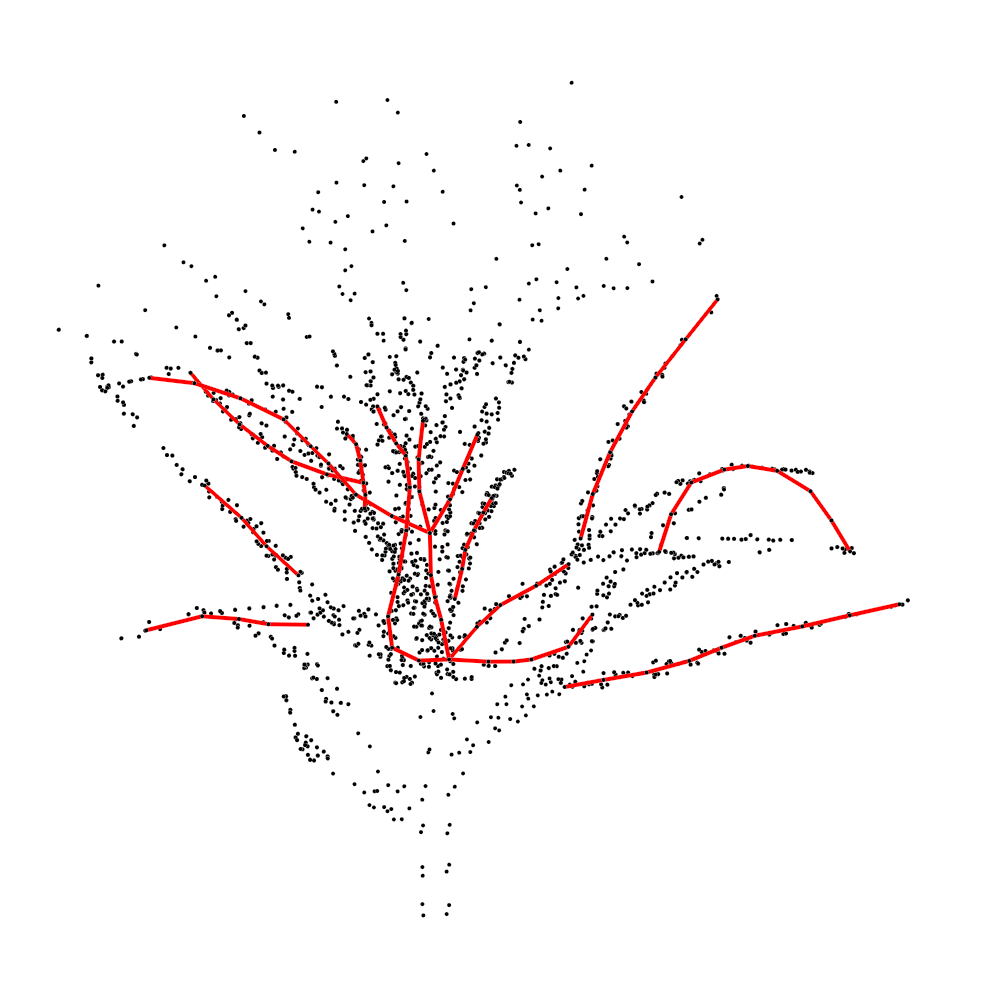}
	}

	\caption{ Skeleton extraction comparisons of point clouds with tree-1,tree-2 and tree-3. The first column shows the ground truth skeletons. The skeletons extracted by $L_1$ method from uniform point cloud are shown in second column.  The skeletons extracted from point clouds with sparse density, noise, missing data, and non-uniform density are shown in the third and fourth, fifth and sixth columns.     }
	\label{result}
\end{figure}

Since the large-sized trees are too complex for $L_1$ method. We select three tree point clouds from small and medium sizes with increased complexity named tree-1,tree-2 and tree-3. In Fig \ref{result}, we validate the performance of $L_1$ method on the point clouds which contain sparse density, noise, missing data, and non-uniform density and evaluate the $L_1$ method from visually and quantitative aspects.

The results show that our dataset is challenging. The first column shows the ground truth skeletons of the tree point clouds. In the second column, when the branch complexity of the tree is low, the $L_1$ method can propose a relatively complete skeleton on a uniformly clean point cloud. With the increasing of branch complexity, the topology correctness of the extracted skeleton decreases. We observe that the $L_1$ method is very sensitive to the variation of point cloud density. In the third column, the skeleton extracted from the sparse density point cloud is generally not effective. In the case of noise, owing to the tight connectivity of the tree branches, $L_1$ treats the noise points as the points of the tree and generates incorrect skeleton point connections. In the absence of data, this method misses fine-scale structure and produces the wrong skeleton. Skeleton extracted from point clouds with uneven density distribution contain several incorrect branches or lose certain branches.

For quantitative comparison, we denote the ground truth  skeleton as $G=(V_g, E_g)$, and the result of the point cloud extraction skeleton algorithm is expressed as $S=(V_s, E_s)$. The purpose of our evaluation is to evaluate the topological correctness of $S$ (the accuracy of the branch and endpoint of $S$) and the centeredness of $S$ (the proximity of $S$ to $G$).

Hausdorff distance\cite{memoli2005a} is a measure of similarity between two sets of points. We use Hausdorff distance to measure the maximum mismatch between the skeleton extracted from the point cloud and the ground truth skeleton:
\begin{equation}
H D(G, S)=\max (\max_{v_{g} \in V_{g}} \min _{v_{s} \in V_{s}} \|v_g-v_s\|, \max_{v_{s} \in V_{s}}\min _{v_{g} \in V_{g}} \|v_s-v_g\|)
\end{equation}

\begin{figure}[htb]
	\centering
	\vfill
	{
		\includegraphics[width=0.5\textwidth]{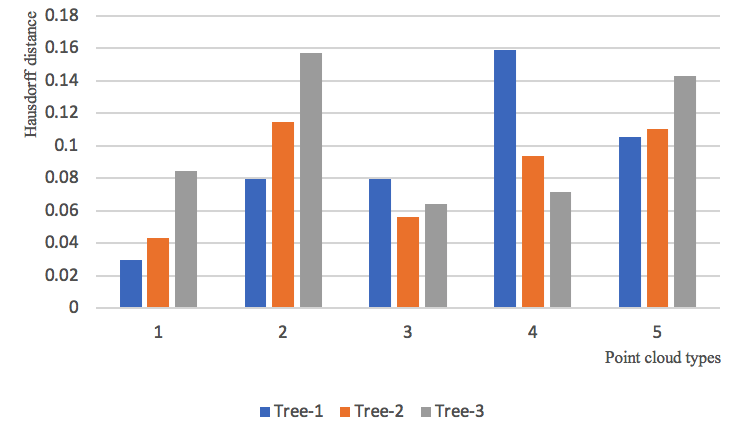}
	}
	\caption{ Haudorff distance between ground truth curve skeletons and the extracted curve skeletons with respect to different tree models.
		Number 1 to 5 represent the skeletons extracted by $L_1$ method from point clouds with sparse density, noise, missing data and uneven density distribution, respectively.
		}
	\label{hausdorff}
\end{figure}

In Fig \ref{hausdorff}, we give the Hausdorff distance value between the ground truth skeletons and the skeletons extracted from different point clouds.

\section{Conclusion}
In this paper, we introduce the process of constructing a tree-structured point cloud dataset. Using implicit surface as an intermediate representation to sample the point cloud from mesh model. At the same time, considering the challenges of skeleton extraction algorithms, we construct four different types of point cloud models. As far as we know, we are the first to propose 3D point cloud dataset with ground truth skeleton. Our dataset is publicly available for evaluation between skeleton extraction algorithms and further research in other fields.

In future work, we plan to build other types of models to augment the standard point cloud dataset. Furthermore, our goal is to propose a comprehensive and consistent quantitative evaluation system to better understand the advantages and limitations of the existing point cloud skeleton extraction methods.
\section*{Acknowledgements}
This work has been partially funded by the Chongqing Research Program of Basic Research and Frontier Technology grant No.cstc2019jcyj-msxmX0033, the National Natural Science Foundation of China grant No.61701051, the Fundamental Research Funds for the Central Universities grant No.2019CDYGYB012.
\bibliographystyle{unsrt} 
\bibliography{mybibfile}

\begin{thebibliography}{10}

\bibitem{Tagliasacchi2009Curve}
Andrea Tagliasacchi, Hao Zhang, and Daniel Cohen-Or.
\newblock Curve skeleton extraction from incomplete point cloud.
\newblock {\em Acm Transactions on Graphics}, 28(3):1--9, 2009.

\bibitem{huang2013l}
Hui Huang, Shihao Wu, Daniel Cohenor, Minglun Gong, Hao Zhang, Guiqing Li, and
  Baoquan Chen.
\newblock L 1 -medial skeleton of point cloud.
\newblock {\em international conference on computer graphics and interactive
  techniques}, 32(4):65, 2013.

\bibitem{Jie20163D}
Mei Jie, Liqiang Zhang, Shihao Wu, Wang Zhen, and Zhang Liang.
\newblock 3d tree modeling from incomplete point clouds via optimization and
  l1-mst.
\newblock {\em International Journal of Geographical Information Science},
  31(5):1--23, 2016.

\bibitem{Song2018Distance}
Chengfang Song, Zhiqiang Pang, Xiaoyuan Jing, and Chunxia Xiao.
\newblock Distance field guided \(l_1\) -median skeleton extraction.
\newblock {\em Visual Computer}, 34(2):243--255, 2018.

\bibitem{Liu_2019_CVPR_Workshops}
Chang Liu, Dezhao Luo, Yifei Zhang, Wei Ke, Fang Wan, and Qixiang Ye.
\newblock Parametric skeleton generation via gaussian mixture models.
\newblock In {\em The IEEE Conference on Computer Vision and Pattern
  Recognition (CVPR) Workshops}, June 2019.

\bibitem{Yang_2019_CVPR_Workshops}
Liping Yang, Diane Oyen, and Brendt Wohlberg.
\newblock A novel algorithm for skeleton extraction from images using
  topological graph analysis.
\newblock In {\em The IEEE Conference on Computer Vision and Pattern
  Recognition (CVPR) Workshops}, June 2019.

\bibitem{Atienza_2019_CVPR_Workshops}
Rowel Atienza.
\newblock Pyramid u-network for skeleton extraction from shape points.
\newblock In {\em The IEEE Conference on Computer Vision and Pattern
  Recognition (CVPR) Workshops}, June 2019.

\bibitem{Demir_2019_CVPR_Workshops}
Ilke Demir, Camilla Hahn, Kathryn Leonard, Geraldine Morin, Dana Rahbani,
  Athina Panotopoulou, Amelie Fondevilla, Elena Balashova, Bastien Durix, and
  Adam Kortylewski.
\newblock Skelneton 2019: Dataset and challenge on deep learning for geometric
  shape understanding.
\newblock In {\em The IEEE Conference on Computer Vision and Pattern
  Recognition (CVPR) Workshops}, June 2019.

\bibitem{Ogniewicz1992Voronoi}
R.~Ogniewicz and M.~Ilg.
\newblock Voronoi skeletons: Theory and applications.
\newblock In {\em IEEE Computer Society Conference on Computer Vision \&
  Pattern Recognition, Cvpr}, 1992.

\bibitem{Bucksch2008CAMPINO}
Alexander Bucksch and Roderik Lindenbergh.
\newblock Campino — a skeletonization method for point cloud processing.
\newblock {\em Isprs Journal of Photogrammetry \& Remote Sensing},
  63(1):115--127, 2008.

\bibitem{Natali2011Graph}
Mattia Natali, Silvia Biasotti, Giuseppe Patanè, and Bianca Falcidieno.
\newblock Graph-based representations of point clouds.
\newblock {\em Graphical Models}, 73(5):151--164, 2011.

\bibitem{Bucksch2010SkelTre}
Alexander Bucksch, Roderik Lindenbergh, and Massimo Menenti.
\newblock Skeltre.
\newblock {\em Visual Computer}, 26(10):1283--1300, 2010.

\bibitem{Livny2010Automatic}
Yotam Livny, Feilong Yan, Matt Olson, Baoquan Chen, Zhang Hao, and Jihad
  El-Sana.
\newblock Automatic reconstruction of tree skeletal structures from point
  clouds.
\newblock {\em Acm Transactions on Graphics}, 29(6):1--8, 2010.

\bibitem{Sharf2010On}
Andrei Sharf, Thomas Lewiner, Ariel Shamir, and Leif Kobbelt.
\newblock On-the-fly curve-skeleton computation for 3d shapes.
\newblock {\em Computer Graphics Forum}, 26(3):323--328, 2010.

\bibitem{Guo2010Analysis}
Li~Guo, Ligang Liu, Hanlin Zheng, and Niloy~J. Mitra.
\newblock Analysis, reconstruction and manipulation using arterial snakes.
\newblock {\em Acm Transactions on Graphics}, 29(6):1, 2010.

\bibitem{Cao2010Point}
Junjie Cao, Andrea Tagliasacchi, Matt Olson, Zhang Hao, and Zhinxun Su.
\newblock Point cloud skeletons via laplacian based contraction.
\newblock In {\em Shape Modeling International Conference}, 2010.

\bibitem{au2008skeleton}
Oscar~Kinchung Au, Chiewlan Tai, Hungkuo Chu, Daniel Cohenor, and Tongyee Lee.
\newblock Skeleton extraction by mesh contraction.
\newblock {\em international conference on computer graphics and interactive
  techniques}, 27(3):44, 2008.

\bibitem{A}
Zhen Wang, Liqiang Zhang, Tian Fang, P~Takis Mathiopoulos, Huamin Qu, Dong
  Chen, and Yuebin Wang.
\newblock A structure-aware global optimization method for reconstructing 3-d
  tree models from terrestrial laser scanning data.
\newblock {\em IEEE Transactions on Geoscience \& Remote Sensing},
  52(9):5653--5669, 2014.

\bibitem{Zhen2016A}
Wang Zhen, Liqiang Zhang, Fang Tian, Xiaohua Tong, and Mei Jie.
\newblock A local structure and direction-aware optimization approach for
  three-dimensional tree modeling.
\newblock {\em IEEE Transactions on Geoscience \& Remote Sensing},
  54(8):4749--4757, 2016.

\bibitem{Qin2019Mass}
Hongxing Qin, Jia Han, Ning Li, Hui Huang, and Baoquan Chen.
\newblock Mass-driven topology-aware curve skeleton extraction from incomplete
  point clouds.
\newblock {\em IEEE Transactions on Visualization and Computer Graphics},
  PP(99):1--1, 2019.

\bibitem{Zhang20163D}
Dejia Zhang, Xie Ning, Liang Shuang, and Jinyuan Jia.
\newblock 3d tree skeletonization from multiple images based on pyrlk optical
  flow.
\newblock {\em Pattern Recognition Letters}, 76(C):49--58, 2016.

\bibitem{demir2019skelneton}
Ilke Demir, Camilla Hahn, Kathryn Leonard, Geraldine Morin, Dana Rahbani,
  Athina Panotopoulou, Amelie Fondevilla, Elena Balashova, Bastien Durix, and
  Adam Kortylewski.
\newblock Skelneton 2019: Dataset and challenge on deep learning for geometric
  shape understanding.
\newblock pages 0--0, 2019.

\bibitem{durix2019the}
Bastien Durix, Sylvie Chambon, Kathryn Leonard, Jeanluc Mari, and Geraldine
  Morin.
\newblock The propagated skeleton: a robust detail-preserving approach.
\newblock pages 343--354, 2019.

\bibitem{ohtake2003multi-level}
Yutaka Ohtake, Alexander Belyaev, Marc Alexa, Greg Turk, and Hanspeter Seidel.
\newblock Multi-level partition of unity implicits.
\newblock {\em international conference on computer graphics and interactive
  techniques}, 22(3):463--470, 2003.

\bibitem{berger2013a}
Matthew Berger, Joshua~A Levine, Luis~Gustavo Nonato, Gabriel Taubin, and
  Claudio~T Silva.
\newblock A benchmark for surface reconstruction.
\newblock {\em ACM Transactions on Graphics}, 32(2):20, 2013.

\bibitem{shen2005interpolating}
Chen Shen, James~F Obrien, and Jonathan~Richard Shewchuk.
\newblock Interpolating and approximating implicit surfaces from polygon soup.
\newblock {\em international conference on computer graphics and interactive
  techniques}, 23(3):896--904, 2005.

\bibitem{vlasic2009dynamic}
Daniel Vlasic, Pieter Peers, Ilya Baran, Paul Debevec, Jovan Popovic, Szymon
  Rusinkiewicz, and Wojciech Matusik.
\newblock Dynamic shape capture using multi-view photometric stereo.
\newblock {\em international conference on computer graphics and interactive
  techniques}, 28(5):174, 2009.

\bibitem{brown2007global}
Benedict Brown and Szymon Rusinkiewicz.
\newblock Global non-rigid alignment of 3-d scans.
\newblock {\em international conference on computer graphics and interactive
  techniques}, 26(3):21, 2007.

\bibitem{hoppe1992surface}
Hugues Hoppe, Tony Derose, Tom Duchamp, John~W Mcdonald, and Werner Stuetzle.
\newblock Surface reconstruction from unorganized points.
\newblock {\em international conference on computer graphics and interactive
  techniques}, 26(2):71--78, 1992.

\bibitem{memoli2005a}
Facundo Memoli and Guillermo Sapiro.
\newblock A theoretical and computational framework for isometry invariant
  recognition of point cloud data.
\newblock {\em Foundations of Computational Mathematics}, 5(3):313--347, 2005.

\end{thebibliography}

\appendix
\appendixpage
\section{}
We have added supplementary material in the Appendix A to show the diversity of our point cloud dataset, which contains ground truth skeletons, point cloud with uniformly distribution, sparse density, noise and missing data, uneven density distribution.
\begin{figure}[htb]
	\centering
	 {
		\includegraphics[width=0.15\textwidth]{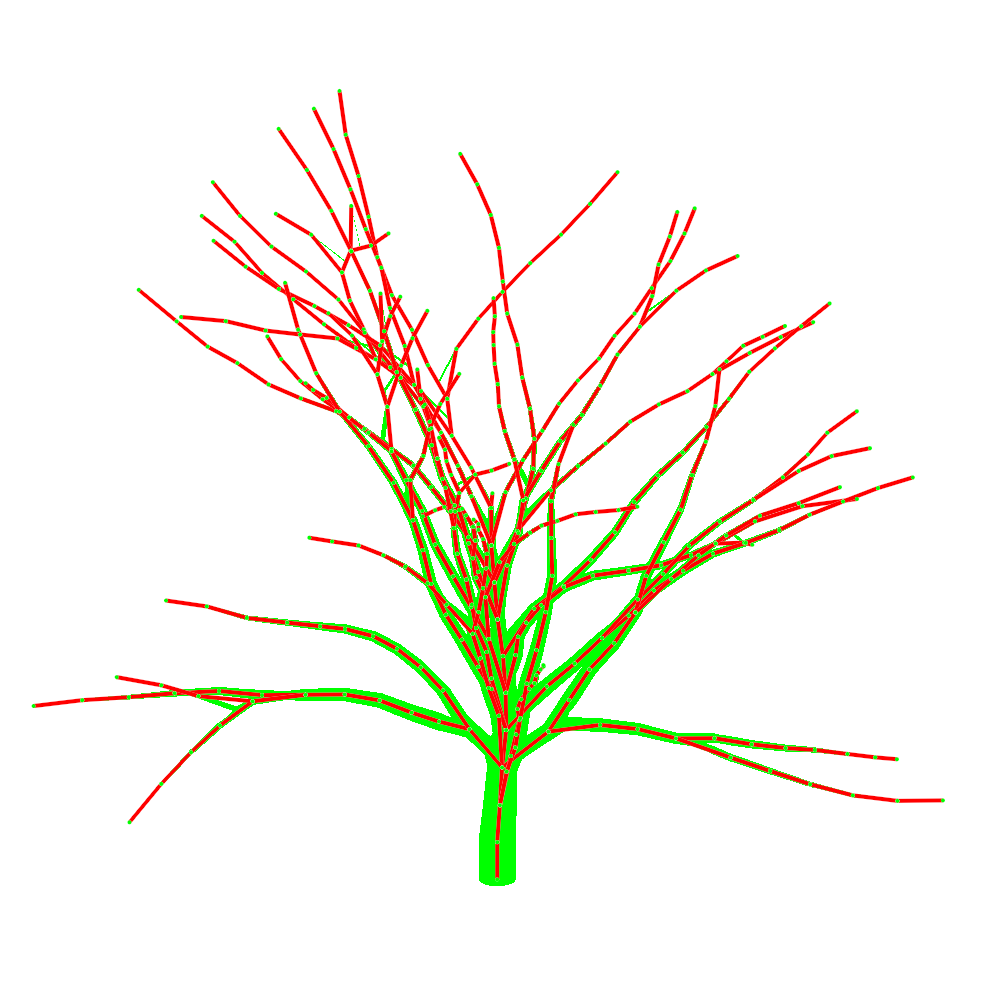}
	}
	{
			\includegraphics[width=0.15\textwidth]{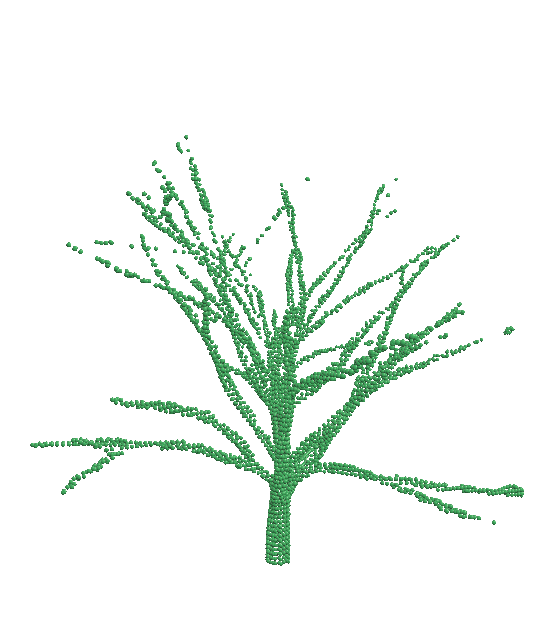}
		}
	{
		\includegraphics[width=0.15\textwidth]{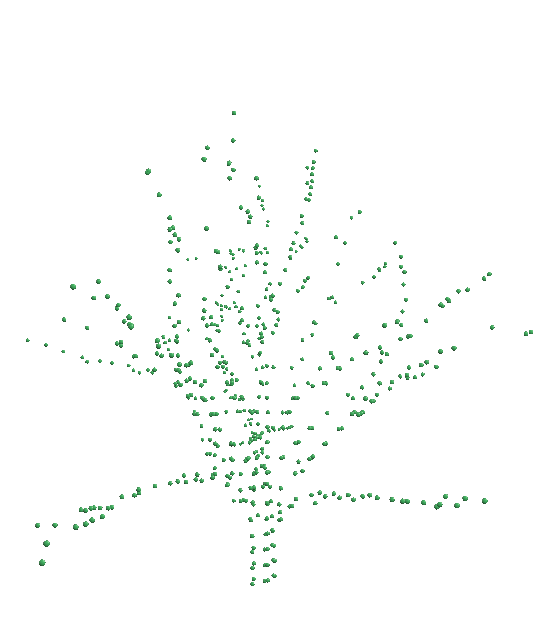}
	}
	 {
		\includegraphics[width=0.15\textwidth]{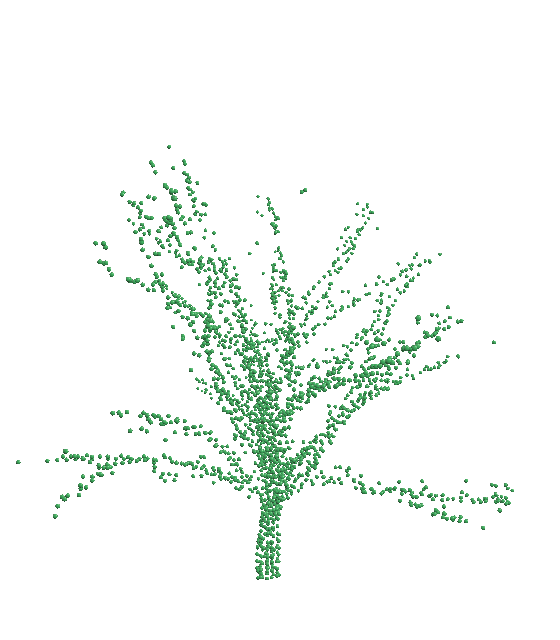}
	}			
	{
	\includegraphics[width=0.15\textwidth]{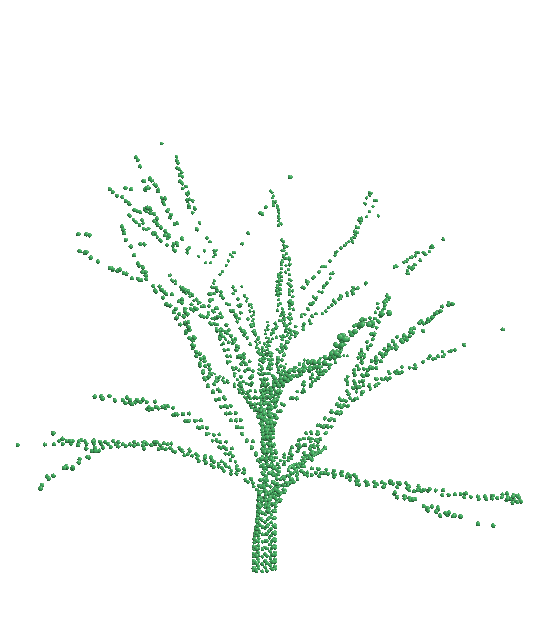}
	}	 
{
	\includegraphics[width=0.15\textwidth]{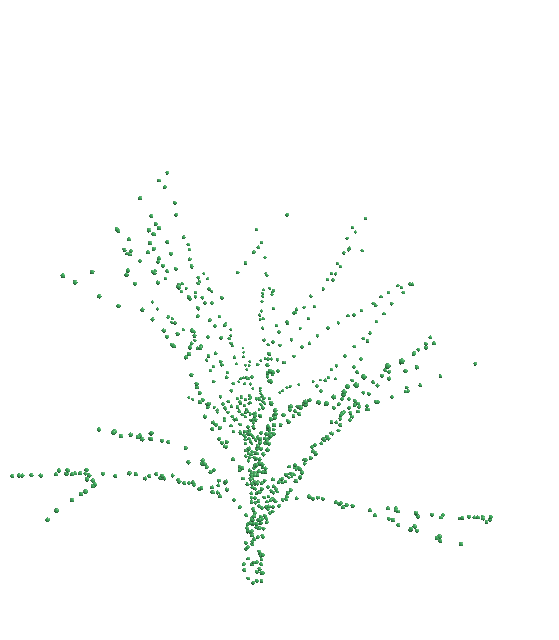}
}

\vfill
\centering
 {
		\includegraphics[width=0.15\textwidth]{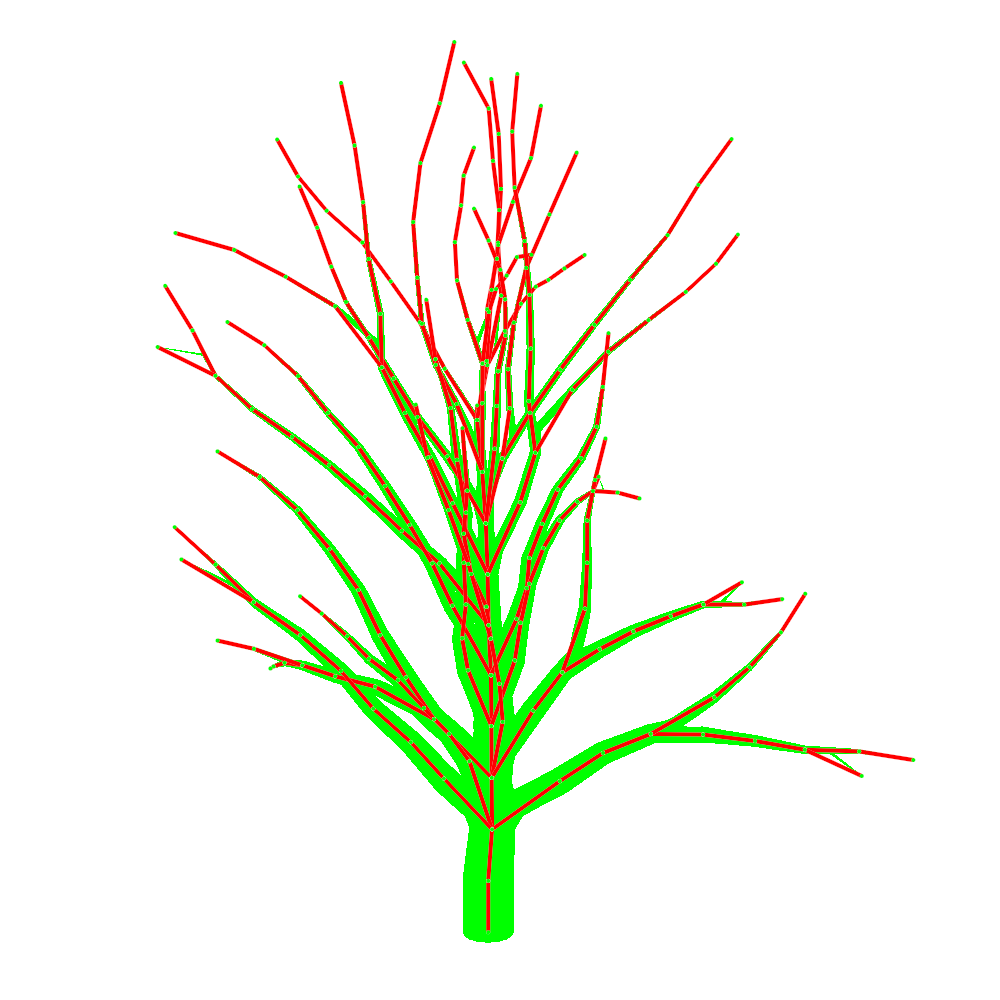}
	}
	 {
		\includegraphics[width=0.15\textwidth]{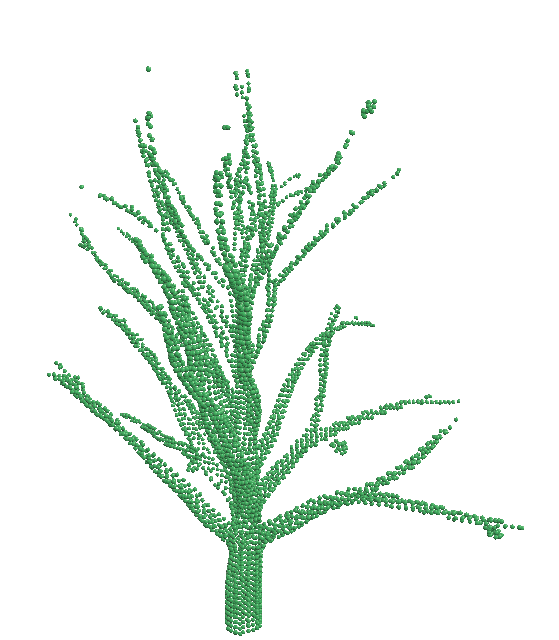}
	}
 {
		\includegraphics[width=0.15\textwidth]{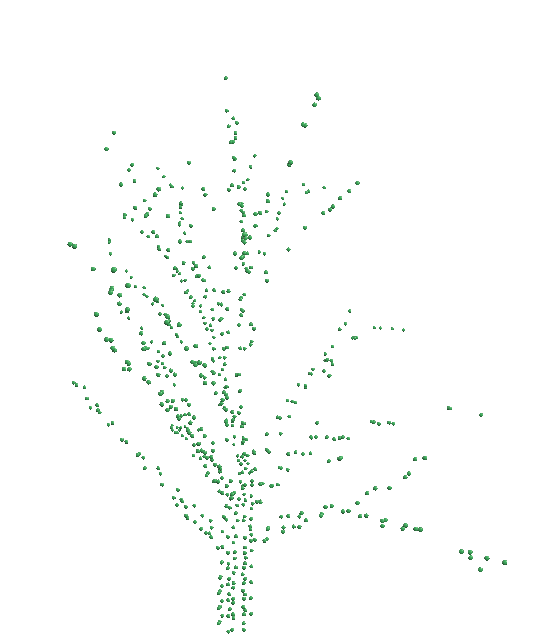}
	}
 {
		\includegraphics[width=0.15\textwidth]{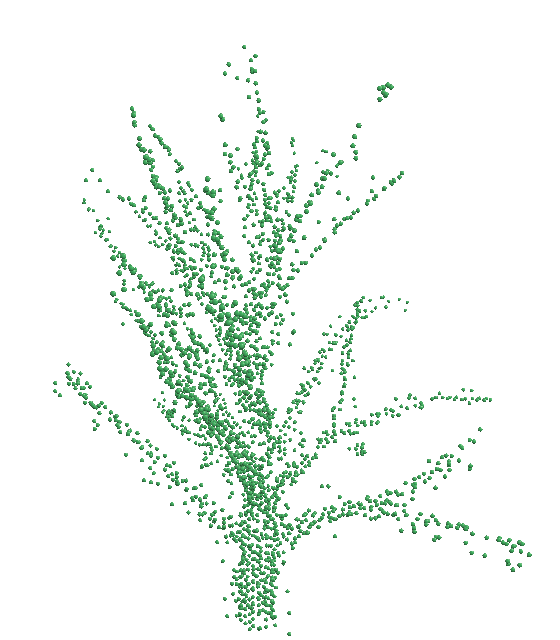}
	}	
{
		\includegraphics[width=0.15\textwidth]{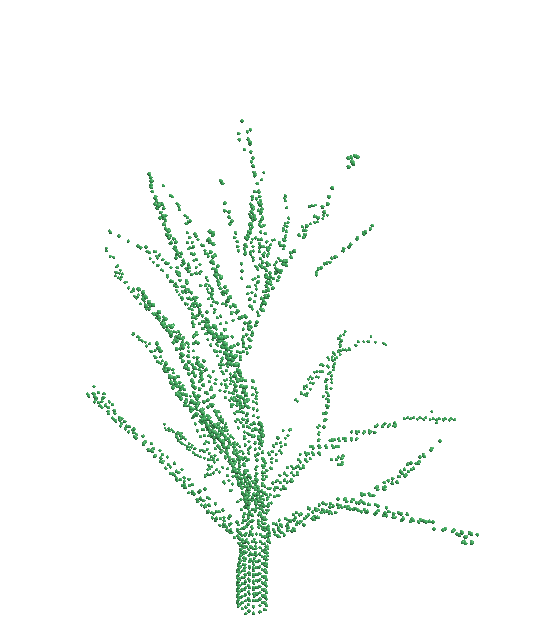}
	 }
{
		\includegraphics[width=0.15\textwidth]{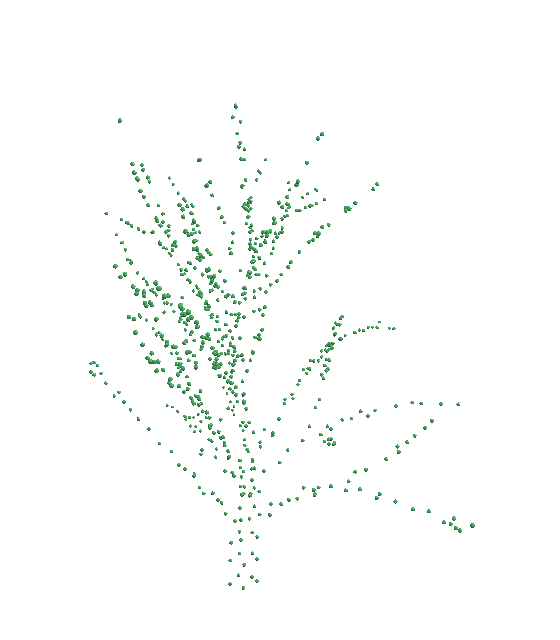}
	}		 
\vfill
	\centering
 {
		\includegraphics[width=0.15\textwidth]{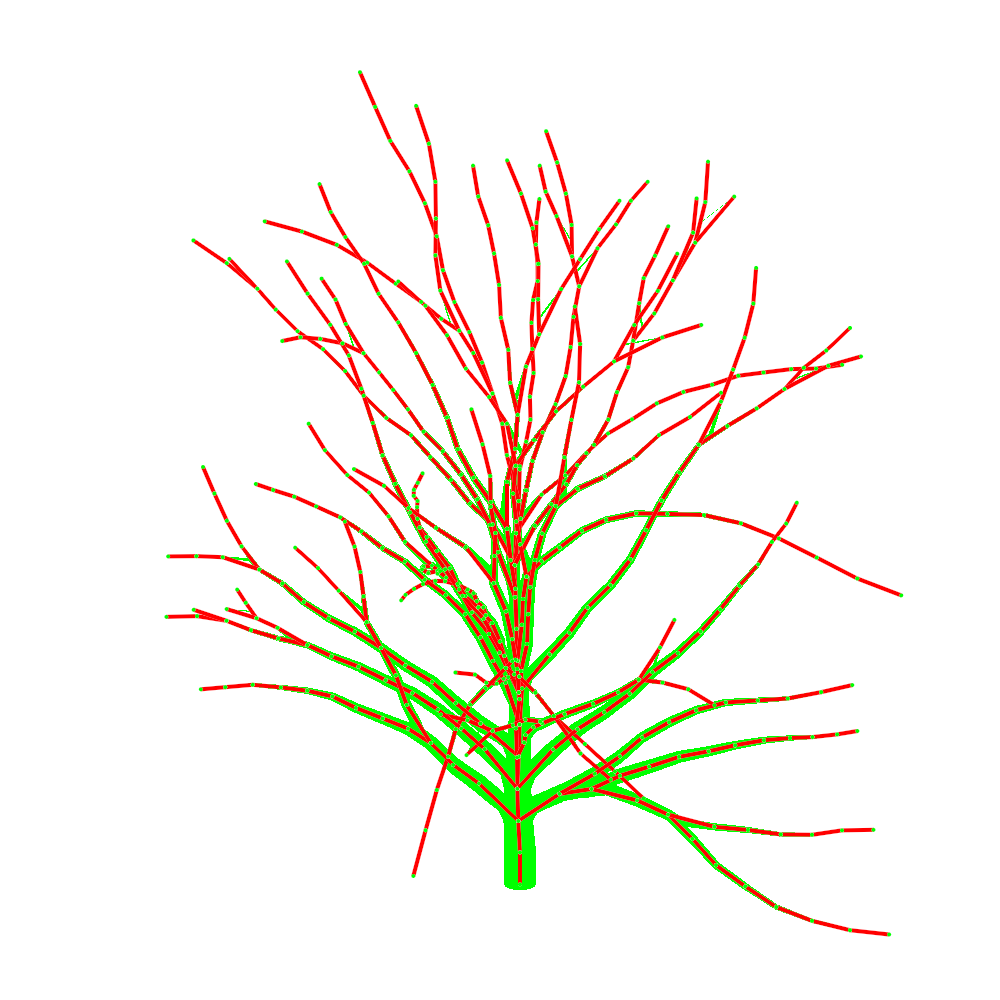}
	}
	 {
		\includegraphics[width=0.15\textwidth]{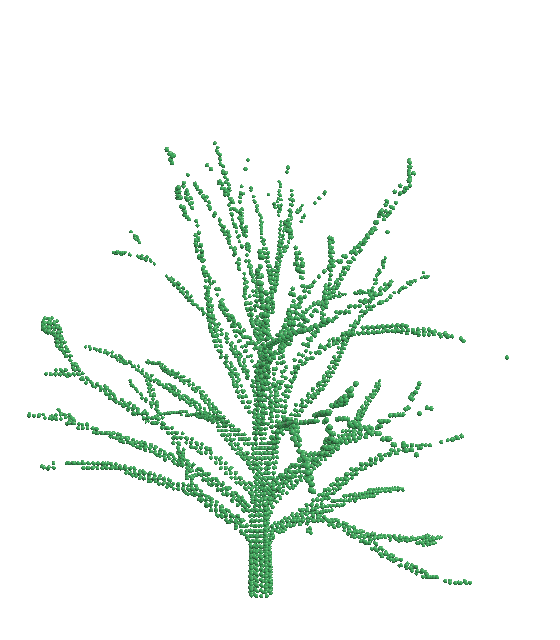}
	}
 {
		\includegraphics[width=0.15\textwidth]{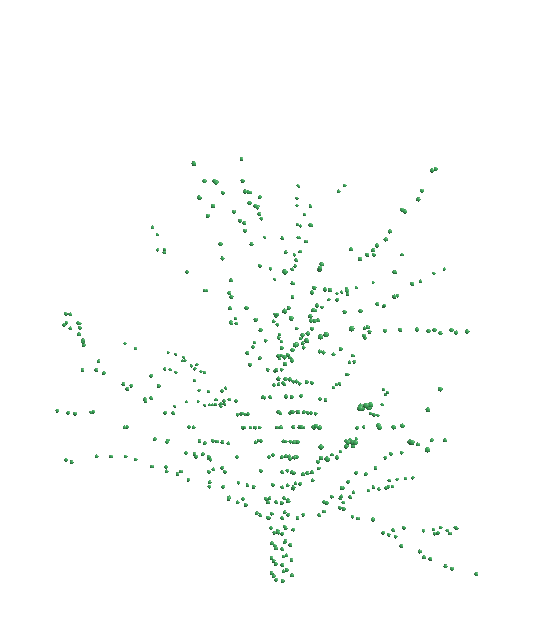}
	}
 {
 	\includegraphics[width=0.15\textwidth]{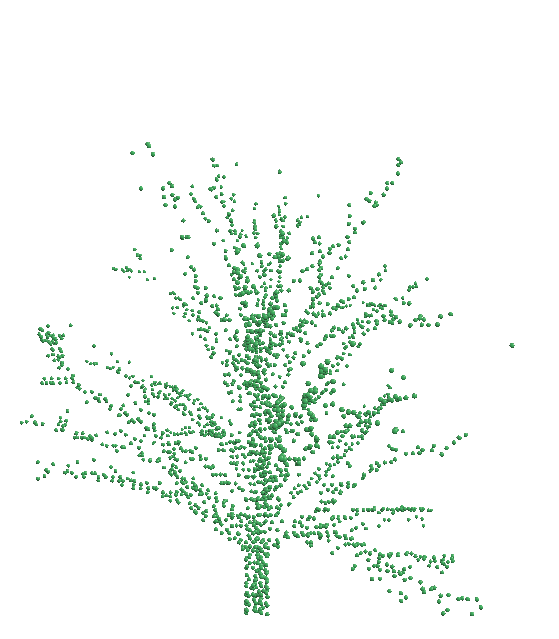}
 }
{
	\includegraphics[width=0.15\textwidth]{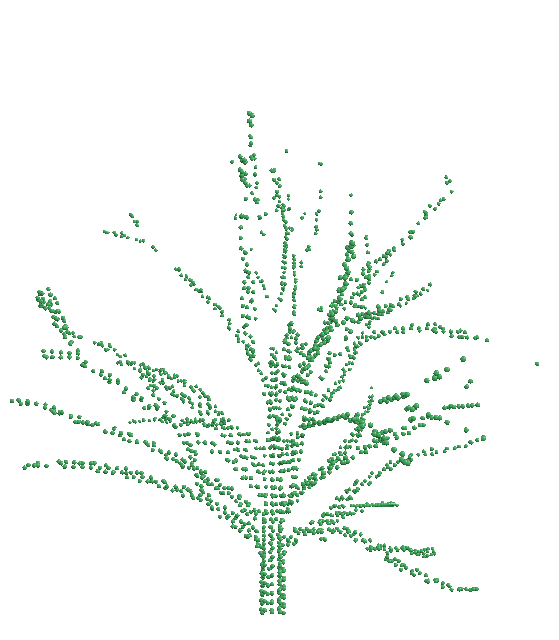}
		 }
	{
		\includegraphics[width=0.15\textwidth]{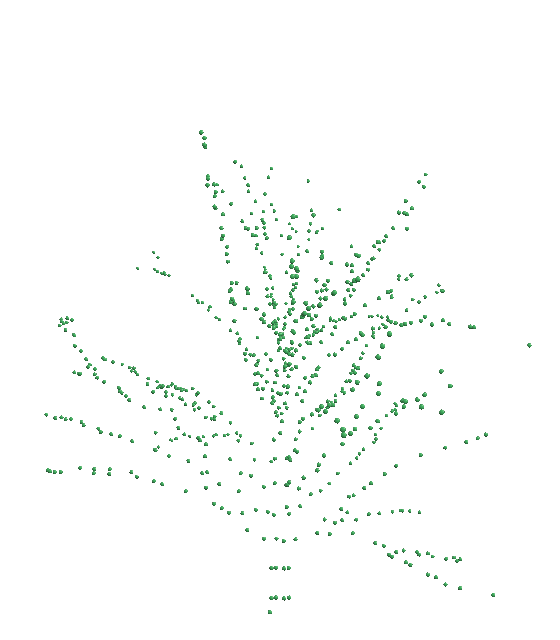}
	}
\vfill
	\centering
 {
	\includegraphics[width=0.15\textwidth]{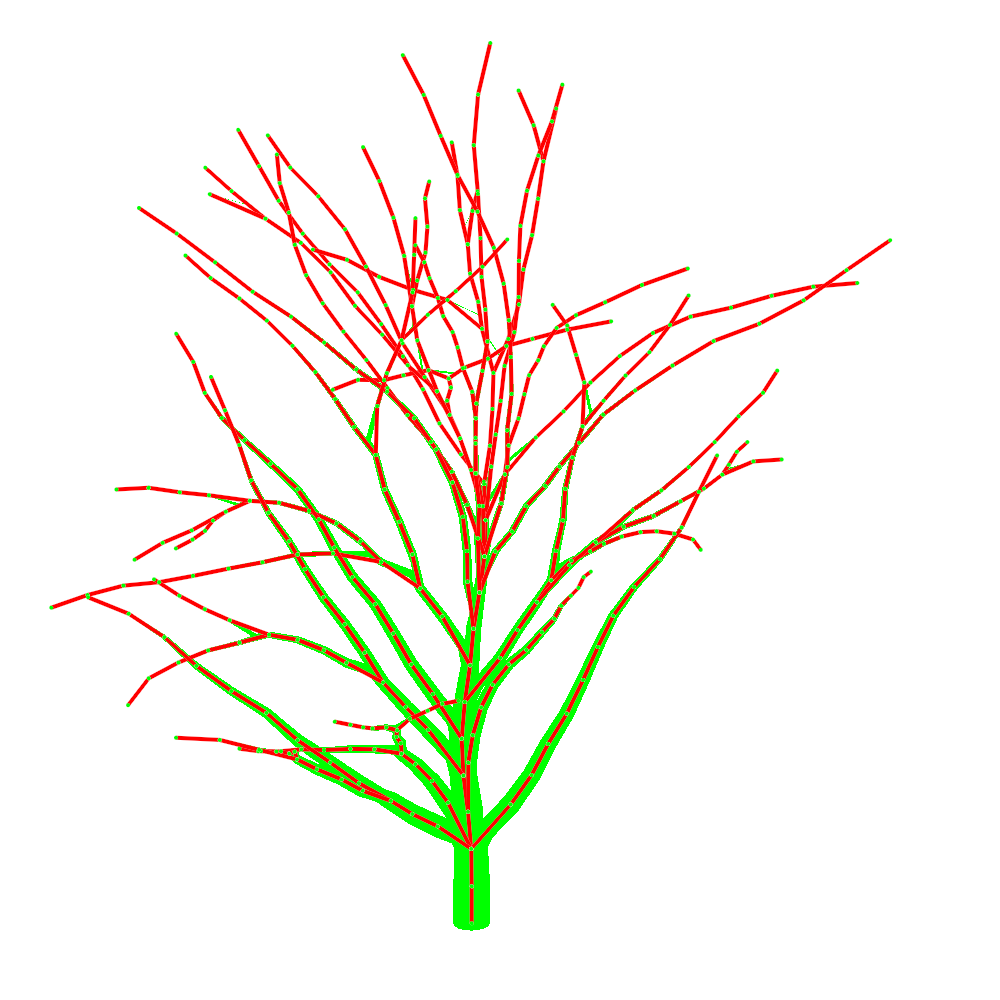}
}
 {
		\includegraphics[width=0.15\textwidth]{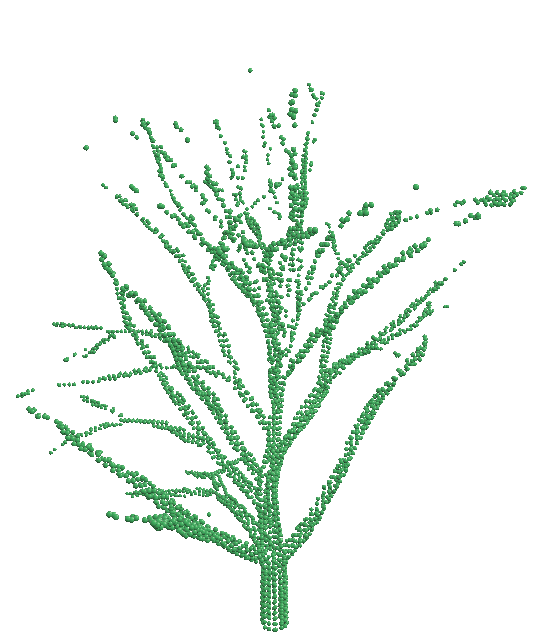}
	}
 {
		\includegraphics[width=0.15\textwidth]{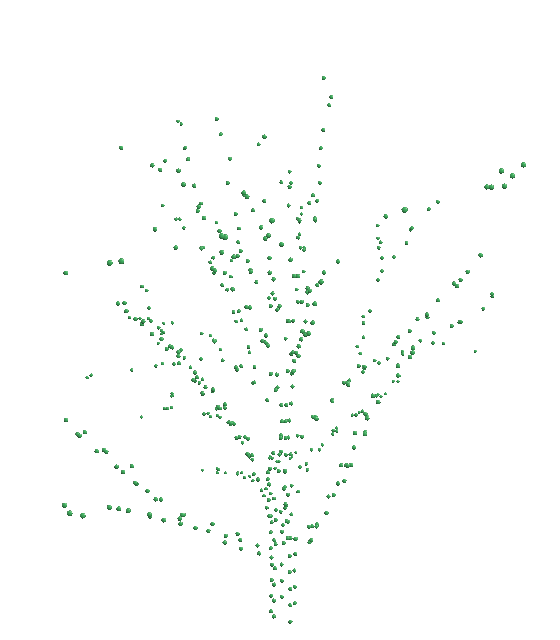}
	}
 {
		\includegraphics[width=0.15\textwidth]{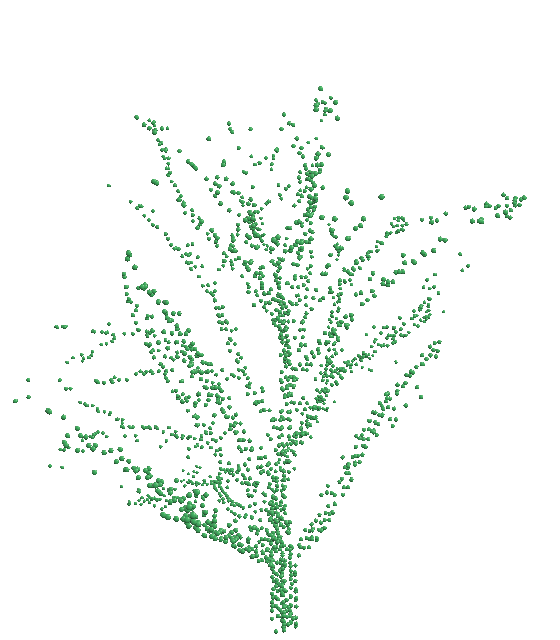}
}
	{
		\includegraphics[width=0.15\textwidth]{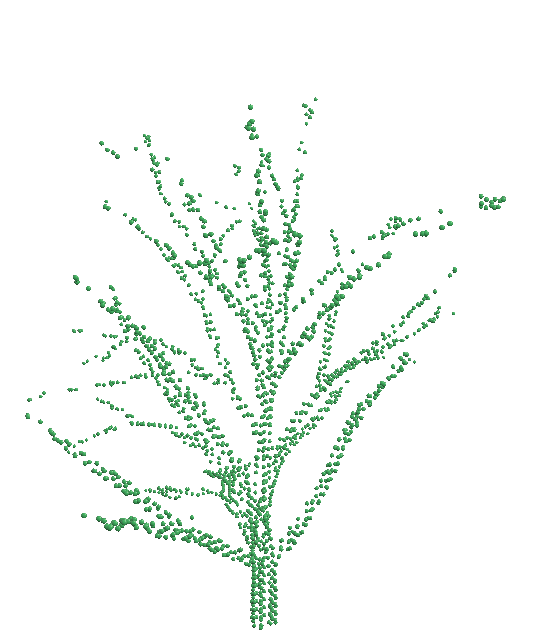}
	}
{
		\includegraphics[width=0.15\textwidth]{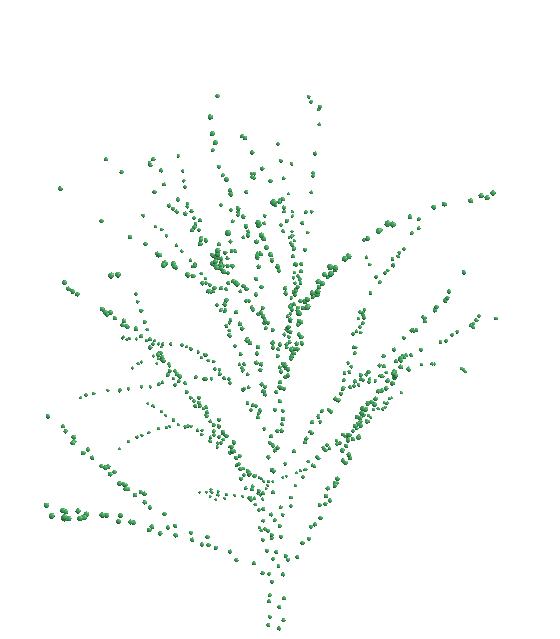}
	}

		\vfill
		\centering
 {
		\includegraphics[width=0.15\textwidth]{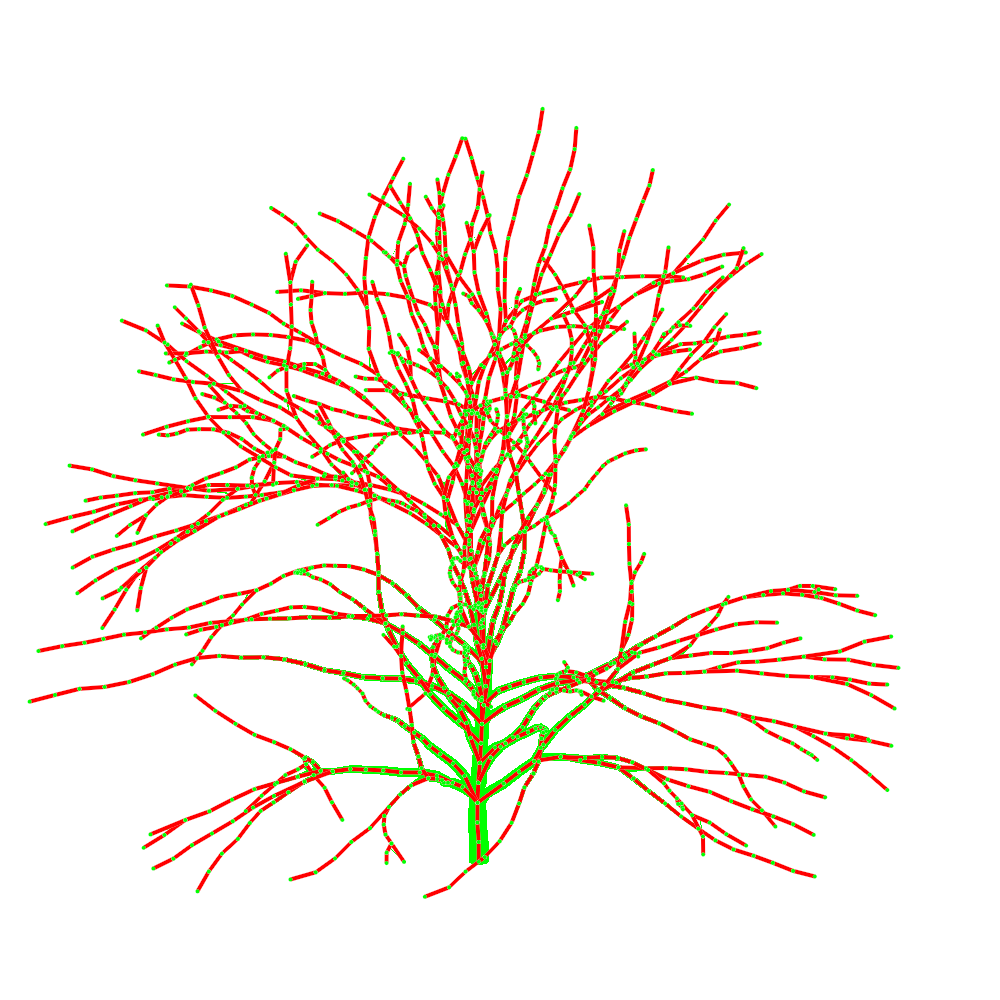}
	}
	 {
		\includegraphics[width=0.15\textwidth]{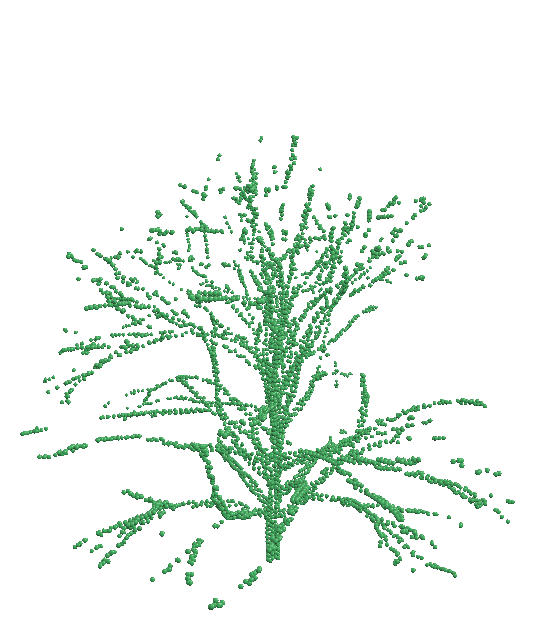}
	}
{
		\includegraphics[width=0.15\textwidth]{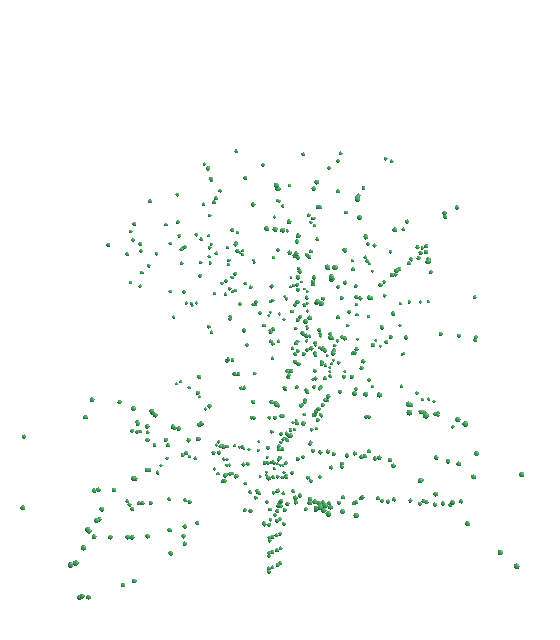}
	}
{
	\includegraphics[width=0.15\textwidth]{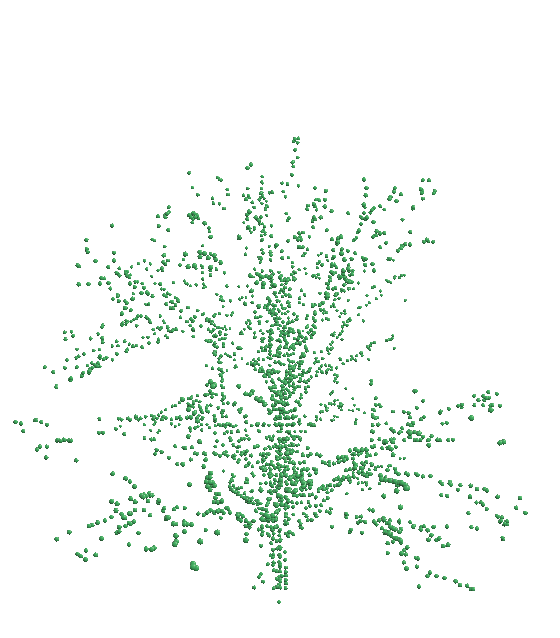}
}
 {
	\includegraphics[width=0.15\textwidth]{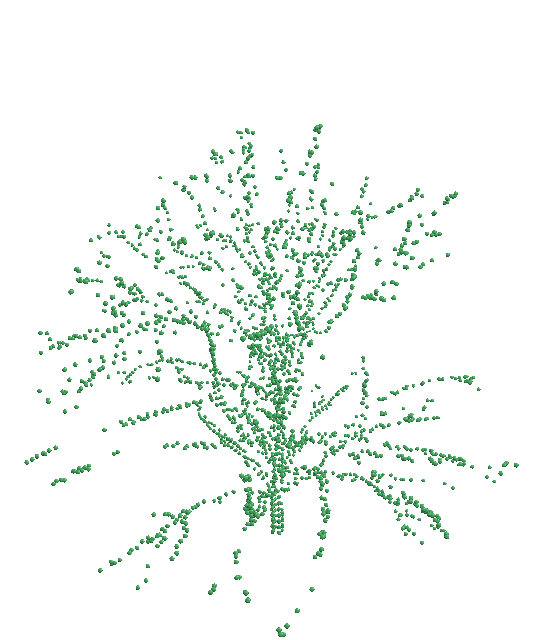}
		}
{
	\includegraphics[width=0.15\textwidth]{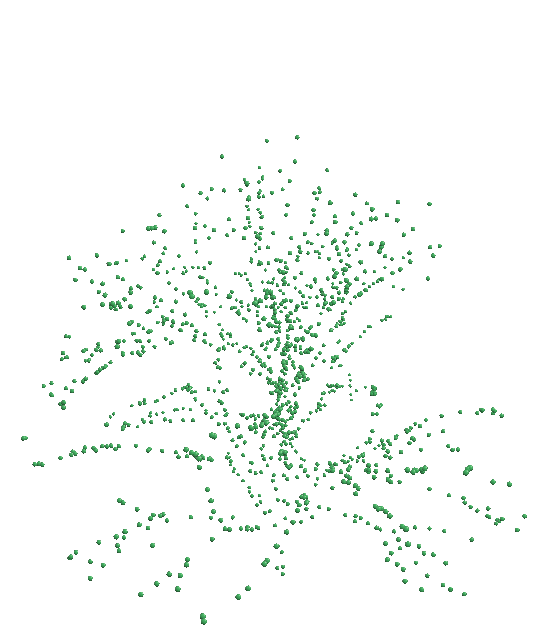}
			}

\vfill
\centering
\subfloat[skeleton model] {
		\includegraphics[width=0.15\textwidth]{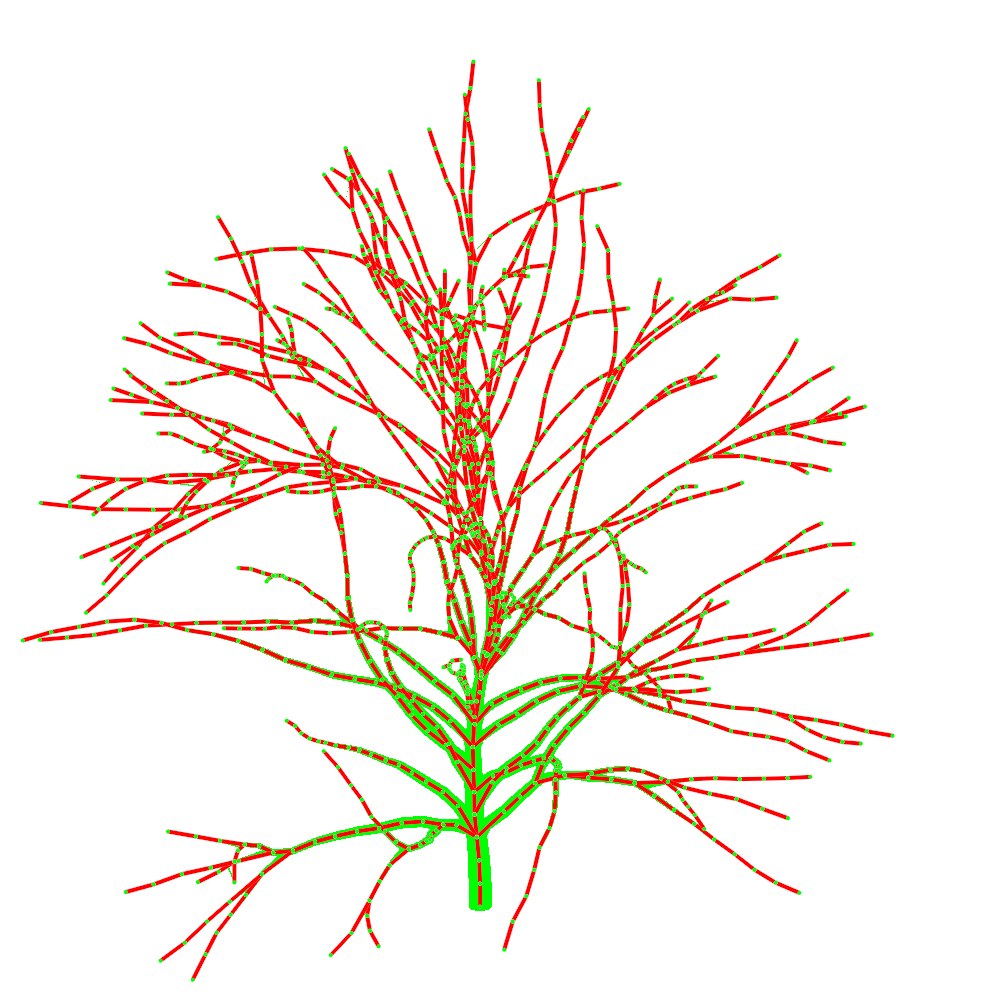}
	}
\subfloat[Point cloud] {
		\includegraphics[width=0.15\textwidth]{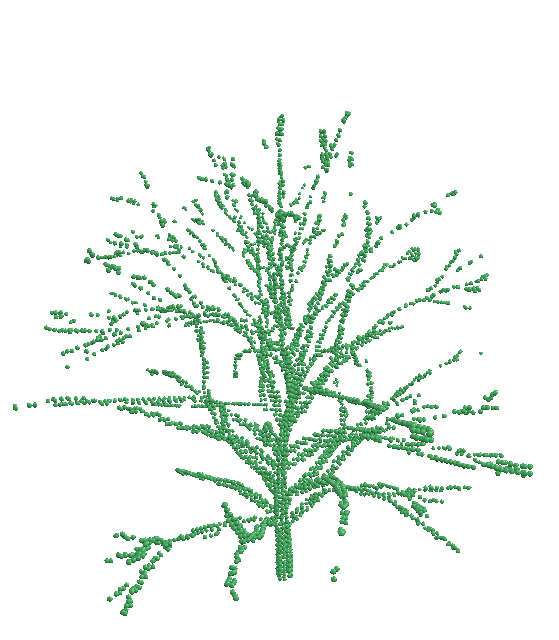}
	}
\subfloat[Defferrnt density] {
		\includegraphics[width=0.15\textwidth]{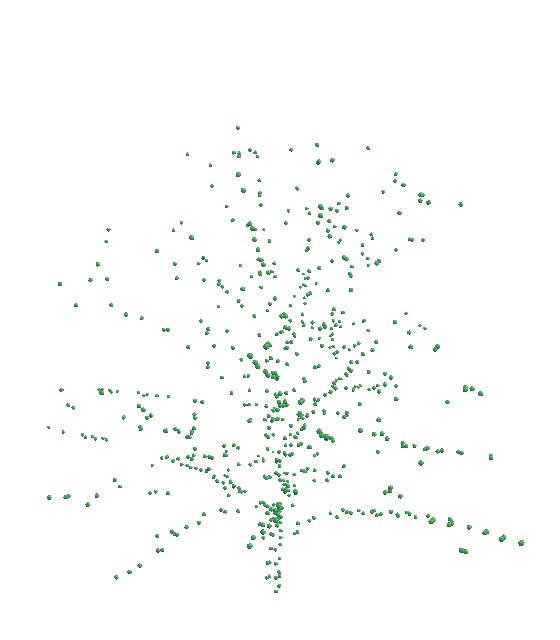}
	}
  \subfloat[Noise data]{
 	\includegraphics[width=0.15\textwidth]{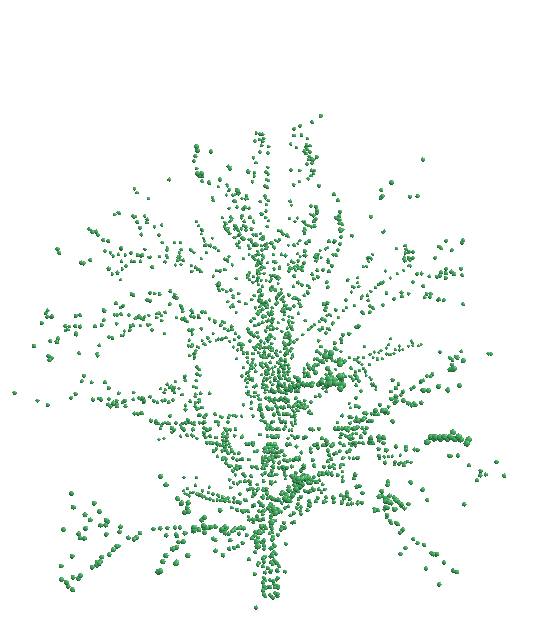}
 }
  \subfloat[Missing data]{
  \includegraphics[width=0.15\textwidth]{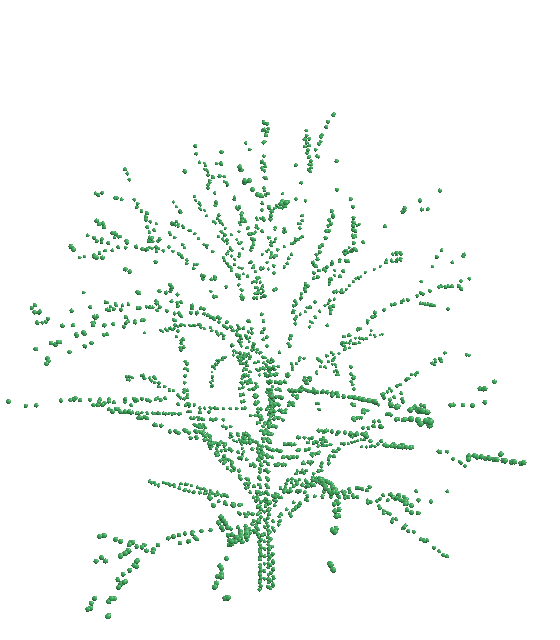}
  }
\subfloat[Uneven distribution]{
		\includegraphics[width=0.15\textwidth]{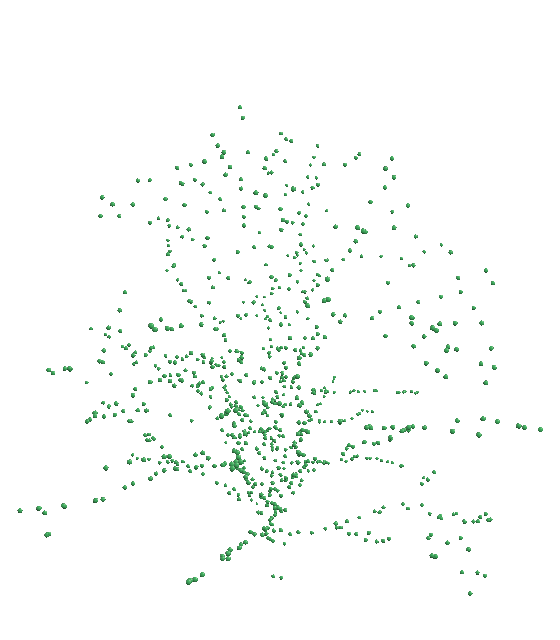}
}

\caption{ Data supplemental of tree-structured point cloud dataset.}
\label{supply}
\end{figure}

\end{document}